\pdfoutput=1

\documentclass[11pt]{article}

\usepackage[preprint]{acl}

\usepackage{times}
\usepackage{latexsym}
\usepackage{multirow}
\usepackage{url}
\usepackage{tcolorbox}  
\usepackage{xcolor}  
\usepackage{amsmath}
\tcbuselibrary{skins}
\usepackage{subcaption}
\usepackage[multiple]{footmisc}
\usepackage[T1]{fontenc}

\usepackage[utf8]{inputenc}

\usepackage{microtype}

\usepackage{inconsolata}

\usepackage{graphicx}

\newtcolorbox{mybox}
{
  enhanced jigsaw,
  drop shadow=black!50!white,
  colback=white
}

\newcommand{\smk}[1]{{\color{red}Shoumik: #1}}

\title{Almost AI, Almost Human:\\ The Challenge of Detecting AI-Polished Writing}

\author{Shoumik Saha \\
  University of Maryland \\
  College Park, USA \\
  \texttt{smksaha@umd.edu} \\\And
  Soheil Feizi \\
  University of Maryland\\
  College Park, USA \\
  \texttt{sfeizi@umd.edu} \\}

\begin{document}
\maketitle
\begin{abstract}
The growing use of large language models (LLMs) for text generation has led to widespread concerns about AI-generated content detection. However, an overlooked challenge is AI-polished text, where human-written content undergoes subtle refinements using AI tools. This raises a critical question: should minimally polished text be classified as AI-generated? Such classification can lead to false plagiarism accusations and misleading claims about AI prevalence in online content. In this study, we systematically evaluate \emph{twelve} state-of-the-art AI-text detectors using our AI-Polished-Text Evaluation \textbf{(APT-Eval) dataset}, which contains $14.7K$ samples refined at varying AI-involvement levels. Our findings reveal that detectors frequently flag even minimally polished text as AI-generated, struggle to differentiate between degrees of AI involvement, and exhibit biases against older and smaller models. These limitations highlight the urgent need for more nuanced detection methodologies.


\end{abstract}

\section{Introduction}



The rapid advancement of LLMs has enabled AI to generate highly fluent, human-like text, raising concerns about detectability and prompting the development of various AI-text detectors \citep{gehrmann2019gltr, mitchell2023detectgpt, hu2023radar}.
However, the distinction between AI-generated and human-written text remains a gray area, particularly when human-authored content is refined using AI tools. \textit{If a human-written text is slightly polished by AI, should it still be classified as human-written, or does it become AI-generated?} Classifying such text as AI-generated can lead to false plagiarism accusations and unfair penalties, especially when detectors flag minimally polished content as AI-generated.\footnote{\href{https://www.usatoday.com/story/life/health-wellness/2025/01/22/college-students-ai-allegations-mental-health/77723194007/}{USAToday News on false AI allegations}}

Additionally, reports suggesting that a significant share of online content is AI-generated -- such as analyses indicating that over $40\%$ of Medium posts are likely AI-written -- often overlook the nuance of AI-polished text, which detection tools may misclassify.\footnote{\href{https://tinyurl.com/3jxktwkr}{Wired news on Medium}}$^{,}$\footnote{\href{https://tinyurl.com/594ydhaj}{NewsBytes report on Medium}}
These sweeping claims risk misrepresenting the actual extent of AI involvement, leading to misleading statistics and misplaced skepticism about human authorship. Motivated by these issues, our study systematically examines how AI-text detectors respond to AI-polished text and whether their classifications are both accurate and fair.

To investigate this issue, we introduce the \textbf{AI-Polished-Text Evaluation (APT-Eval)} dataset of size \textbf{14.7K}, which systematically examines how AI-text detectors respond to varying degrees of AI involvement in human writing. Our dataset is built from pre-existing human-written samples that are refined using different LLMs, such as GPT-4o \citep{openaiGPT4}, Llama3-70B \citep{dubey2024llama}, DeepSeek-V3 \citep{liu2024deepseek}, etc., applying degree and percentage based modifications. This allows us to assess how detectors respond to minor and major AI polishing. We analyze the classification accuracy, false positive rates, and domain-specific sensitivities of \textbf{12 state-of-the-art detectors}, spanning model-based, metric-based, and commercial systems.

Our findings reveal critical weaknesses in existing AI-text detection systems. AI-text detectors exhibit alarmingly high false positive rates, often flagging very minimally polished text as AI-generated. For instance, minimal polishing with GPT-4o can lead to detection rates ranging from $10\%$ to $75\%$, depending on the detector. Furthermore, detectors struggle to differentiate between minor and major AI refinements, suggesting that they may not be as reliable as previously assumed. 
We also uncover biases against smaller or older LLMs, where polishing done by less advanced models is more likely to be flagged than text refined by state-of-the-art LLMs. On average, $46\%$ of samples polished by LLaMA2-7B are classified as AI-generated, whereas this drops to $23\%$ for DeepSeek-V3-polished samples.
Furthermore, our study uncovers domain-specific inconsistencies in detection accuracy. Detectors show the greatest vulnerability for the `speech' domain, while exhibiting comparatively higher robustness to `paper-abstract' texts.

These findings raise concerns about the fairness and generalizability of current detection methods. By shedding light on these issues, our research provides valuable insights into the evolving challenges of AI-assisted writing and the limitations of current AI-text detection methodologies. Our code and dataset are publicly available: \url{https://github.com/ShoumikSaha/ai-polished-text.git}



\section{APT(AI-Polished-Text) Eval Dataset}
\subsection{Initial Dataset}
In this study, we begin with purely human-written texts (HWT) and refine them using various large language models (LLMs). Building on the work of \citet{zhang2024llm}, we utilize HWT samples from their `MixSet' dataset. These samples are carefully selected based on two key criteria: (1) they were created prior to the widespread adoption of LLMs, and (2) they span six distinct domains. For clarity, we refer to this baseline HWT dataset as the `No-Polish-HWT' set. This set comprises 300 samples, with 50 samples per domain (details in Table \ref{tab:dataset_hwt_basic}).

\subsection{Dataset Preparation}
As we generate the AI-polished versions of our No-Polish-HWT samples, we adjust the level of AI/LLM involvement. We employ two distinct polishing strategies: --
\begin{enumerate}
    \item \textbf{Degree-Based Polishing:} The LLM is prompted to refine the text in four varying degrees of modification: -- (1) extremely-minor, (2) minor, (3) slightly-major, and (4) major.

    \item \textbf{Percentage-Based Polishing:} The LLM is instructed to modify a fixed percentage $(p\%)$ of words in a given text. The percentage is systematically varied across the following values: $p\% = \{1, 5, 10, 20, 35, 50, 75 \}\%$.
\end{enumerate}

As a result, each HWT sample is transformed into 11 distinct AI-polished variants. For the LLM-polishing, we employ five different models: GPT-4o, Llama3.1-70B, Llama3-8B, Llama2-7B, and DeepSeek-V3. Each model is carefully prompted to generate the highest-quality output, preserving the original semantics of the text (details provided in Appendix \ref{app:llm_prompt}). Figure \ref{fig:extreme_minor_gpt_sample} illustrates a randomly selected sample from our dataset, which has been polished by GPT-4o with an extremely minor modification.


\begin{figure}[htbp]
    \centering
    \includegraphics[width=\linewidth]{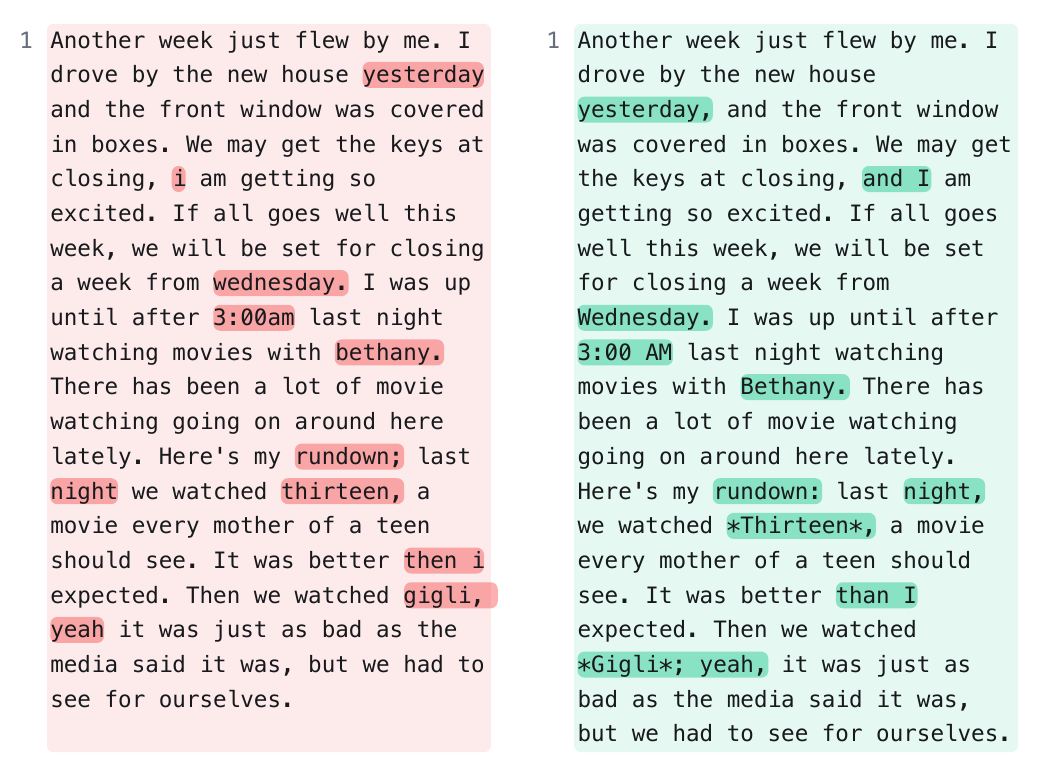}
    \caption{Random sample from our APT Eval dataset. Original HWT on left; Polished version on right.}
    \label{fig:extreme_minor_gpt_sample}
\end{figure}

\subsection{Dataset Analysis}
\begin{figure}[htbp]
    \centering
    \includegraphics[width=\linewidth]{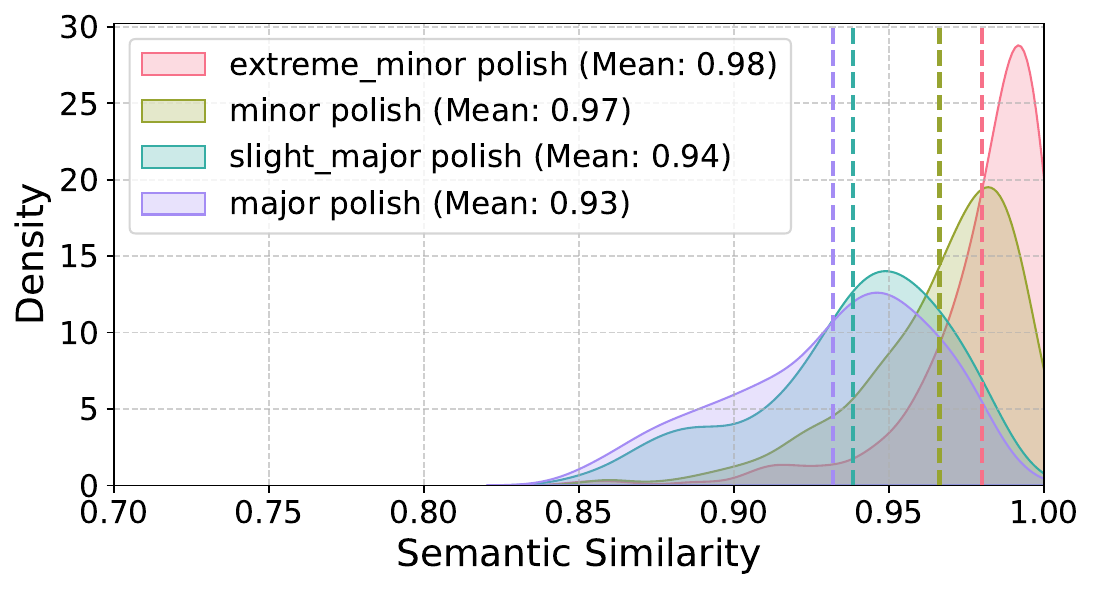}
    \caption{Distribution of Semantic Similarity for Degree-based AI-Polished Texts by GPT-4o.}
    \label{fig:sem_sim_gpt4}
\end{figure}

To assess the differences and deviations between pure HWT and AI-polished text, we employ three key metrics: Cosine semantic similarity, Jaccard distance, and Levenshtein distance. For semantic similarity, we compute the cosine similarity between the embeddings of the original and AI-polished texts (APT) using the BERT-base model. To ensure that the polished samples retain a strong resemblance to the original text, we filter out any samples with a semantic similarity below $0.85$. Figure \ref{fig:sem_sim_gpt4} shows the distribution for APTs (degree-based) with their mean value (see all metrics, and plots in Appendix \ref{app:data_analysis}).

\begin{table}[htbp]
\centering
\resizebox{\linewidth}{!}{%
\begin{tabular}{l|ccccc|c}
\multirow{2}{*}{\textbf{\begin{tabular}[c]{@{}l@{}}Polish \\ Type\end{tabular}}} & \multicolumn{5}{c|}{\textbf{Polisher LLM}} & \multirow{2}{*}{\textbf{Total}} \\ \cline{2-6}
 &
  \multicolumn{1}{l}{\textbf{GPT-4o}} &
  \textbf{\begin{tabular}[c]{@{}c@{}}Llama3.1\\ 70B\end{tabular}} &
  \textbf{\begin{tabular}[c]{@{}c@{}}Llama3\\ 8B\end{tabular}} &
  \textbf{\begin{tabular}[c]{@{}c@{}}Llama2\\ 7B\end{tabular}} &
  \textbf{\begin{tabular}[c]{@{}c@{}}DeepSeek\\ V3\end{tabular}} &
   \\ \hline
\textbf{\begin{tabular}[c]{@{}l@{}}no-polish /\\ pure HWT\end{tabular}}          & -      & -      & -      & -      & -      & 300                             \\ \hline
\textbf{\begin{tabular}[c]{@{}l@{}}Degree\\ based\end{tabular}}                  & 1152   & 1085   & 1125   & 744    & 1141   & 5547                            \\ \hline
\textbf{\begin{tabular}[c]{@{}l@{}}Percentage\\ based\end{tabular}}              & 2072   & 2048   & 1977   & 1282   & 1787   & 9166                            \\ \hline
\textbf{Total}                                                                   & 3224   & 3133   & 3102   & 2026   & 2928   & \textbf{14713}                 
\end{tabular}%
}
\caption{Our APT Eval Dataset}
\label{tab:apt_eval}
\end{table}

After filtering, the final APT-Eval dataset consists of \textbf{14.7K} samples, providing a robust benchmark for evaluating AI-text detection systems. Table \ref{tab:apt_eval} shows the total number of samples for each strategy and polisher (more details in table \ref{tab:apt_eval_details}).

\section{AI-text Detectors}
In this work, we evaluate a total of ten detectors from three different categories: 
\begin{enumerate}
    \item \textbf{Model-based:} RADAR \citep{hu2023radar}, RoBERTa-Base (ChatGPT) \citep{guo-etal-2023-hc3}, RoBERTa-Base (GPT2), and RoBERTa-Large (GPT2) \citep{gpt2OutputDataset}.
    
    \item \textbf{Metric-based:} GLTR \citep{gehrmann2019gltr}, DetectGPT \citep{mitchell2023detectgpt}, FastDetectGPT \citep{bao2023fast}, LLMDet \citep{wu2023llmdet}, Binoculars \citep{hans2024spotting}.
    \item \textbf{Commercial:} ZeroGPT \footnote{\url{https://www.zerogpt.com/}}, GPTZero \footnote{\url{https://gptzero.me/}}, Pangram\footnote{\url{https://www.pangram.com/}}.
\end{enumerate}

\subsection{Detectors' Threshold}
AI-text detectors generate a scalar score or prediction based on a given sequence of input tokens. To transform this score into a binary classification, an appropriate threshold must be determined. \citet{dugan2024raid} highlight that a naive threshold selection can lead to poor accuracy or a high false positive rate (FPR). Therefore, we optimize the threshold for each detector to achieve maximum accuracy in detecting HWT and AI-text.

We evaluate these detectors on $300$ samples of our `no-polish-HWT' (pure human-written) set and $300$ samples of pure AI-generated texts from the dataset of \citet{zhang2024llm}. Most detectors achieve $70\% - 88\%$ accuracy, with a false positive rate of $1\% - 8\%$. Table \ref{tab:detector_threshold} shows the detector-specific threshold with their accuracy and FPR.


\section{Key Findings}
We evaluate the detectors on our APT-Eval dataset from multiple perspectives to analyze their response to AI-polished text. Our key findings are as follows --

\subsection{Alarming false positive rate by AI-text detectors for minor polishing.} 
Though most detectors can achieve a low FPR on pure HWT (Table \ref{tab:detector_threshold}), most of them give a high FPR for any polishing, especially for extremely minor and minor polishing. 
For example, GLTR, with a $6.83\%$ FPR on pure HWT, classifies $40.87\%$ of extremely minor and $42.81\%$ of minor-polished GPT-4o texts as AI-text.
This trend extends to percentage-based polishing -- GLTR flagging $26.85\%$ of texts with only $1\%$ AI edits.
The issue persists across LLM polishers, with classification rates of: $34.11\%$ (DeepSeek-V3), 
$39.19\%$ (Llama3.1-70B), $44.86\%$ (Llama3-8B), and $52.31\%$ (Llama2-7B) for extremely minor AI-polishing. 
Figure \ref{fig:combined_result_main} visualizes these AI-detection rates, with further details in Appendix \ref{app:result_all_detectors}. 

For high‑stakes tasks like AI-text detection -- where even a $5\%$ FPR is very high -- every detector we evaluated flagged a higher number of minimally polished, human‑written texts than their unedited counterparts (Figure \ref{fig:combined_result_main}). Commercial systems are no exception: after applying only extremely minor edits with LLaMA-2-7B, the share of samples detected as AI-text jumped to \(32.31\%\) for ZeroGPT, \(42.56\%\) for Pangram, and \(64.71\%\) for GPTZero.

\begin{figure*}
    \centering
    \begin{minipage}{\textwidth} 
        \centering
        \begin{subfigure}{0.39\textwidth}
            \centering
            \includegraphics[width=\textwidth]{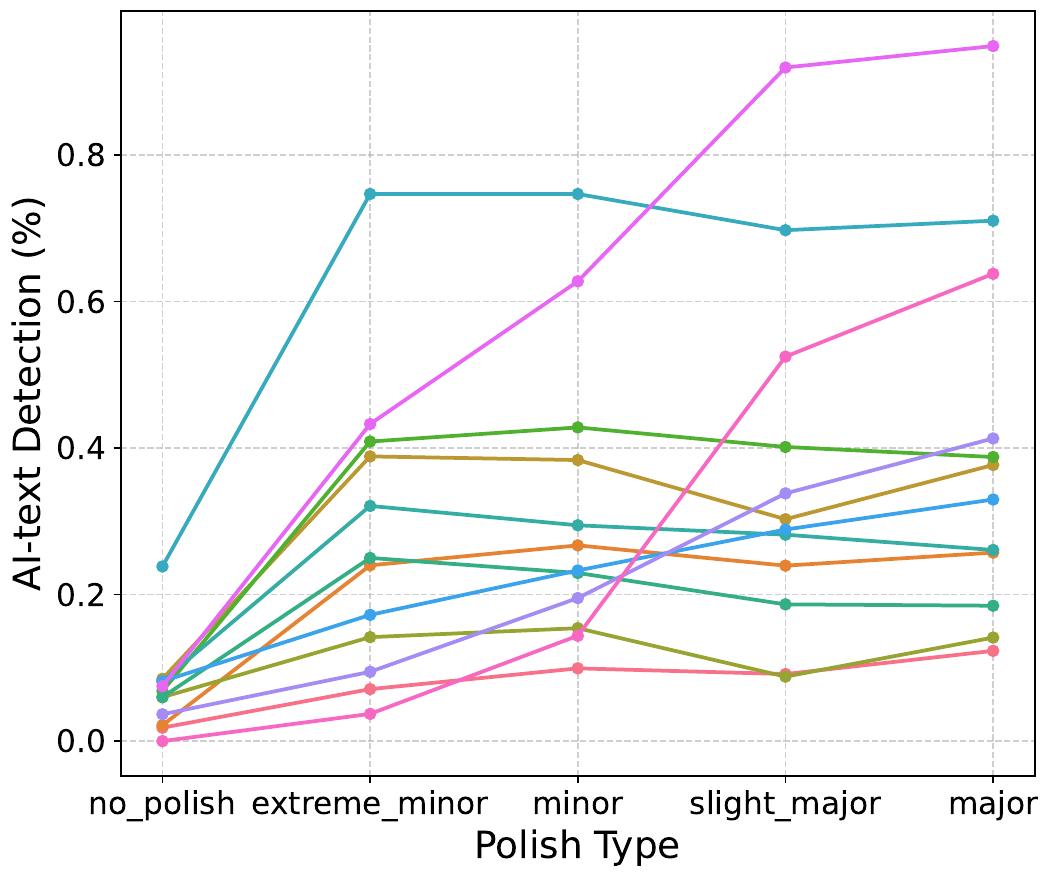}
            \caption{GPT-4o}
            \label{fig:gpt4_acc_overall}
        \end{subfigure}
        \hspace{2pt}
        \begin{subfigure}{0.55\textwidth}
            \centering
            \includegraphics[width=\textwidth]{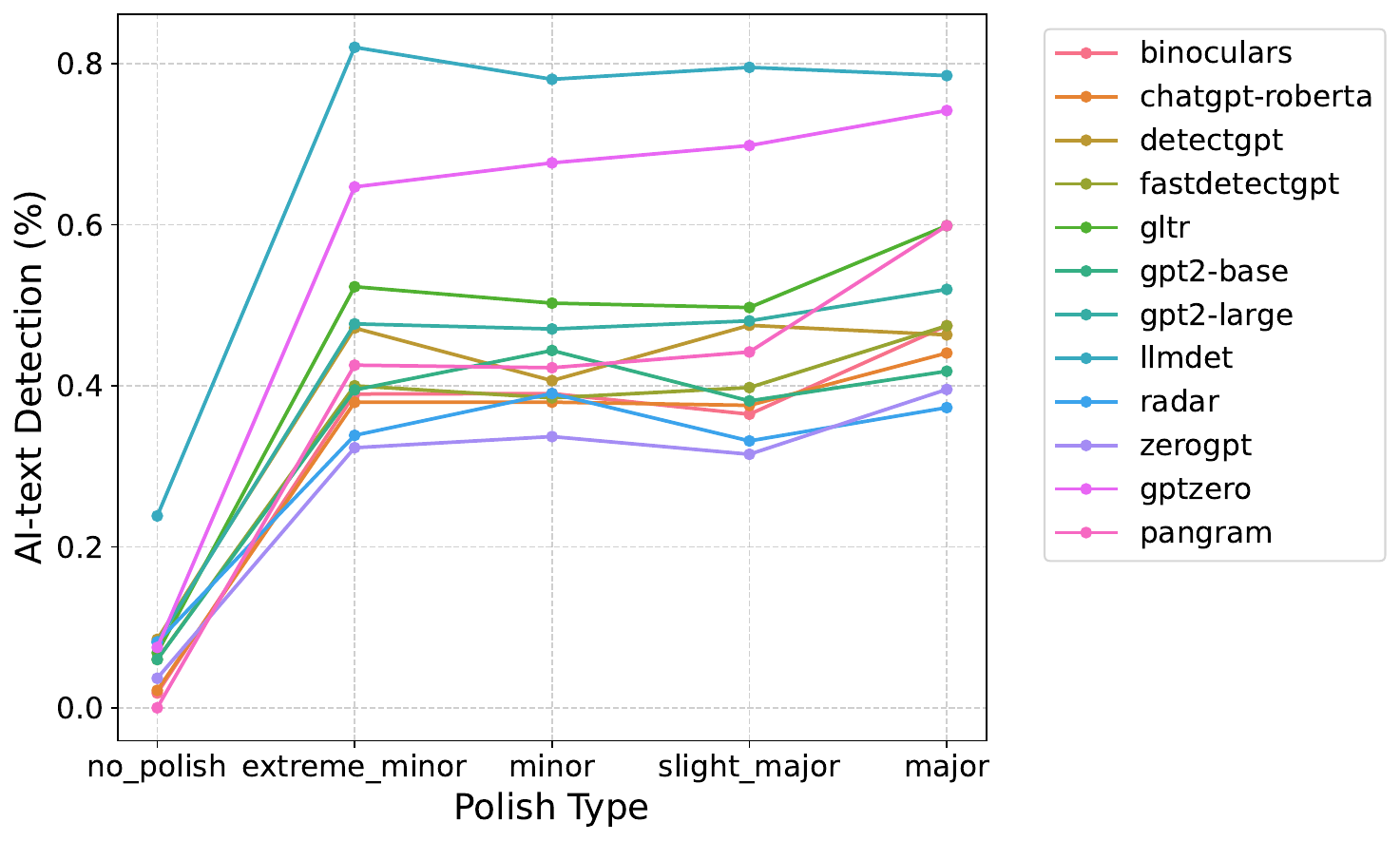}
            \caption{Llama2-7b}
            \label{fig:llama2_acc_overall}
        \end{subfigure}
    \end{minipage}
    
    \caption{AI-text detection rate for degree-based AI-polished-texts (APT) by all detectors.}
    \label{fig:combined_result_main}
\end{figure*}

\subsection{Most AI-text detectors fail to distinguish between minor and major polishing.} 

Most detectors not only flag a large portion of minor-polished texts but also struggle to differentiate between the degrees of AI involvement. For example, RoBERTa-large classifies $47.69\%$ of minor-polished texts as AI-generated, yet its rate for major-polished texts is only slightly higher at $51.98\%$.


Surprisingly, some detectors mark out fewer major-polished texts than extremely minor ones, revealing a lack of sensitivity to modification extent. As shown in Figure \ref{fig:combined_result_main}, detectors like DetectGPT, FastDetectGPT, GLTR, RoBERTa-base, RoBERTa-large, and LLMDet follow this trend. FastDetectGPT, for instance, detects $10.07\%$ of texts with $1\%$ AI edits as AI, but only $9.59\%$ for $75\%$ polishing (Figure \ref{fig:result_combined_prct}). 

\begin{figure}
    \centering
    \includegraphics[width=\linewidth]{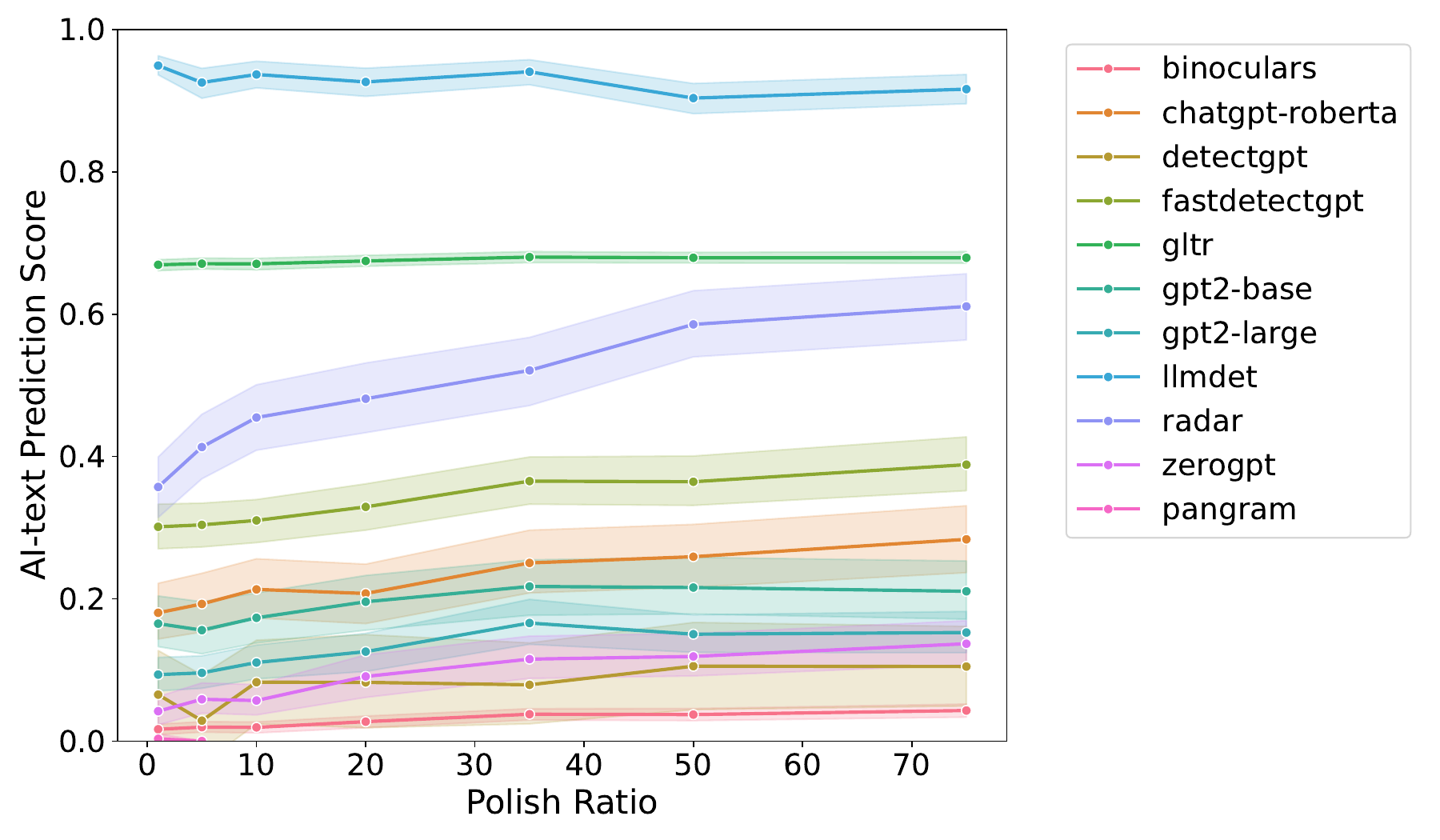}
    \caption{AI-text prediction score with 95\% confidence interval for percentage-based AI-polished-texts by Llama3-8B.}
    \label{fig:range_plot_prct_llama}
\end{figure}

To provide a richer understanding of the detectors' behavior beyond binary classification, we also analyze the raw probability score. Figure \ref{fig:range_plot_prct_llama} shows the mean prediction logits across different percents of AI-polishing by Llama3-8B, along with 95\% confidence intervals. 
Notably, most detectors exhibit minimal variation in logits across different polishing degrees -- revealing a key limitation: their inability to reliably distinguish between subtle and substantial levels of LLM-driven refinement.

\subsection{Most detectors penalize more if the polisher LLM is older or smaller.}
\begin{figure}
    \centering
    \includegraphics[width=\linewidth]{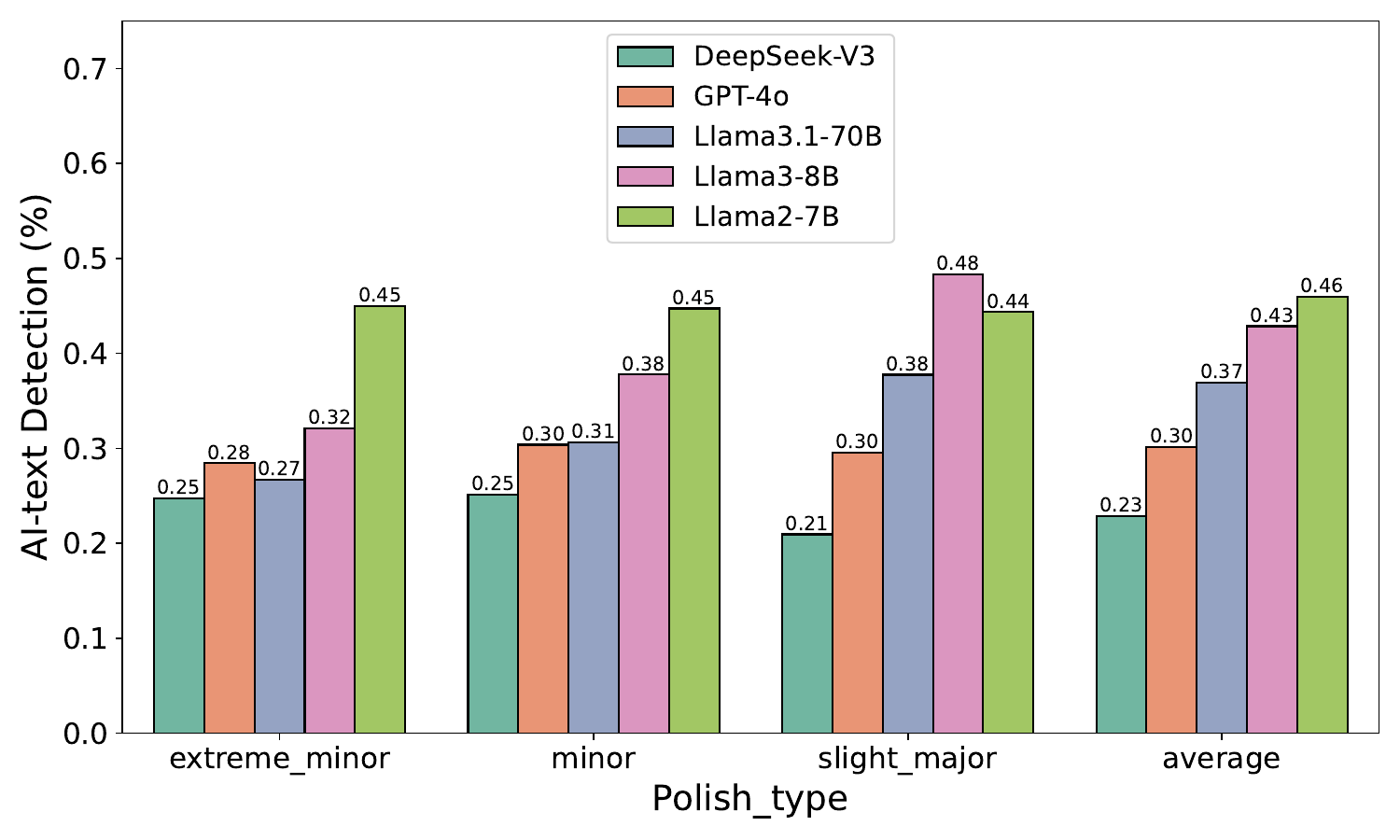}
    \caption{AI-text detection rate for degree-based AI-polished-texts from different polisher LLMs.}
    \label{fig:polisher_llm}
\end{figure}

We analyze whether AI-text detectors exhibit biases across different LLMs and find a higher AI-detection rate for smaller and older models (Figure \ref{fig:polisher_llm}). For extremely minor polishing, Llama-2 has a higher AI-detection rate of $45\%$, while DeepSeek-V3, GPT-4o, and Llama-3 models range from $25\%$ to $32\%$. The same trend was also found for our percentage-based polishing (Figure \ref{fig:polisher_vary_ratio}).
Among the polishers, the latest released LLM DeepSeek-V3 (Dec 2024) gets the lowest AI-detection rate on average as $23\%$.
The probable reason can be -- with time, newer LLMs have become increasingly adept at generating human-like text, making detection more challenging over time. 
However, such an imbalance can create unfair scenarios, where a student using Llama-2 is flagged for minor polishing while another using Llama-3.1 is found innocent. 

\subsection{Some domains are more sensitive than others.}
\begin{figure}
    \centering
    \includegraphics[width=0.85\linewidth]{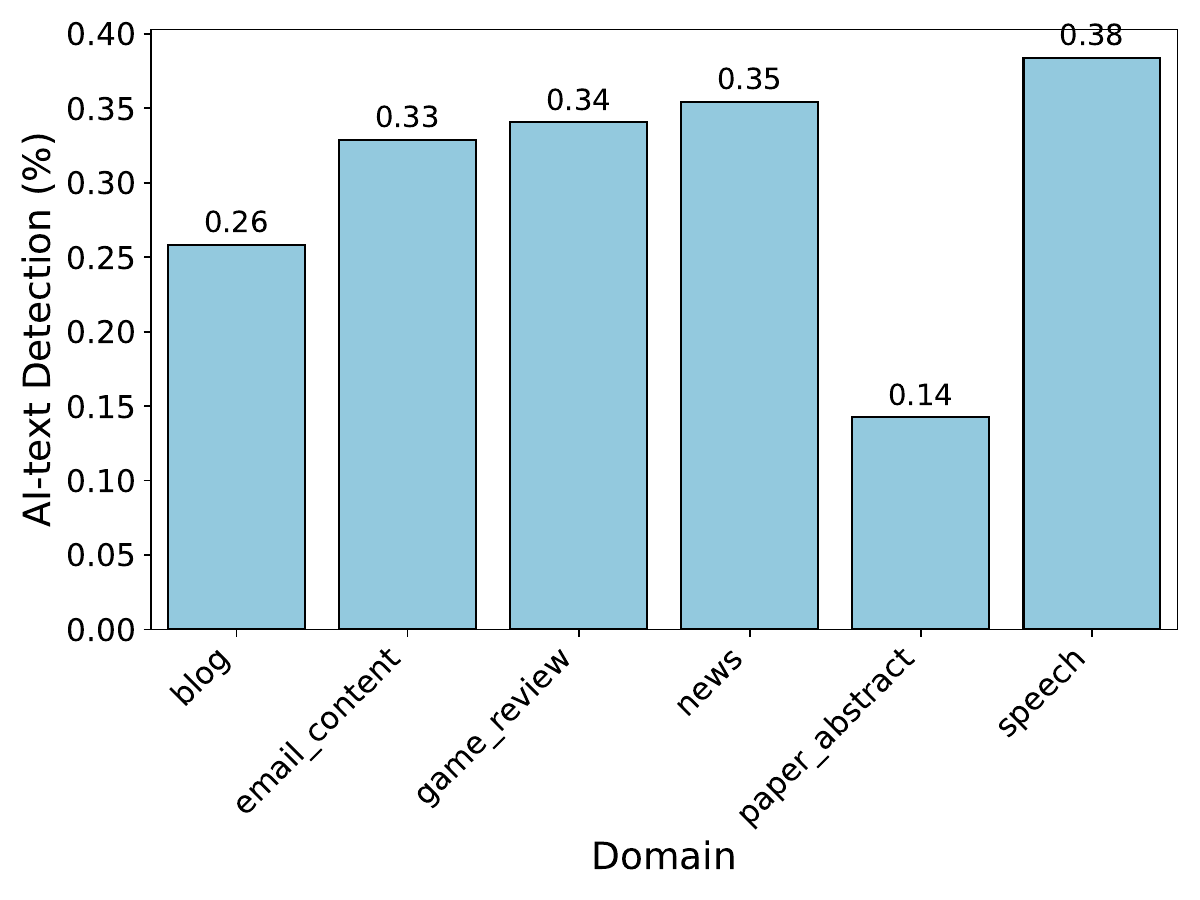}
    \caption{Average AI-detection rate for different domains (GPT-4o polishing)}
    \label{fig:domain_ext_minor_gpt}
\end{figure}

Since our HWT dataset spans six domains, we analyze detectors' AI-detection rate across them. Some domains are flagged more than others -- `speech' has the highest rate ($33\% - 56\%$ for extreme-minor polishing), while `paper\_abstract' has the lowest ($16\% - 31\%$). This trend is noticed at any level of polishing. Figure \ref{fig:domain_ext_minor_gpt} demonstrates the average AI-detection rate across different domains for GPT-4o polished texts. More detailed results are in Figure \ref{fig:domain_all_results} (Appendix \ref{app:domain_results}).

Interestingly, detection rates do not always correlate with polishing levels. As shown in figure \ref{fig:domain_change_plot}, for larger models like GPT-4o, DeepSeek-V3, and Llama3.1-70B, the AI-detection rate for `paper\_abstract' decreases as polishing increases -- likely because with more freedom in polishing, they tend to generate more-human-like texts. 


\section{Related Work}

Detecting AI-generated text is crucial as models become more human-like. Traditional methods use statistical metrics like perplexity and n-gram frequency \citep{gehrmann2019gltr, wu2023llmdet, hans2024spotting}, while others rely on machine learning classifiers like BERT and RoBERTa \citep{hu2023radar, guo-etal-2023-hc3, solaiman2019release}. However, these approaches mainly differentiate pure AI and human-written text.

Some prior studies have explored paraphrasing as a means to evade AI detectors \citep{sadasivan2023can, krishna2023paraphrasing}, but they do not specifically evaluate the unreliability of detection models in AI-polished text scenarios, which we address. Other works \citep{dugan2023real, zeng2024towards} focus on detecting the boundary between HWT and AI-generated text, treating the sentences as distinct entities. \citet{verma2023ghostbuster} explores how perturbations to AI-text can affect detection outcomes.
More recent research \citep{gao2024llm, yang2024chatgpt} has investigated LLM-assisted text polishing, but without considering varied degrees of AI involvement.   
We extend this by systematically analyzing AI-polished text across multiple levels, assessing detection limitations.


\section{Discussion}
The findings of this study reveal several critical limitations in current AI-text detectors, particularly in distinguishing between human-authored content that has been subtly refined by AI and fully machine-generated text. A key concern is the high false positive rate associated with minimally polished text. Many detectors classify such lightly edited content as AI-generated, which poses serious risks of unjust accusations of plagiarism or academic dishonesty.

To address this, we recommend moving beyond binary classification frameworks and adopting tiered or probabilistic labeling schemes that reflect varying degrees of AI involvement. A promising direction for future work is to train detectors not only on purely human-written and fully AI-generated texts but also on AI-polished samples. We hope our released APT-Eval dataset will serve as a valuable resource for developing and evaluating such models. Furthermore, rather than producing a definitive label, detectors should output prediction probabilities, enabling users to better interpret and trust the system’s verdict.

We also observe detector biases against older or smaller LLMs, and recommend training on a more diverse range of LLM outputs—including from newer, more human-like models—to improve fairness. Domain-specific biases can also be mitigated by fine-tuning detectors for particular genres or text types. Also, in the early stage of the pipeline, there can be another model to find the text-domain that will trigger a specific detector accordingly.

Lastly, we emphasize the importance of interpretability and human oversight in detection tools. Developing interpretable detectors that can highlight suspicious segments or stylistic anomalies will empower users to make informed decisions. In high-stakes scenarios, integrating human-in-the-loop review mechanisms can further enhance the reliability and fairness of the process. Ultimately, addressing these challenges requires a multi-faceted approach that balances technical sophistication with transparency, fairness, and adaptability.

\section{Conclusion}

Our study exposes key flaws in AI-text detectors when handling AI-polished text, showing high false positive rates and difficulty distinguishing minor from major AI refinements. Detectors also exhibit biases against older or smaller models, raising fairness concerns. We highlight the need for more nuanced detection methods and release our APT-Eval dataset to support further research.

\section*{Limitations}
While our study provides valuable insights into the challenges of AI-text detection for AI-polished texts, several limitations should be acknowledged. First, our dataset, APT-Eval, is built using a specific set of LLMs (GPT-4o, Llama3-70B, etc.), which may not fully represent the diversity of AI models available. Future research should explore a broader range of models to assess generalizability. Additionally, while our dataset spans six distinct domains of human-written text (HWT), incorporating more domains could provide a more comprehensive evaluation of AI-text detection across different writing contexts.

Second, our findings highlight biases in detection models, particularly against smaller or older LLMs, but further investigation is needed to understand the root causes of these biases. Moreover, while this study focuses on identifying limitations in current AI-text detection systems, the development of more nuanced, fine-grained detection frameworks remains an open challenge. Future work should explore adaptive AI-text detectors capable of distinguishing varying levels of AI involvement, ensuring both accuracy and fairness in AI-assisted writing evaluation.

\section*{Acknowledgments}
This project was supported in part by a grant from an NSF CAREER AWARD 1942230, ONR YIP award N00014-22-1-2271, ARO’s Early Career Program Award 310902-00001, Army Grant No. W911NF2120076, the NSF award CCF2212458, NSF Award No. 2229885 (NSF Institute for Trustworthy AI in Law and Society, TRAILS), a MURI grant 14262683, an award from meta 314593-00001 and an award from Capital One.

\bibliography{main}

\newpage
\onecolumn

\appendix

\section{Dataset} \label{app:dataset}
\subsection{Dataset Details} \label{app:data_details}

\begin{table*}[htbp]
\centering
\resizebox{\textwidth}{!}{%
\begin{tabular}{ll|ccccc|c}
\multicolumn{1}{l|}{\multirow{2}{*}{\textbf{Polish Type}}} & \multirow{2}{*}{\textbf{Polish Operation}} & \multicolumn{5}{c|}{\textbf{Polisher LLM}} & \multirow{2}{*}{\textbf{Total}} \\ \cline{3-7}
\multicolumn{1}{l|}{} &  & \textbf{GPT-4o} & \textbf{\begin{tabular}[c]{@{}c@{}}Llama3.1\\ 70B\end{tabular}} & \textbf{\begin{tabular}[c]{@{}c@{}}Llama3\\ 8B\end{tabular}} & \textbf{\begin{tabular}[c]{@{}c@{}}Llama2\\ 7B\end{tabular}} & \textbf{\begin{tabular}[c]{@{}c@{}}DeepSeek\\ V3\end{tabular}} &  \\ \hline
\multicolumn{1}{l|}{} & \textbf{\begin{tabular}[c]{@{}l@{}}no-polish \\ pure HWT\end{tabular}} & - & - & - & - & - & 300 \\ \hline
\multicolumn{1}{l|}{\multirow{4}{*}{\textbf{Degree-based}}} & \textbf{extreme-minor} & 297 & 297 & 293 & 196 & 300 & 1383 \\
\multicolumn{1}{l|}{} & \textbf{minor} & 293 & 294 & 286 & 188 & 297 & 1358 \\
\multicolumn{1}{l|}{} & \textbf{slight-major} & 285 & 282 & 280 & 182 & 293 & 1322 \\
\multicolumn{1}{l|}{} & \textbf{major} & 277 & 212 & 266 & 178 & 251 & 1184 \\ \hline
\multicolumn{1}{l|}{\multirow{7}{*}{\textbf{Percentage-based}}} & \textbf{1\%} & 299 & 299 & 294 & 193 & 265 & 1350 \\
\multicolumn{1}{l|}{} & \textbf{5\%} & 298 & 299 & 283 & 173 & 244 & 1297 \\
\multicolumn{1}{l|}{} & \textbf{10\%} & 297 & 298 & 285 & 176 & 225 & 1281 \\
\multicolumn{1}{l|}{} & \textbf{20\%} & 297 & 296 & 277 & 177 & 257 & 1304 \\
\multicolumn{1}{l|}{} & \textbf{35\%} & 295 & 294 & 287 & 199 & 247 & 1322 \\
\multicolumn{1}{l|}{} & \textbf{50\%} & 293 & 285 & 279 & 176 & 273 & 1306 \\
\multicolumn{1}{l|}{} & \textbf{75\%} & 293 & 277 & 272 & 188 & 276 & 1306 \\ \hline
\multicolumn{2}{c|}{\textbf{Total}} & 3224 & 3133 & 3102 & 2026 & 2928 & \textbf{14713}
\end{tabular}%
}
\caption{Details of our APT-Eval dataset}
\label{tab:apt_eval_details}
\end{table*}

\begin{table}[htbp]
    \centering
    \resizebox{0.7\linewidth}{!}{%
    \begin{tabular}{lll}
    \textbf{Domain}         & \textbf{Year} & \textbf{Source} \\ \hline
    Blog Content   & 2006 &  Blog \citep{schler2006effects}     \\
    Email Content  & 2015 &  Enron email dataset \citep{enronDataset}     \\
    News Content   & 2006 &  BBC news \citep{greene2006practical}      \\
    Game Reviews   & 2021 &  Steam reviews \citep{steamReviews2021} \\
    Paper Abstract & 2022 &  ArXiv-10  \citep{farhangi2022protoformer}    \\
    Speech Content & 2021 &  Ted Talk  \citep{tedTalks2021}    \\ \hline
    \end{tabular}
    }
    \caption{Details on our `no-polish-HWT' set of our Dataset.}
    \label{tab:dataset_hwt_basic}
\end{table}

\subsection{Prompts to Polisher LLM}\label{app:llm_prompt}

\begin{figure}[htbp]
    \centering
    \begin{mybox}
    \textbf{System Prompt: } You are a helpful chatbot who always responds with helpful information. You are asked to provide a polished version of the following text. Only generate the polished text.\\
    \textbf{User Prompt: } Polish the given original text below with $\texttt{\{polish\_type\}}$ polishing. The difference between original and polished text must be $\texttt{\{polish\_type\}}$. The semantic meaning of polished text must be the same as original text. The given original text: $\texttt{\{original\_text\}}$
    \end{mybox}
    \caption{Prompt for the Degree-based AI-polishing}
    \label{fig:prompt_gpt_degree_polish}
\end{figure}

\begin{figure}[htbp]
    \centering
    \begin{mybox}
    \textbf{System Prompt: } You are a helpful chatbot who always responds with helpful information. You are asked to provide a polished version of the following text. Only generate the polished text.\\
    \textbf{User Prompt: } Polish the given text below. The text has a total of $\texttt{\{text\_length\}}$ words. Make sure that you edit exactly $\texttt{\{polish\_word\_limit\}}$ words. Do not change or polish more than $\texttt{\{polish\_word\_limit\}}$ words. Also, make sure that the semantic meaning does not change with polishing. Only output the polished text, nothing else. The given text: $\texttt{\{original\_text\}}$
    \end{mybox}
    \caption{Prompt for the Percentage-based AI-polishing}
    \label{fig:prompt_gpt_percentage_polish}
\end{figure}

\newpage
\subsection{Dataset Analysis} \label{app:data_analysis}

\begin{figure*}[htbp]
    \centering
    \begin{minipage}{\textwidth} 
        \centering
        \begin{subfigure}{0.24\textwidth}
            \centering
            \includegraphics[width=\textwidth]{images/data_plots/multiple_semantic_similarity_polish_type_gpt.pdf}
            \caption{GPT-4o}
        \end{subfigure}
        \hfill
        \begin{subfigure}{0.24\textwidth}
            \centering
            \includegraphics[width=\textwidth]{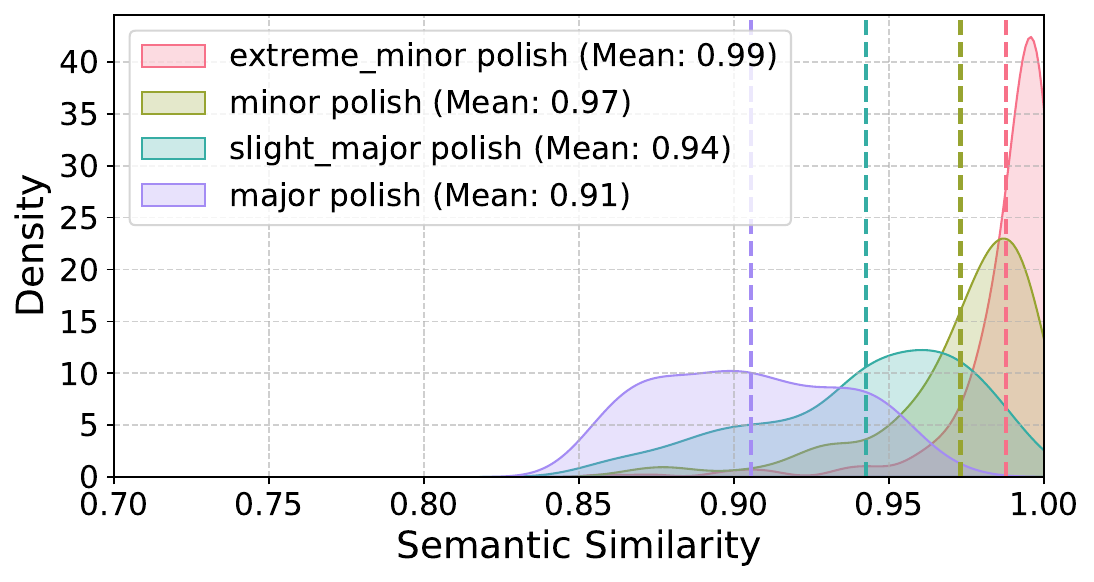}
            \caption{Llama3.1-70b}
        \end{subfigure}
        \hfill
        \begin{subfigure}{0.24\textwidth}
            \centering
            \includegraphics[width=\textwidth]{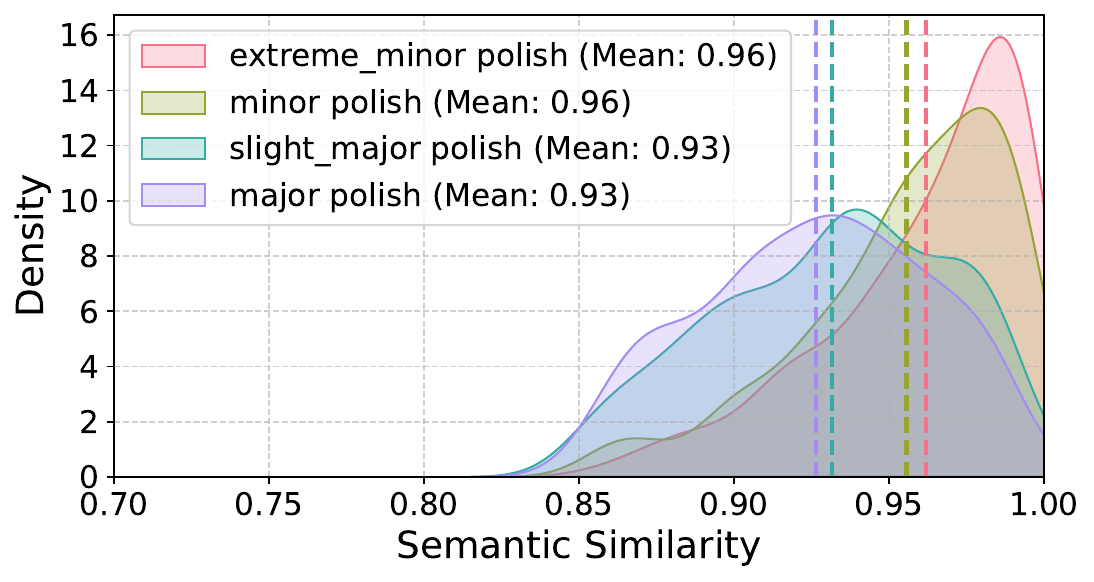}
            \caption{Llama3-8b}
        \end{subfigure}
        \hfill
        \begin{subfigure}{0.24\textwidth}
            \centering
            \includegraphics[width=\textwidth]{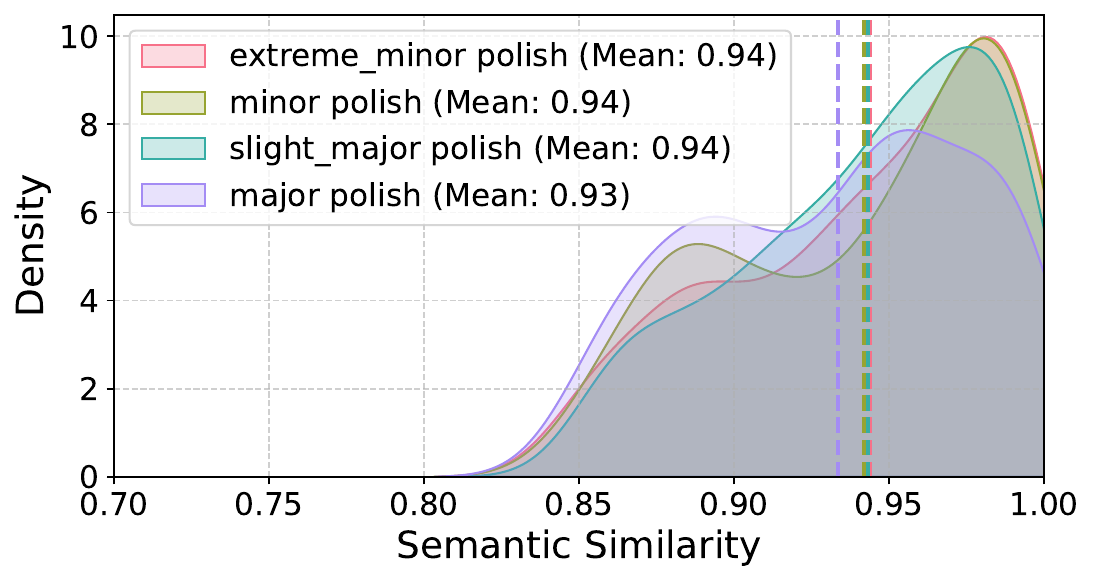}
            \caption{Llama2-7b}
        \end{subfigure}
        \\ 
        \vspace{0.2cm}
        \small \textbf{Distribution of Semantic Similarity}
    \end{minipage}
    
    \vspace{0.5cm} 
    \begin{minipage}{\textwidth}
    \centering
        \begin{subfigure}{0.24\textwidth}
            \centering
            \includegraphics[width=\textwidth]{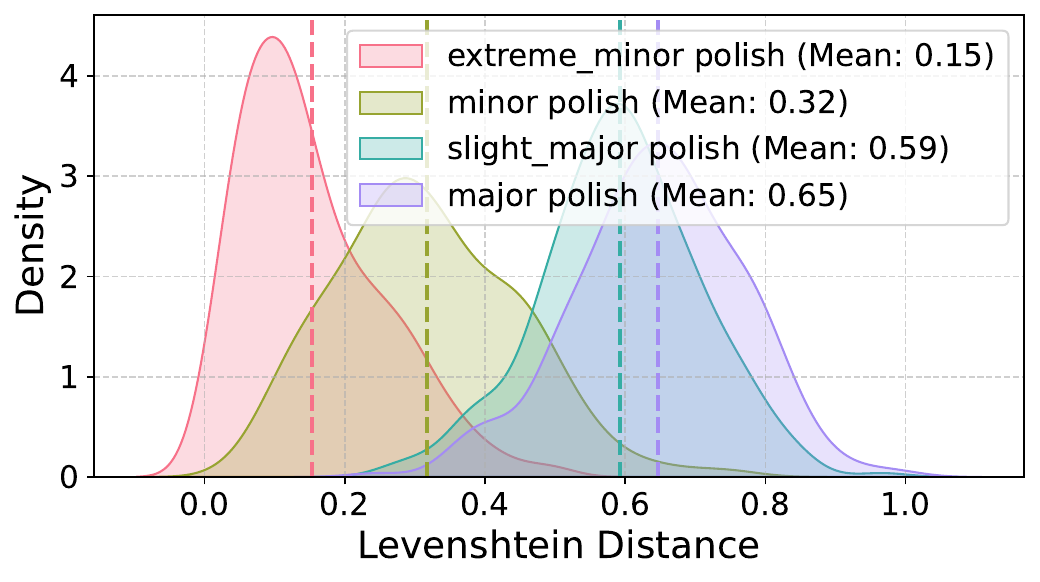}
            \caption{GPT-4o}
        \end{subfigure}
        \hfill
        \begin{subfigure}{0.24\textwidth}
            \centering
            \includegraphics[width=\textwidth]{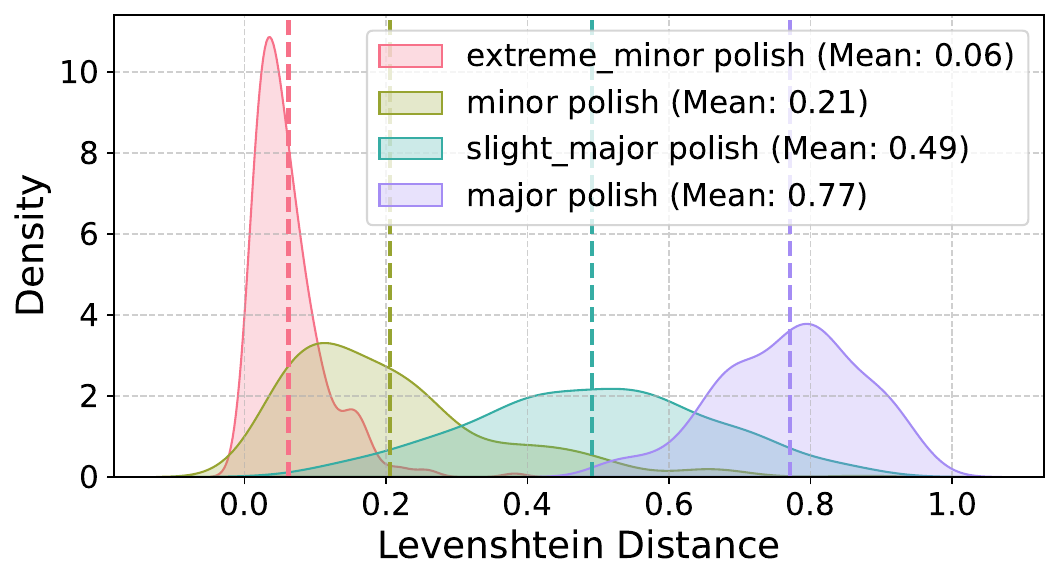}
            \caption{Llama3.1-70b}
        \end{subfigure}
        \hfill
        \begin{subfigure}{0.24\textwidth}
            \centering
            \includegraphics[width=\textwidth]{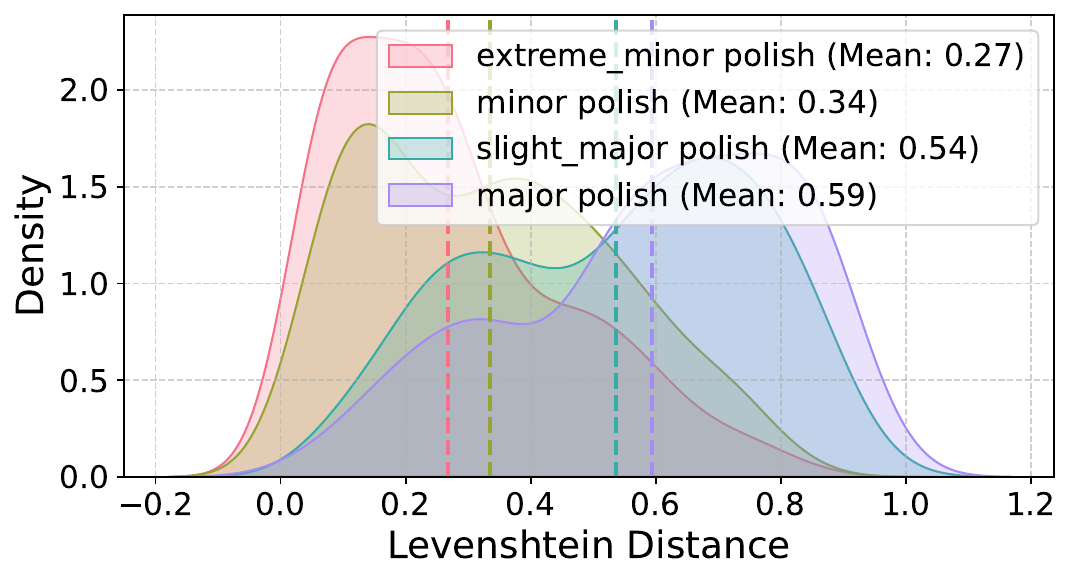}
            \caption{Llama3-8b}
        \end{subfigure}
        \hfill
        \begin{subfigure}{0.24\textwidth}
            \centering
            \includegraphics[width=\textwidth]{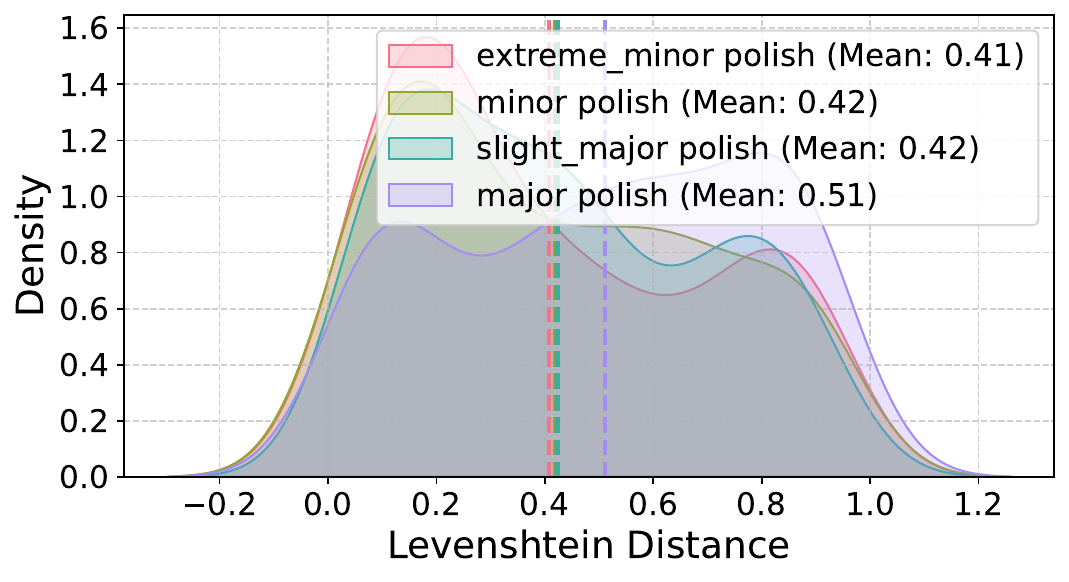}
            \caption{Llama2-7b}
        \end{subfigure}
        \\ 
        \vspace{0.2cm}
        \small \textbf{Distribution of Levenshtein Distance}
    \end{minipage}

    \vspace{0.5cm} 
    \begin{minipage}{\textwidth}
    \centering
        \begin{subfigure}{0.24\textwidth}
            \centering
            \includegraphics[width=\textwidth]{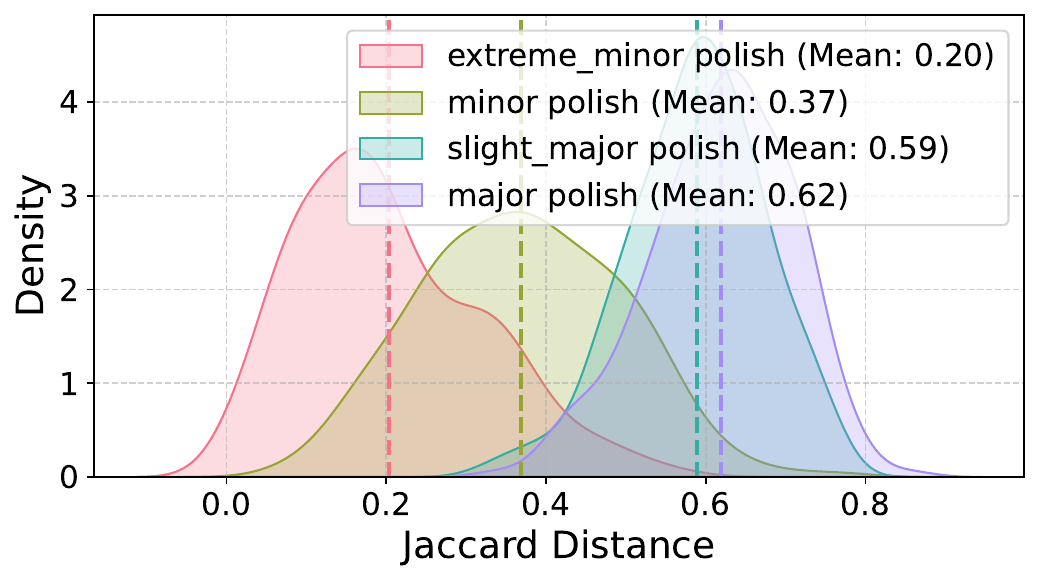}
            \caption{GPT-4o}
        \end{subfigure}
        \hfill
        \begin{subfigure}{0.24\textwidth}
            \centering
            \includegraphics[width=\textwidth]{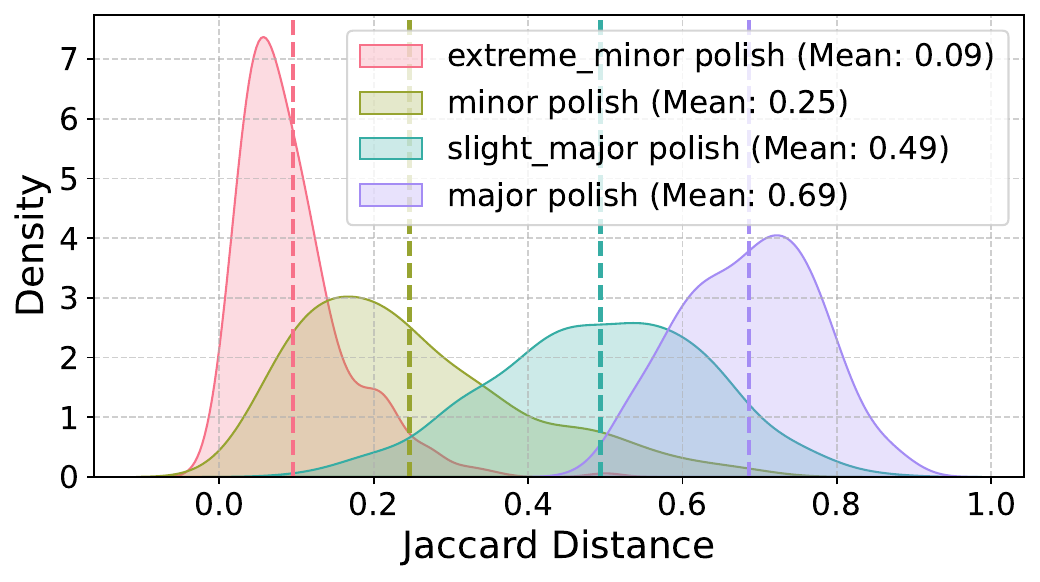}
            \caption{Llama3.1-70b}
        \end{subfigure}
        \hfill
        \begin{subfigure}{0.24\textwidth}
            \centering
            \includegraphics[width=\textwidth]{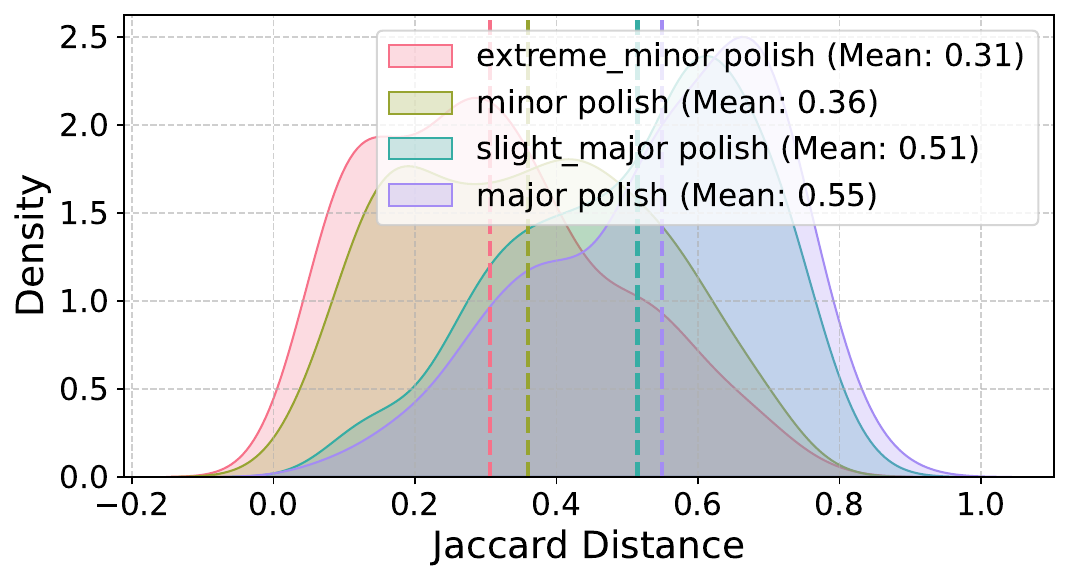}
            \caption{Llama3-8b}
        \end{subfigure}
        \hfill
        \begin{subfigure}{0.24\textwidth}
            \centering
            \includegraphics[width=\textwidth]{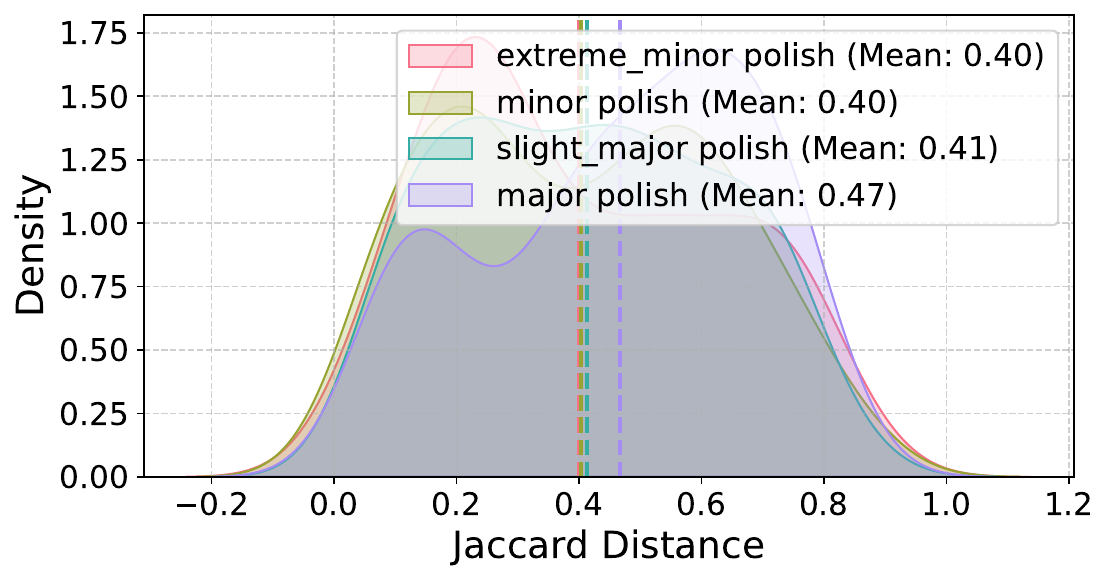}
            \caption{Llama2-7b}
        \end{subfigure}
        \\ 
        \vspace{0.2cm}
        \small \textbf{Distribution of Jaccard Distance}
    \end{minipage}
    
    \caption{Distribution of cosine semantic similarity, levenshtein distance, and jaccard distance for \textbf{degree-based} AI-polished texts by different polisher.}
    \label{fig:dataset_metric_degree}
\end{figure*}


\begin{figure*}[htbp]
    \centering
    \begin{minipage}{\textwidth} 
        \centering
        \begin{subfigure}{0.24\textwidth}
            \centering
            \includegraphics[width=\textwidth]{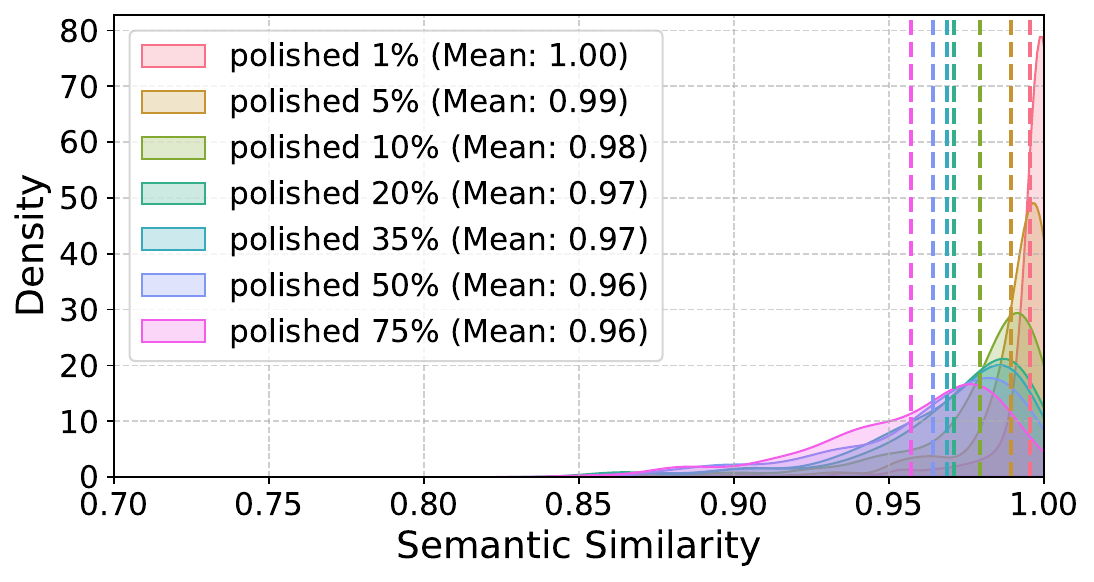}
            \caption{GPT-4o}
        \end{subfigure}
        \hfill
        \begin{subfigure}{0.24\textwidth}
            \centering
            \includegraphics[width=\textwidth]{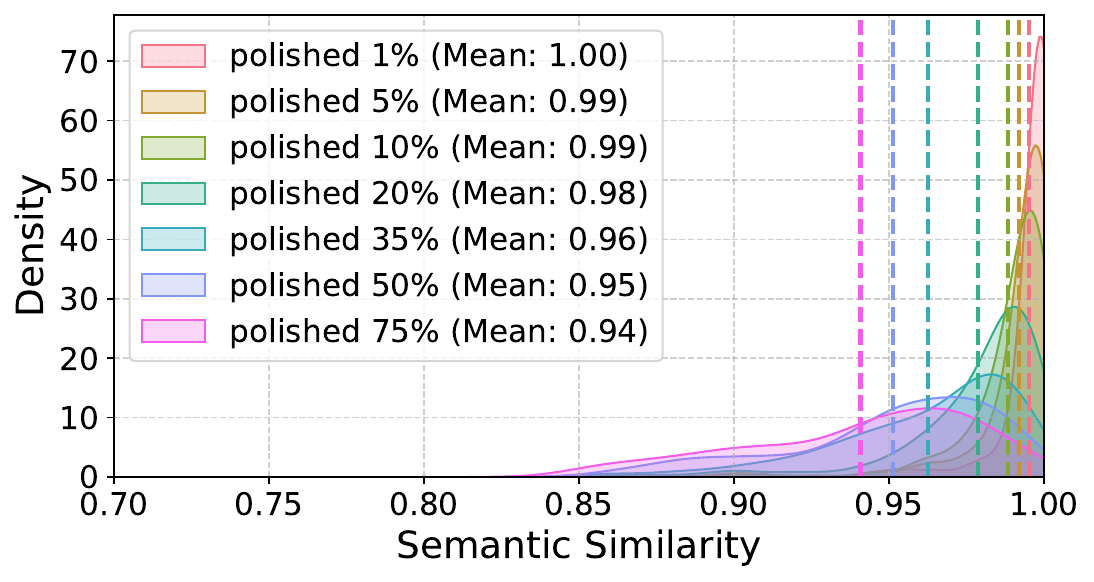}
            \caption{Llama3.1-70b}
        \end{subfigure}
        \hfill
        \begin{subfigure}{0.24\textwidth}
            \centering
            \includegraphics[width=\textwidth]{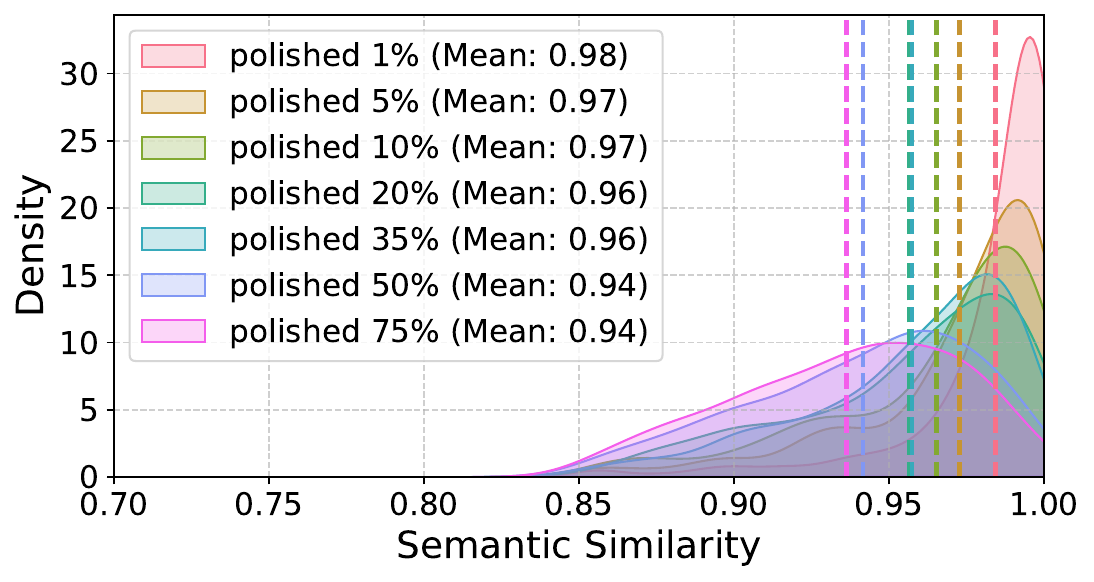}
            \caption{Llama3-8b}
        \end{subfigure}
        \hfill
        \begin{subfigure}{0.24\textwidth}
            \centering
            \includegraphics[width=\textwidth]{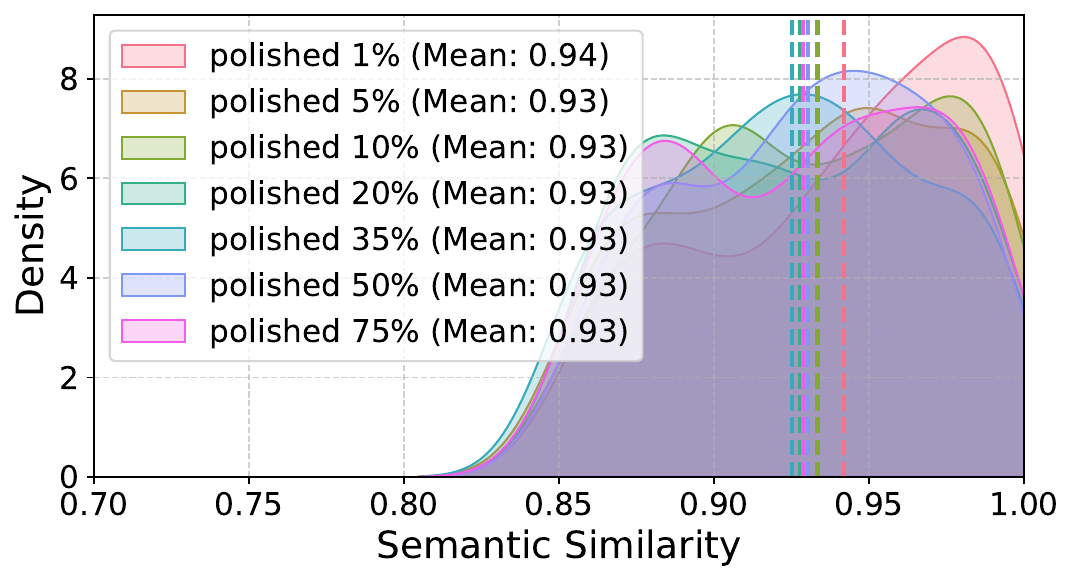}
            \caption{Llama2-7b}
        \end{subfigure}
        \\ 
        \vspace{0.2cm}
        \small \textbf{Distribution of Semantic Similarity}
    \end{minipage}
    
    \vspace{0.5cm} 
    \begin{minipage}{\textwidth}
    \centering
        \begin{subfigure}{0.24\textwidth}
            \centering
            \includegraphics[width=\textwidth]{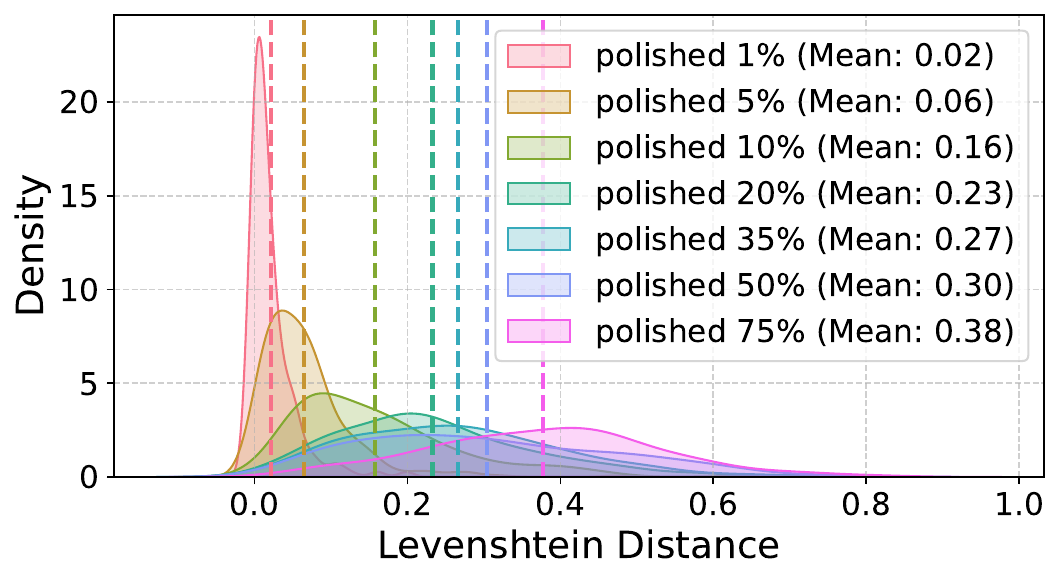}
            \caption{GPT-4o}
        \end{subfigure}
        \hfill
        \begin{subfigure}{0.24\textwidth}
            \centering
            \includegraphics[width=\textwidth]{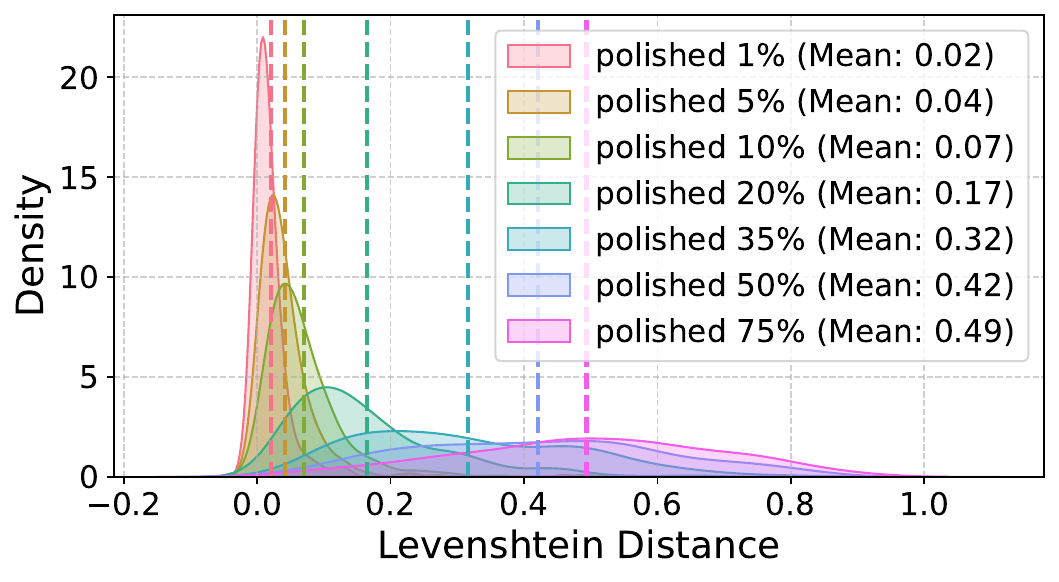}
            \caption{Llama3.1-70b}
        \end{subfigure}
        \hfill
        \begin{subfigure}{0.24\textwidth}
            \centering
            \includegraphics[width=\textwidth]{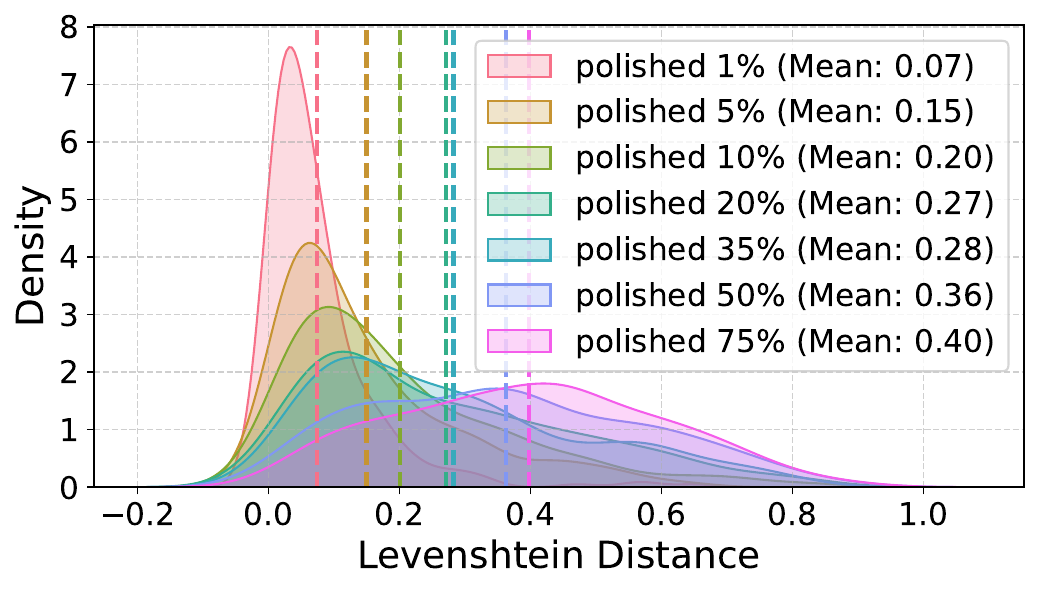}
            \caption{Llama3-8b}
        \end{subfigure}
        \hfill
        \begin{subfigure}{0.24\textwidth}
            \centering
            \includegraphics[width=\textwidth]{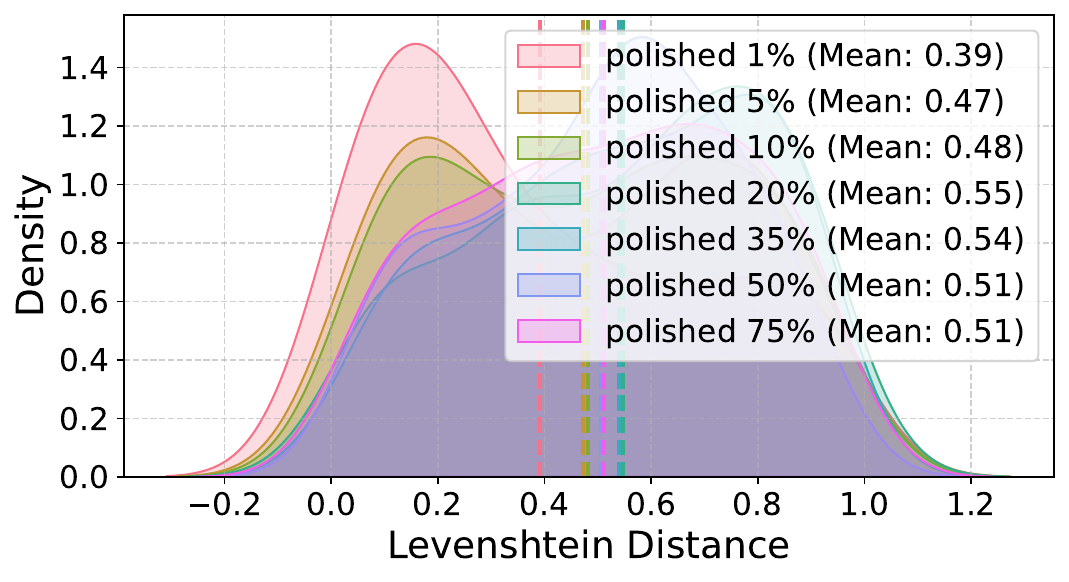}
            \caption{Llama2-7b}
        \end{subfigure}
        \\ 
        \vspace{0.2cm}
        \small \textbf{Distribution of Levenshtein Distance}
    \end{minipage}

    \vspace{0.5cm} 
    \begin{minipage}{\textwidth}
    \centering
        \begin{subfigure}{0.24\textwidth}
            \centering
            \includegraphics[width=\textwidth]{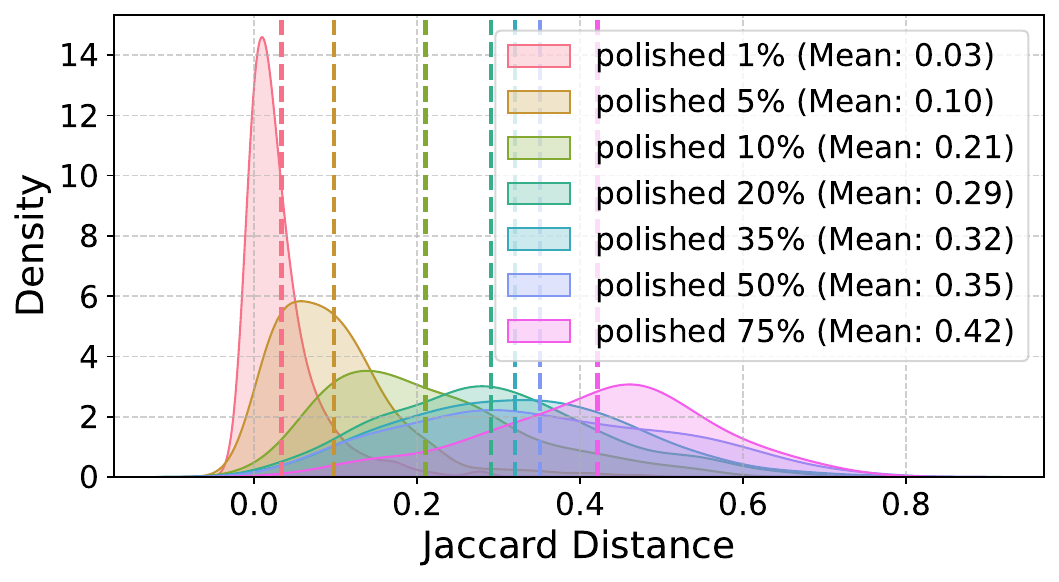}
            \caption{GPT-4o}
        \end{subfigure}
        \hfill
        \begin{subfigure}{0.24\textwidth}
            \centering
            \includegraphics[width=\textwidth]{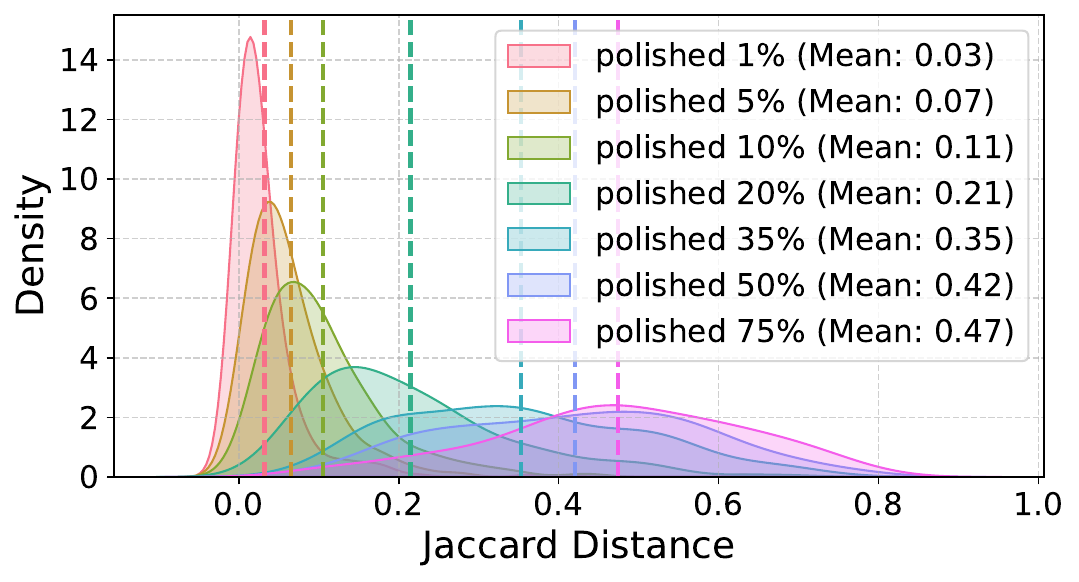}
            \caption{Llama3.1-70b}
        \end{subfigure}
        \hfill
        \begin{subfigure}{0.24\textwidth}
            \centering
            \includegraphics[width=\textwidth]{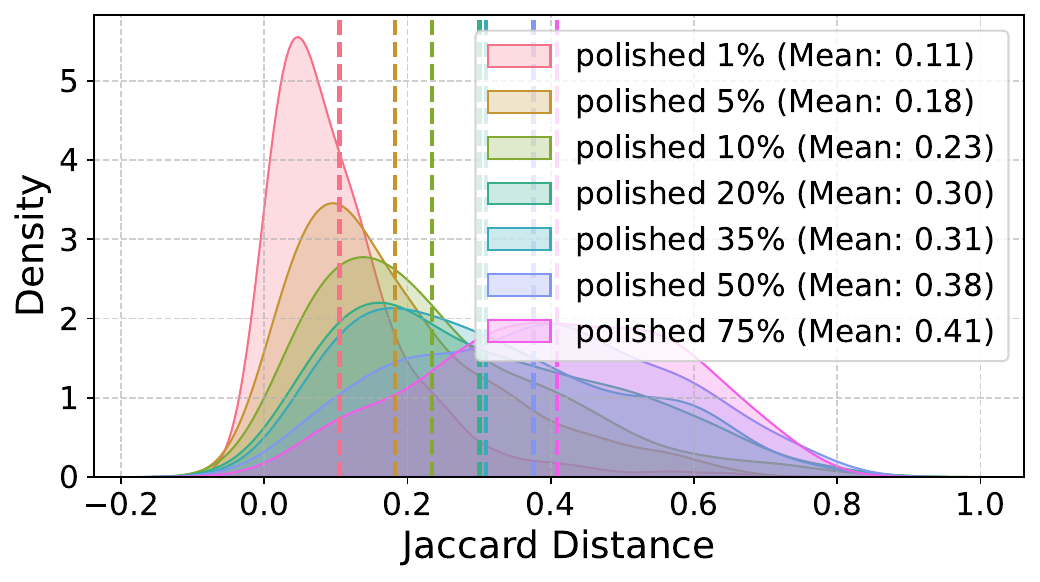}
            \caption{Llama3-8b}
        \end{subfigure}
        \hfill
        \begin{subfigure}{0.24\textwidth}
            \centering
            \includegraphics[width=\textwidth]{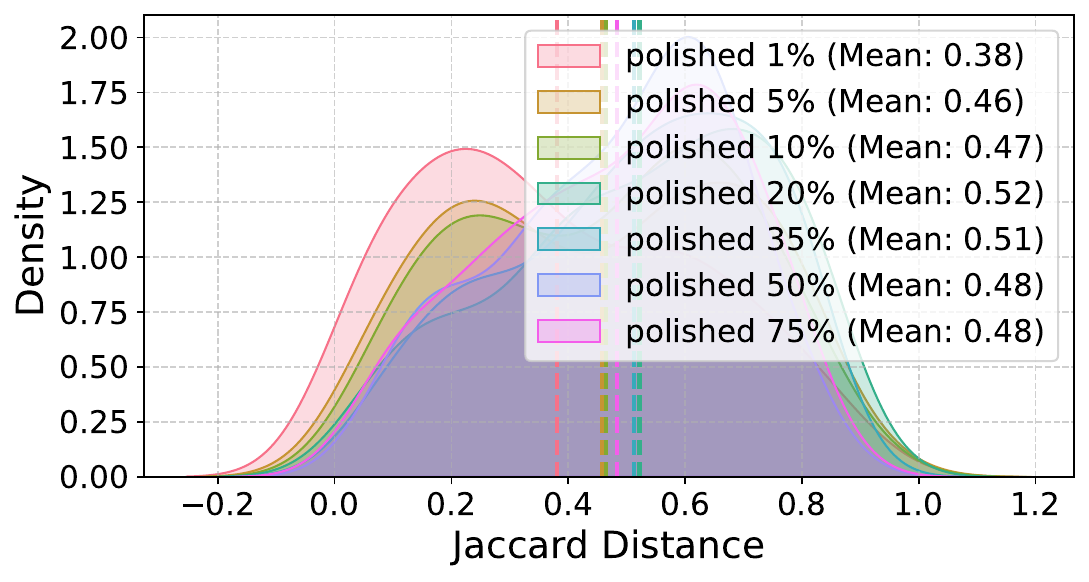}
            \caption{Llama2-7b}
        \end{subfigure}
        \\ 
        \vspace{0.2cm}
        \small \textbf{Distribution of Jaccard Distance}
    \end{minipage}
    
    \caption{Distribution of cosine semantic similarity, levenshtein distance, and jaccard distance for \textbf{percentage-based} AI-polished texts by different polisher.}
    \label{fig:dataset_metric_prct}
\end{figure*}

\begin{table*}[htbp]
\centering
\resizebox{\textwidth}{!}{%
\begin{tabular}{l|lll|lll}
\multirow{2}{*}{\textbf{\begin{tabular}[c]{@{}l@{}}Polish\\ Type\end{tabular}}} & \multicolumn{3}{c|}{\textbf{GPT-4o}} & \multicolumn{3}{c}{\textbf{Llama3.1-70B}} \\ \cline{2-7} 
 &
  \textbf{\begin{tabular}[c]{@{}l@{}}Mean Semantic \\ Similarity\end{tabular}} &
  \textbf{\begin{tabular}[c]{@{}l@{}}Mean Levensthein \\ Distance\end{tabular}} &
  \textbf{\begin{tabular}[c]{@{}l@{}}Mean Jaccard \\ Distance\end{tabular}} &
  \textbf{\begin{tabular}[c]{@{}l@{}}Mean Semantic \\ Similarity\end{tabular}} &
  \textbf{\begin{tabular}[c]{@{}l@{}}Mean Levensthein \\ Distance\end{tabular}} &
  \textbf{\begin{tabular}[c]{@{}l@{}}Mean Jaccard \\ Distance\end{tabular}} \\ \hline
\textbf{extreme-minor}                                                          & 0.98       & 0.15       & 0.20       & 0.99         & 0.06         & 0.09        \\
\textbf{minor}                                                                  & 0.97       & 0.35       & 0.37       & 0.97         & 0.21         & 0.25        \\
\textbf{slight-major}                                                           & 0.94       & 0.59       & 0.59       & 0.94         & 0.49         & 0.49        \\
\textbf{major}                                                                  & 0.93       & 0.65       & 0.62       & 0.91         & 0.77         & 0.69       
\end{tabular}%
}
\caption{Similarity and Distance between original HWT and AI-polished texts (degree-based)}
\label{tab:dataset_metric_degree}
\end{table*}

\begin{table*}[htbp]
\centering
\resizebox{\textwidth}{!}{%
    \begin{tabular}{l|lll|lll}
    \multirow{2}{*}{\textbf{Polish \%}} & \multicolumn{3}{c|}{\textbf{GPT-4o}} & \multicolumn{3}{c}{\textbf{Llama3.1-70B}} \\ \cline{2-7} 
     &
      \textbf{\begin{tabular}[c]{@{}l@{}}Mean Semantic \\ Similarity\end{tabular}} &
      \textbf{\begin{tabular}[c]{@{}l@{}}Mean Levensthein \\ Distance\end{tabular}} &
      \textbf{\begin{tabular}[c]{@{}l@{}}Mean Jaccard \\ Distance\end{tabular}} &
      \textbf{\begin{tabular}[c]{@{}l@{}}Mean Semantic \\ Similarity\end{tabular}} &
      \textbf{\begin{tabular}[c]{@{}l@{}}Mean Levensthein \\ Distance\end{tabular}} &
      \textbf{\begin{tabular}[c]{@{}l@{}}Mean Jaccard \\ Distance\end{tabular}} \\ \hline
    1                                   & 1.00       & 0.02       & 0.03       & 1.00         & 0.02         & 0.03        \\
    5                                   & 0.99       & 0.06       & 0.10       & 0.99         & 0.04         & 0.07        \\
    10                                  & 0.98       & 0.16       & 0.21       & 0.99         & 0.07         & 0.11        \\
    20                                  & 0.97       & 0.23       & 0.29       & 0.98         & 0.17         & 0.21        \\
    35                                  & 0.97       & 0.27       & 0.32       & 0.96         & 0.32         & 0.35        \\
    50                                  & 0.96       & 0.30       & 0.35       & 0.95         & 0.42         & 0.42        \\
    75                                  & 0.96       & 0.38       & 0.42       & 0.94         & 0.49         & 0.47       
    \end{tabular}%
}
\caption{Similarity and Distance between original HWT and AI-polished texts (percentage-based)}
\label{tab:dataset_metric_prct}
\end{table*}

\clearpage
\newpage

\subsection{Difference Between APT-Eval and MixSet Dataset}
While our APT-Eval dataset builds upon the human-written portion of the MixSet dataset, it is fundamentally different in its objectives and design. MixSet includes AI-polished text using token- and sentence-level paraphrasing, but it does not control or quantify the degree of AI involvement in the polishing process. In contrast, our APT-Eval dataset systematically explores varying levels of AI assistance -- both in terms of degree-based (e.g., extremely minor to major edits) and percentage-based modifications. Our goal is to introduce a fine-grained continuum of AI involvement, allowing us to rigorously evaluate how current detectors respond to subtle versus substantial AI polish, rather than just simple polish.

\subsection{Samples of our APT-Eval Dataset}
\begin{figure*}[htbp]
    \centering
    \begin{minipage}{\textwidth} 
        \centering
        \begin{subfigure}{0.7\textwidth}
            \centering
            \includegraphics[width=\textwidth]{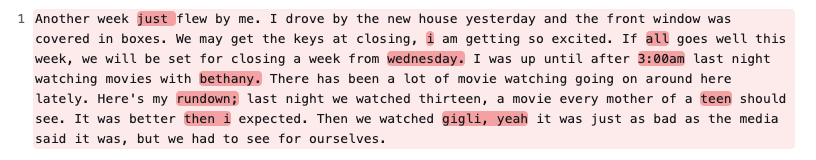}
            \caption{Original HWT}
        \end{subfigure}
    \end{minipage}

    \vspace{0.5cm} 
    \begin{minipage}{\textwidth} 
        \centering
        \begin{subfigure}{0.7\textwidth}
            \centering
            \includegraphics[width=\textwidth]{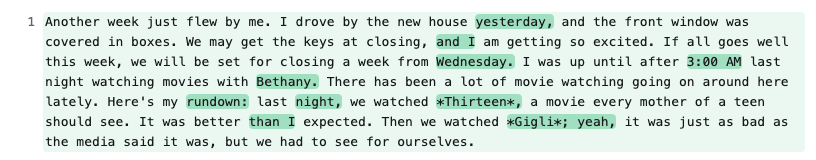}
            \caption{GPT-4o Polished}
        \end{subfigure}
    \end{minipage}

    \vspace{0.5cm} 
    \begin{minipage}{\textwidth} 
        \centering
        \begin{subfigure}{0.7\textwidth}
            \centering
            \includegraphics[width=\textwidth]{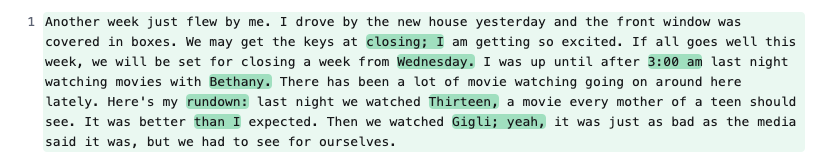}
            \caption{Llama3.1-70b Polished}
        \end{subfigure}
    \end{minipage}

    \vspace{0.5cm} 
    \begin{minipage}{\textwidth} 
        \centering
        \begin{subfigure}{0.7\textwidth}
            \centering
            \includegraphics[width=\textwidth]{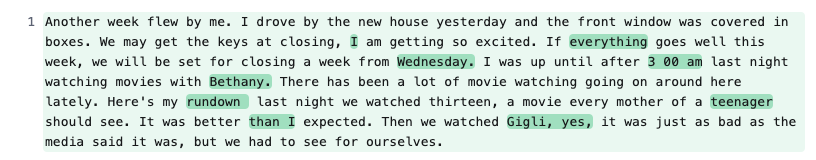}
            \caption{Llama3-8b Polished}
        \end{subfigure}
    \end{minipage}

    \vspace{0.5cm} 
    \begin{minipage}{\textwidth} 
        \centering
        \begin{subfigure}{0.7\textwidth}
            \centering
            \includegraphics[width=\textwidth]{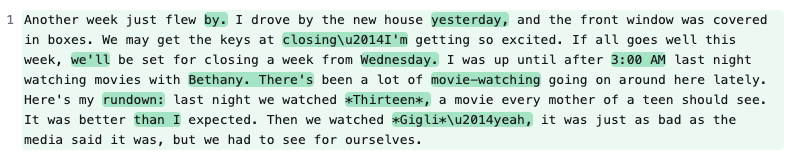}
            \caption{DeepSeek-V3 Polished}
        \end{subfigure}
    \end{minipage}

    \caption{Example sample from our APT-Eval dataset, where the original HWT is polished (extreme-minor) by different polisher LLMs.}
    \label{fig:sample_exmpl_1}
\end{figure*}

\begin{figure*}[htbp]
    \centering
    \begin{minipage}{\textwidth} 
        \centering
        \begin{subfigure}{0.7\textwidth}
            \centering
            \includegraphics[width=\textwidth]{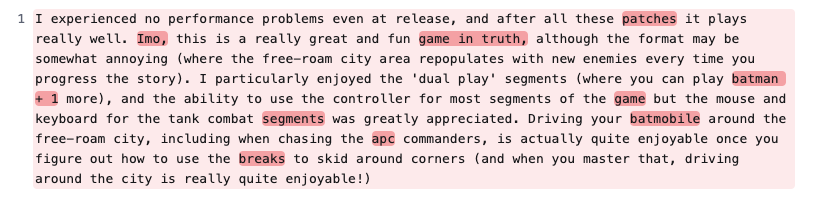}
            \caption{Original HWT}
        \end{subfigure}
    \end{minipage}

    \vspace{0.5cm} 
    \begin{minipage}{\textwidth} 
        \centering
        \begin{subfigure}{0.7\textwidth}
            \centering
            \includegraphics[width=\textwidth]{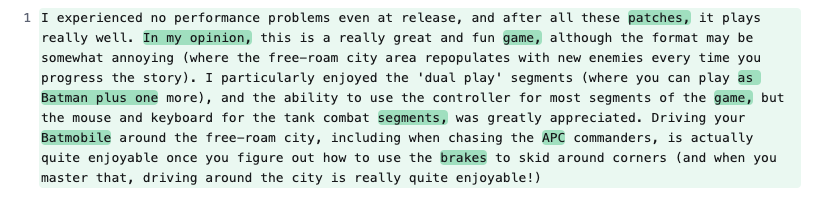}
            \caption{Extreme-minor Polishing}
        \end{subfigure}
    \end{minipage}

    \vspace{0.5cm} 
    \begin{minipage}{\textwidth} 
        \centering
        \begin{subfigure}{0.7\textwidth}
            \centering
            \includegraphics[width=\textwidth]{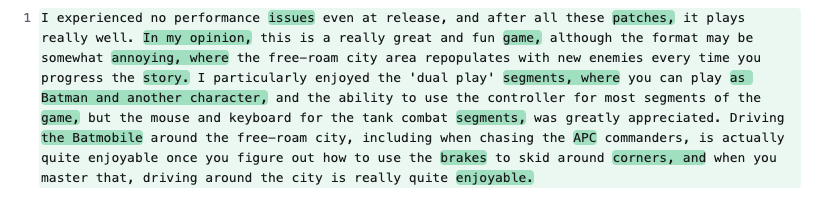}
            \caption{Minor Polishing}
        \end{subfigure}
    \end{minipage}

    \vspace{0.5cm} 
    \begin{minipage}{\textwidth} 
        \centering
        \begin{subfigure}{0.7\textwidth}
            \centering
            \includegraphics[width=\textwidth]{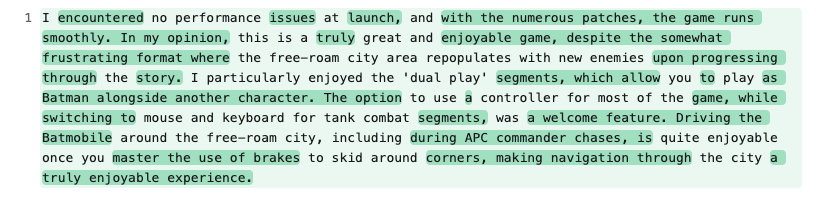}
            \caption{Slight-major Polishing}
        \end{subfigure}
    \end{minipage}

    \vspace{0.5cm} 
    \begin{minipage}{\textwidth} 
        \centering
        \begin{subfigure}{0.7\textwidth}
            \centering
            \includegraphics[width=\textwidth]{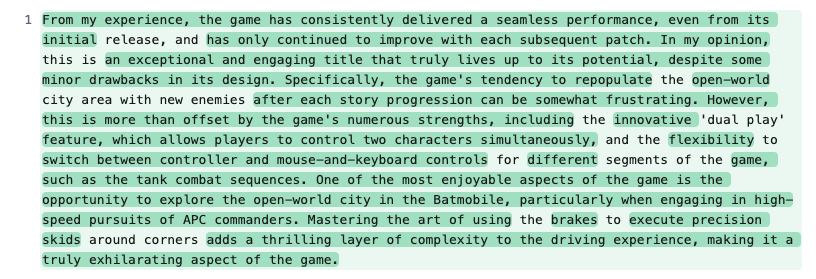}
            \caption{Major Polishing}
        \end{subfigure}
    \end{minipage}

    \caption{Example sample from our APT-Eval dataset, where the original HWT is polished with different degrees by Llama3.1-70B model.}
    \label{fig:sample_exmpl_110}
\end{figure*}

\clearpage
\newpage
\section{AI-text Detectors} \label{app:detector}
We developed our code-base on the framework of RAID \cite{dugan2024raid}, and kept the hyperparameters for all detectors the same as RAID for a fair evaluation. 

Additionally, we identify the threshold corresponding to a $5\%$ false positive rate (FPR) -- or the lowest possible FPR if it exceeds $5\%$. We notice that -- for most detectors, the thresholds for `best accuracy' and `5\% FPR' do not vary much. Moreover, since our primary focus is on misclassification rates for both minor-polished and major-polished texts, optimizing the threshold for overall accuracy is more appropriate than minimizing FPR alone.


Table \ref{tab:detector_threshold} shows the threshold that we found by optimizing the accuracy, and used for the evaluation of our APT-Eval dataset. 

\begin{table*}[h!]
\centering
\resizebox{0.7\textwidth}{!}{%
\begin{tabular}{lllll}
                                       & \textbf{Detector}             & \textbf{Threshold} & \textbf{Accuracy} & \textbf{FPR}    \\ \hline
\multirow{4}{*}{\textbf{Model-Based}}  & \textbf{RADAR}                & 0.8989             & 0.8017            & 0.082           \\
                                       & \textbf{RoBERTa (ChatGPT)}    & 0.333              & 0.8617            & 0.0217          \\
                                       & \textbf{RoBERTa-Base (GPT2)}  & 0.091              & 0.7917            & 0.06            \\
                                       & \textbf{RoBERTa-Large (GPT2)} & 0.0408             & 0.8               & 0.0817          \\ \hline
\multirow{5}{*}{\textbf{Metric-Based}} & \textbf{GLTR}                 & 0.7038             & 0.845             & 0.0683          \\
                                       & \textbf{DetectGPT}            & 0.355              & 0.725             & 0.085           \\
                                       & \textbf{Fast-DetectGPT}       & 0.778              & 0.8317            & 0.06            \\
                                       & \textbf{LLMDet}               & 0.9798             & 0.605             & 0.2383          \\
                                       & \textbf{Binoculars}           & 0.1075             & \textbf{0.88}     & \textit{0.018}  \\ \hline
\multirow{3}{*}{\textbf{Commercial}}   & \textbf{ZeroGPT}              & 0.2525             & 0.8067            & 0.0367          \\
                                       & \textbf{GPTZero}              & 0.03               & 0.862             & 0.075           \\ 
                                       & \textbf{Pangram}              & 0.01               & \textit{0.875}             & \textbf{0.00}   \\ \hline
\end{tabular}%
}
\caption{Detector-based Threshold, Accuracy, and False Positive Rate. The best performance is in bold, and the second best is in italics.}
\label{tab:detector_threshold}
\end{table*}


For computational resources, we employed:
\begin{itemize}
    \item One NVIDIA RTX A5000 GPU for running model-based and metric-based detectors.
    \item One NVIDIA RTX A6000 GPU for the Binoculars detector.
    \item API subscriptions for ZeroGPT, GPTZero and Pangram
\end{itemize}

\newpage
\section{Results and Findings} \label{app:result}
\subsection{Results for All Detectors} \label{app:result_all_detectors}
\begin{figure*}[h!]
    \centering
    \begin{minipage}{\textwidth} 
        \centering
        \begin{subfigure}{0.48\textwidth}
            \centering
            \includegraphics[width=\textwidth]{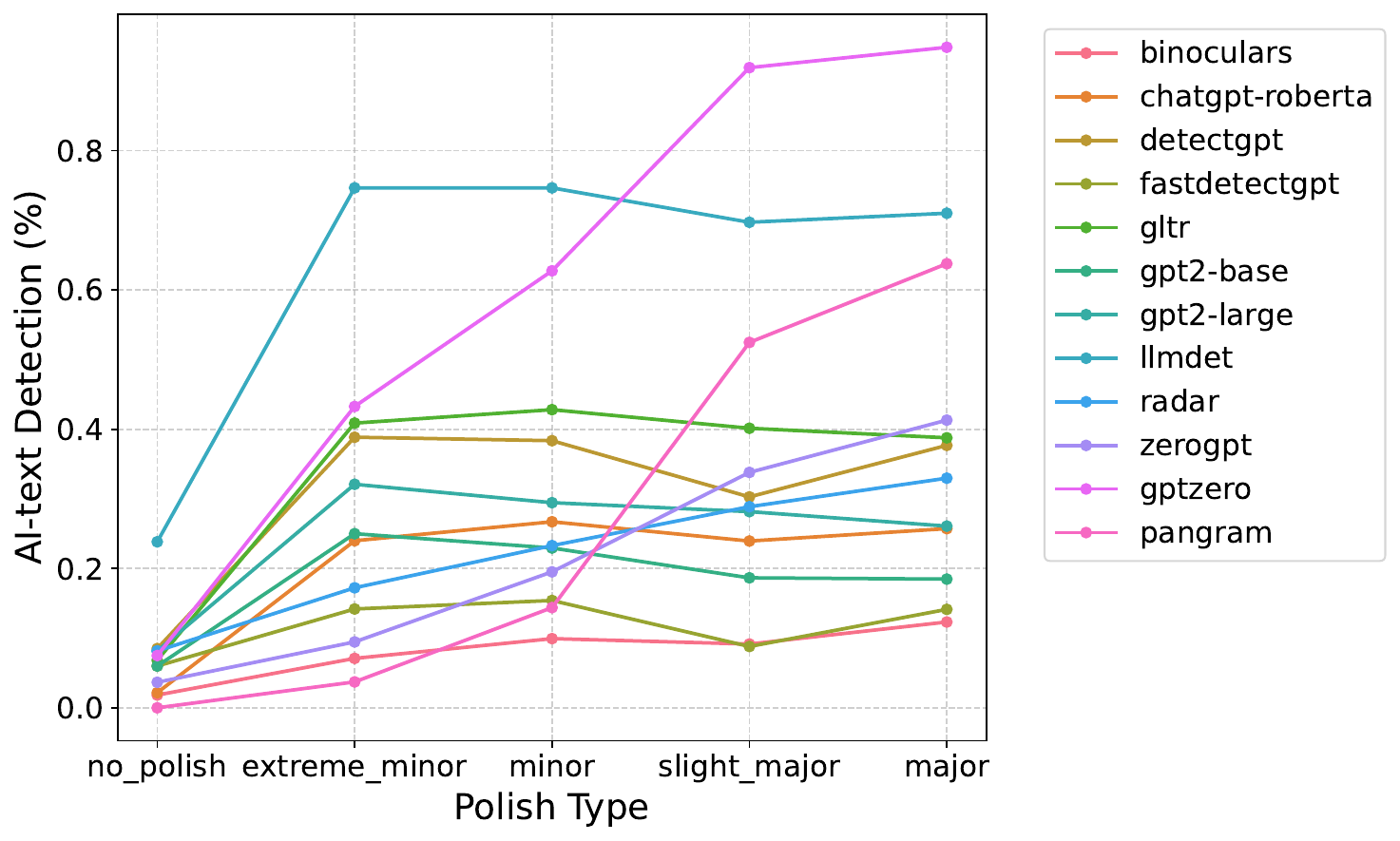}
            \caption{GPT-4o}
        \end{subfigure}
        \hfill
        \begin{subfigure}{0.48\textwidth}
            \centering
            \includegraphics[width=\textwidth]{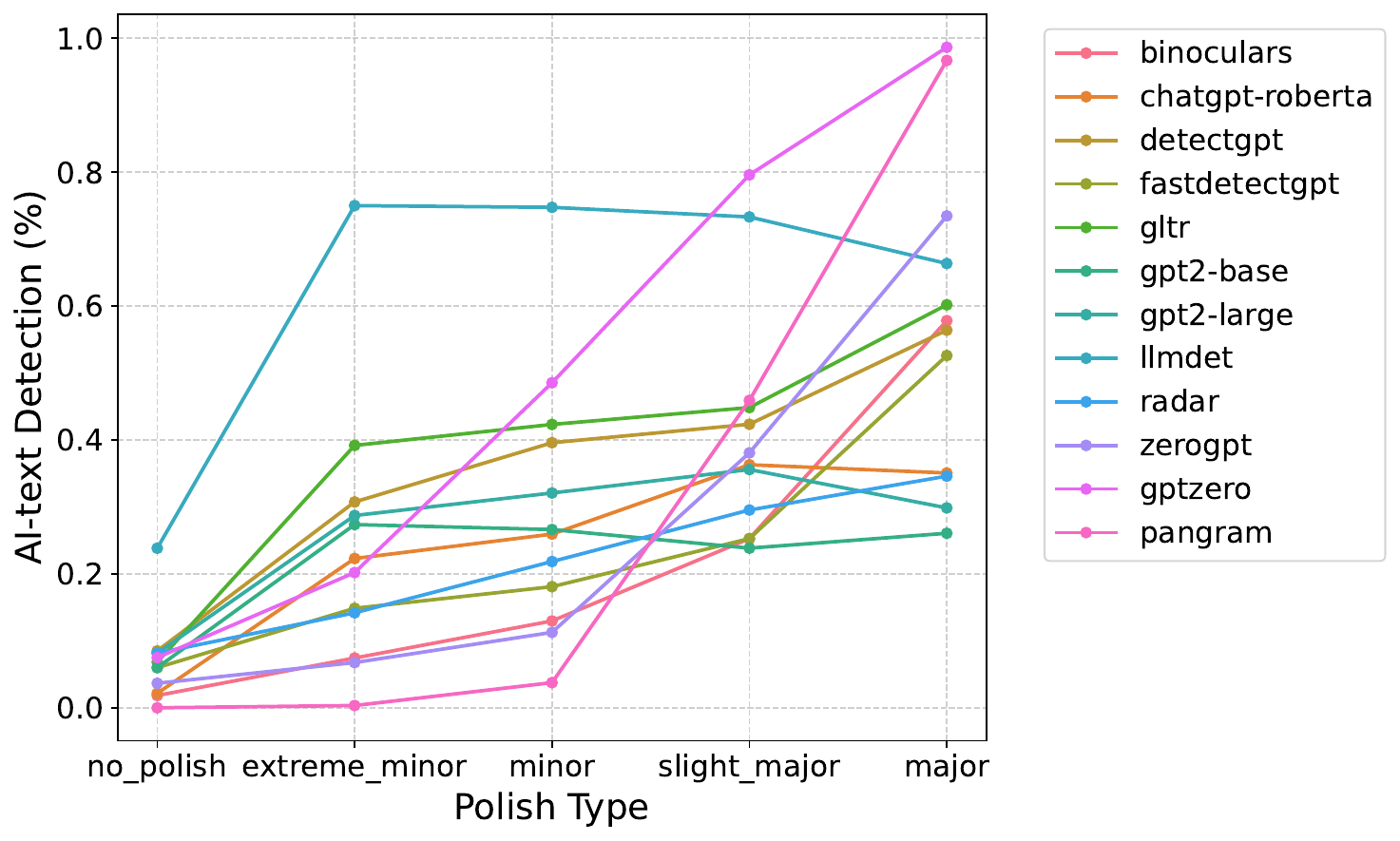}
            \caption{Llama3.1-70b}
        \end{subfigure}
    \end{minipage}
    
    \vspace{0.5cm} 
    \begin{minipage}{\textwidth}
    \centering
        \begin{subfigure}{0.48\textwidth}
            \centering
            \includegraphics[width=\textwidth]{images_new/result_plots/combined_model_polish_type_vs_mgt_llama2.pdf}
            \caption{Llama2-7B}
        \end{subfigure}
        \hfill
        \begin{subfigure}{0.48\textwidth}
            \centering
            \includegraphics[width=\textwidth]{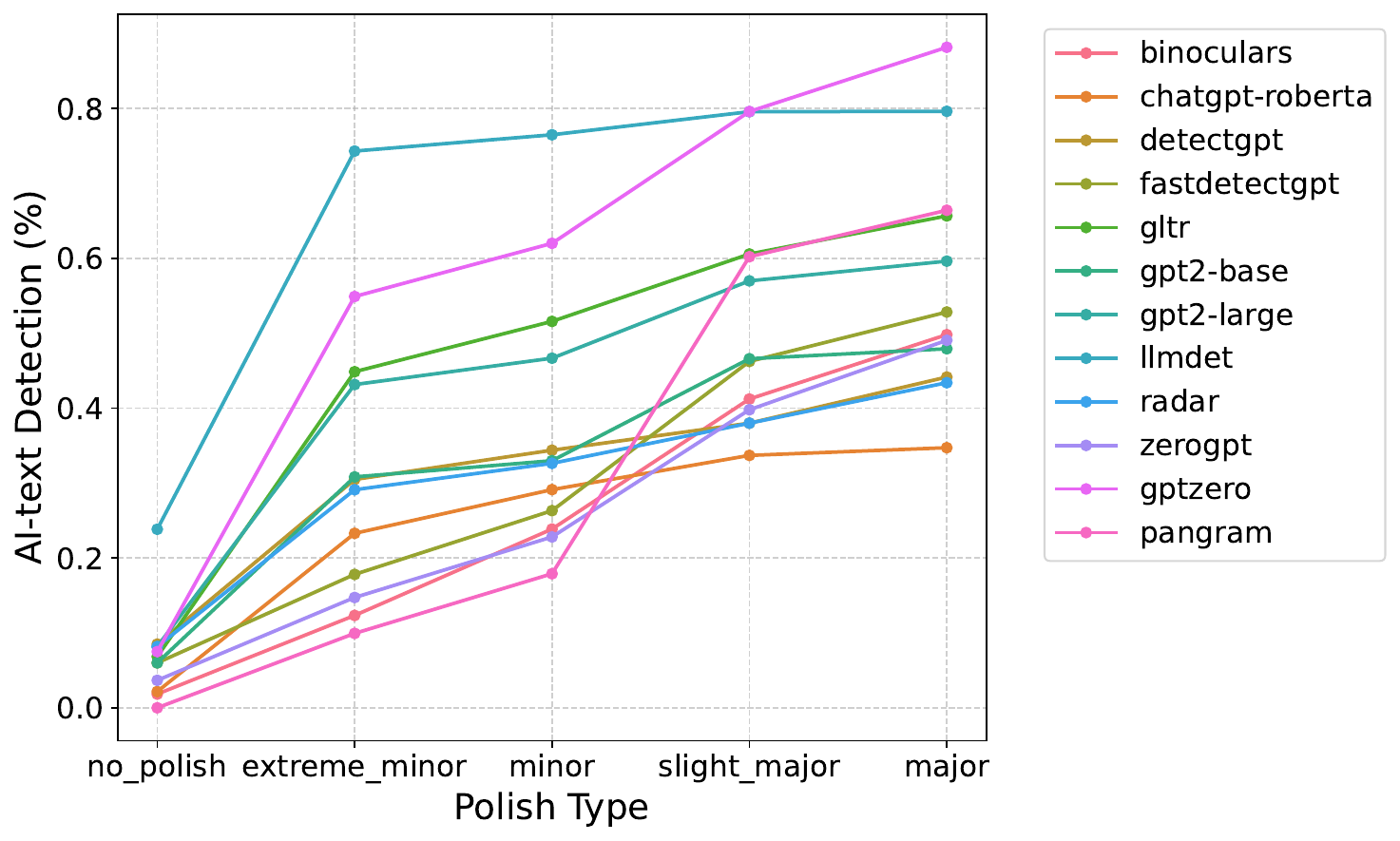}
            \caption{Llama3-8B}
        \end{subfigure}
    \end{minipage}

    \vspace{0.5cm} 
    \begin{minipage}{\textwidth}
    \centering
        \begin{subfigure}{0.48\textwidth}
            \centering
            \includegraphics[width=\textwidth]{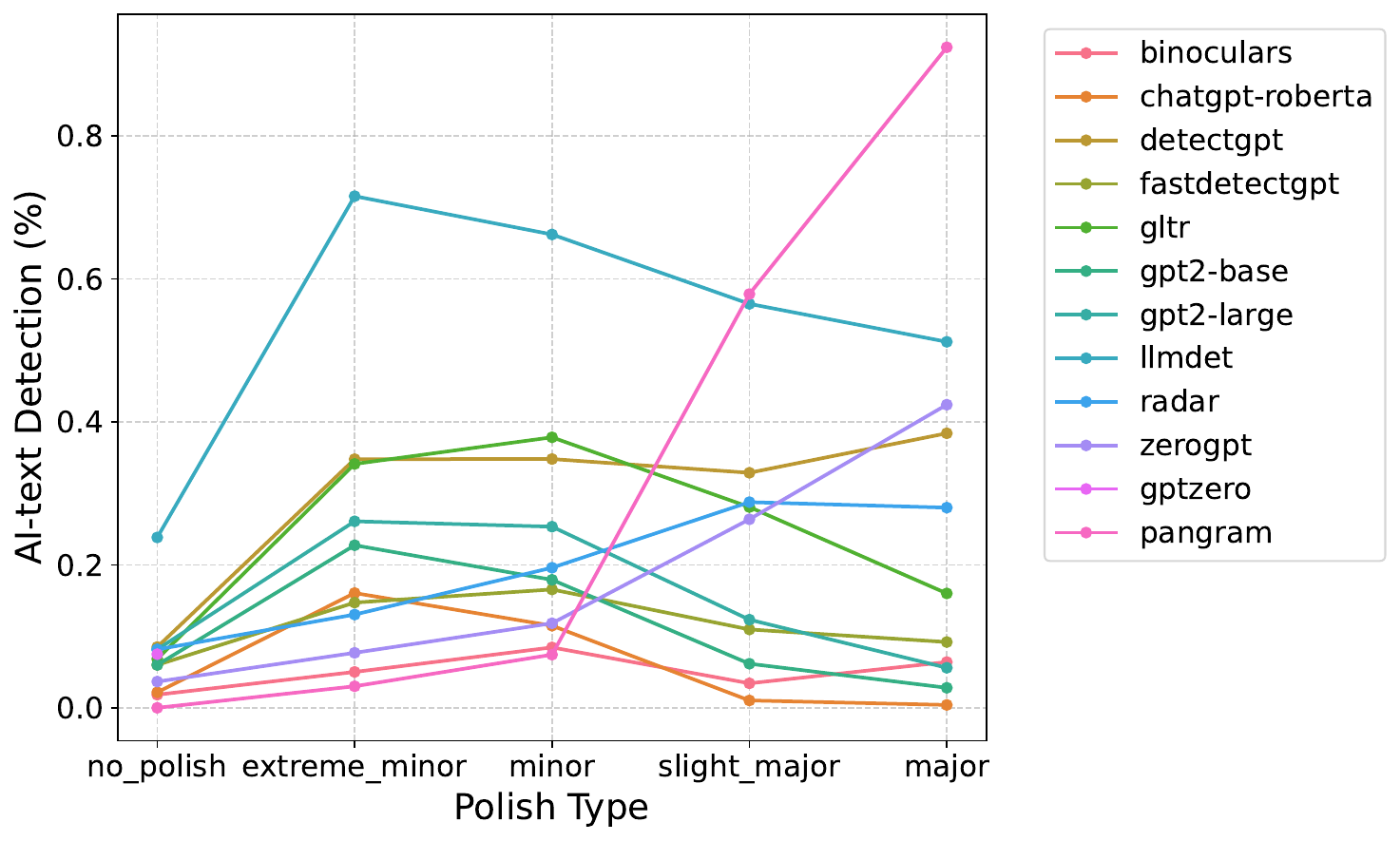}
            \caption{DeepSeek-V3}
        \end{subfigure}
    \end{minipage}
    
    \caption{AI-text detection rate for \textbf{degree-based} AI-polished-texts (APT) by all detectors.}
    \label{fig:result_combined_degree}
\end{figure*}

\begin{figure*}[h!]
    \centering
    \begin{minipage}{\textwidth} 
        \centering
        \begin{subfigure}{0.48\textwidth}
            \centering
            \includegraphics[width=\textwidth]{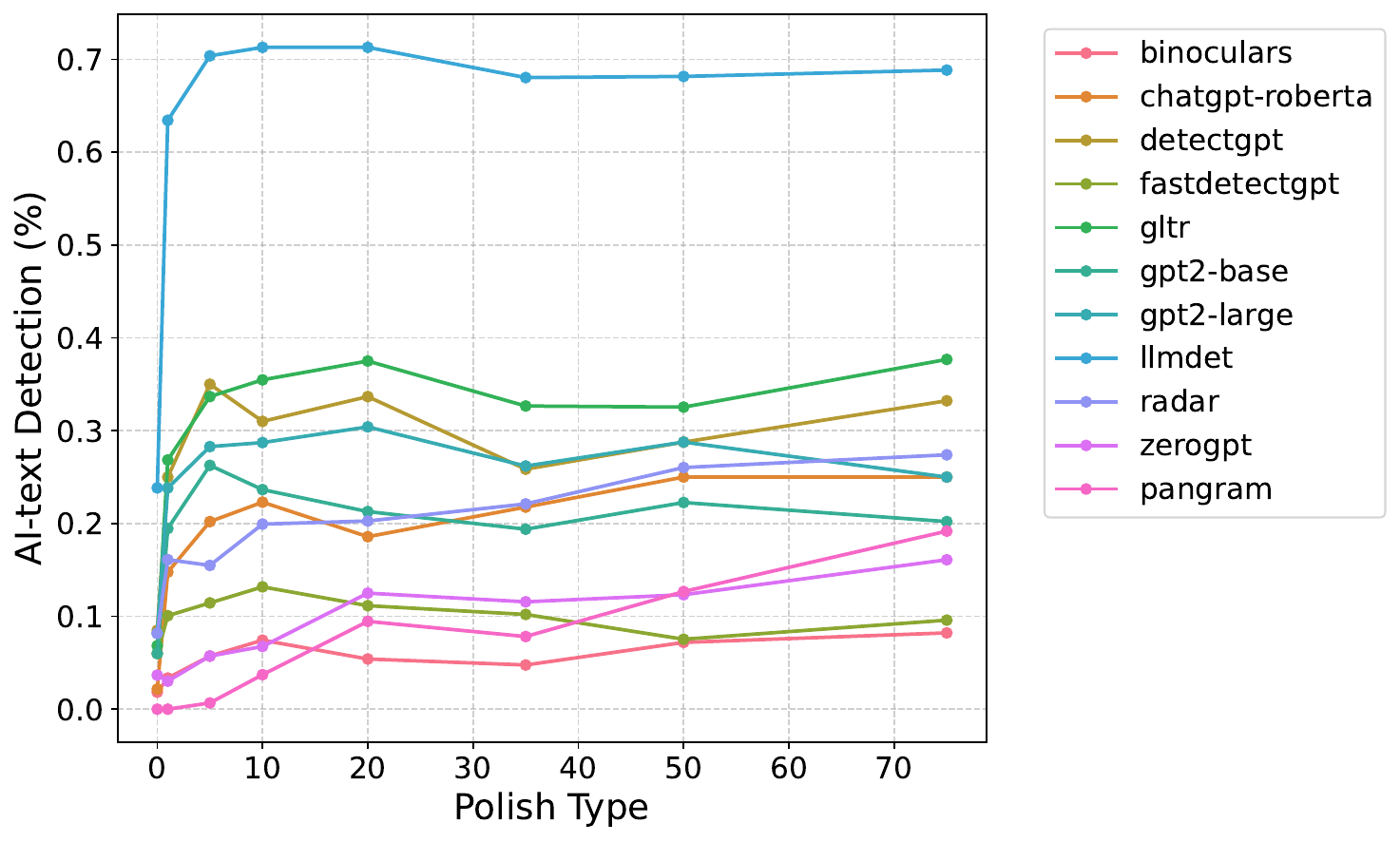}
            \caption{GPT-4o}
        \end{subfigure}
        \hfill
        \begin{subfigure}{0.48\textwidth}
            \centering
            \includegraphics[width=\textwidth]{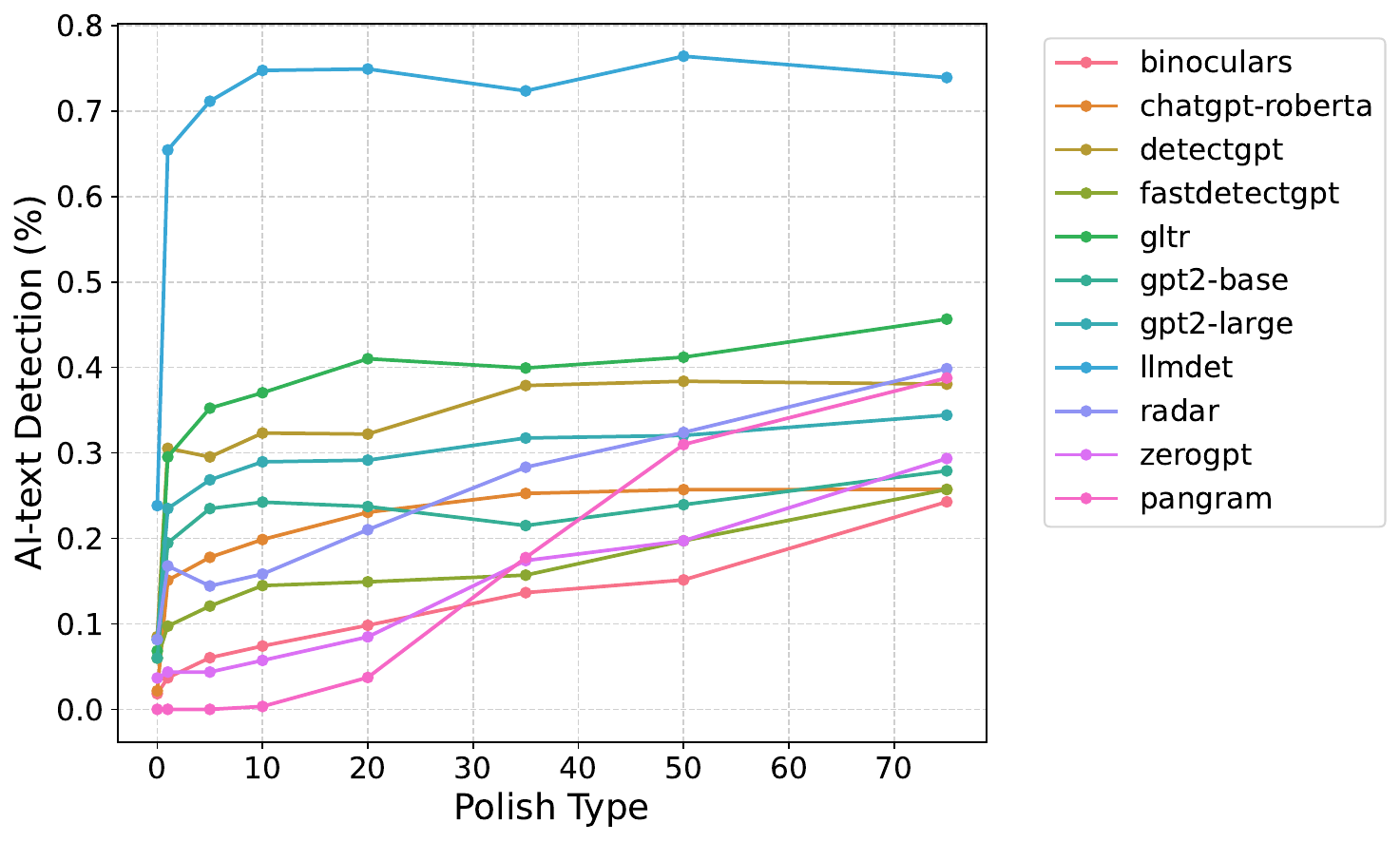}
            \caption{Llama3.1-70b}
        \end{subfigure}
    \end{minipage}
    
    \vspace{0.5cm} 
    \begin{minipage}{\textwidth}
    \centering
        \begin{subfigure}{0.48\textwidth}
            \centering
            \includegraphics[width=\textwidth]{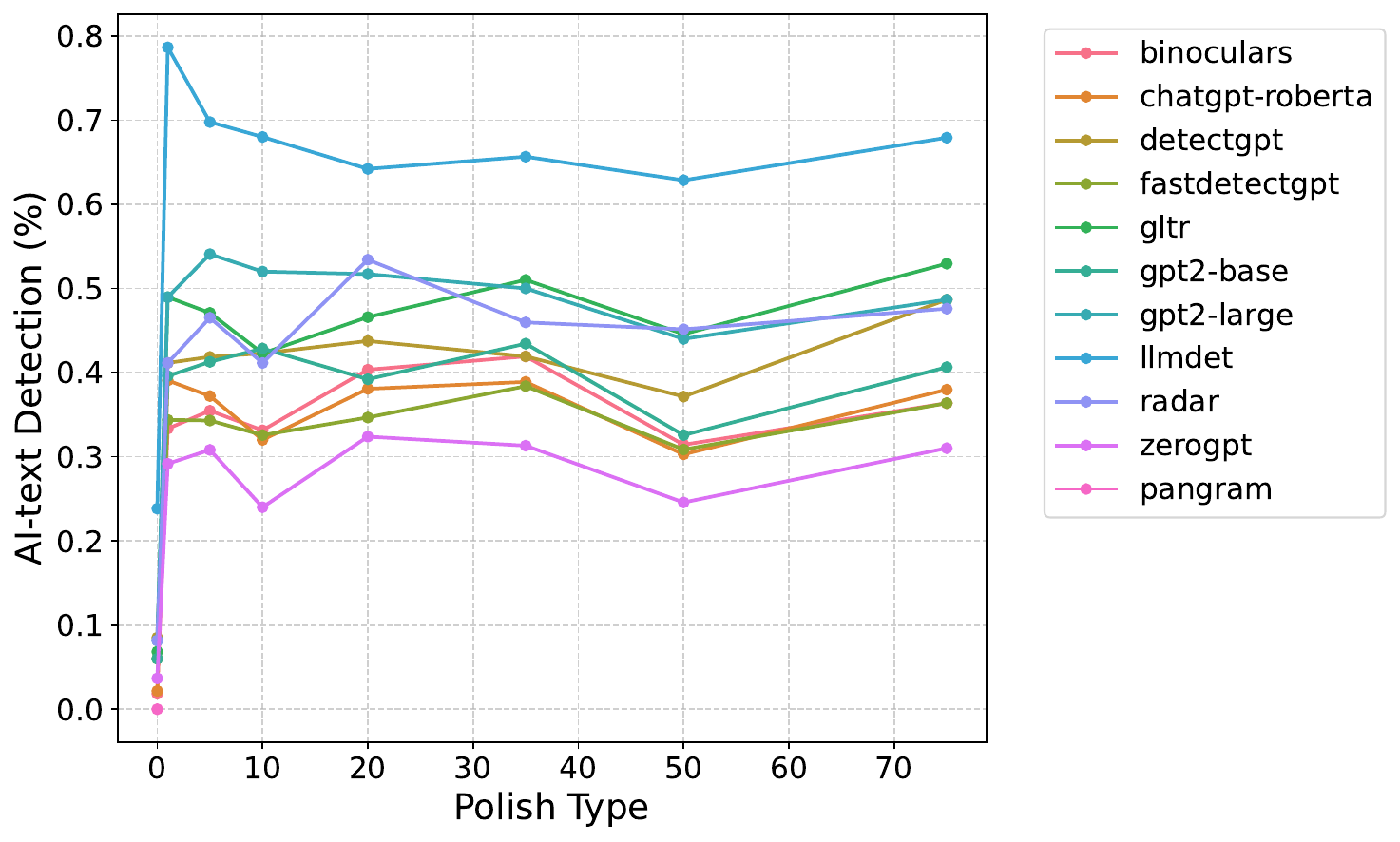}
            \caption{Llama2-7B}
        \end{subfigure}
        \hfill
        \begin{subfigure}{0.48\textwidth}
            \centering
            \includegraphics[width=\textwidth]{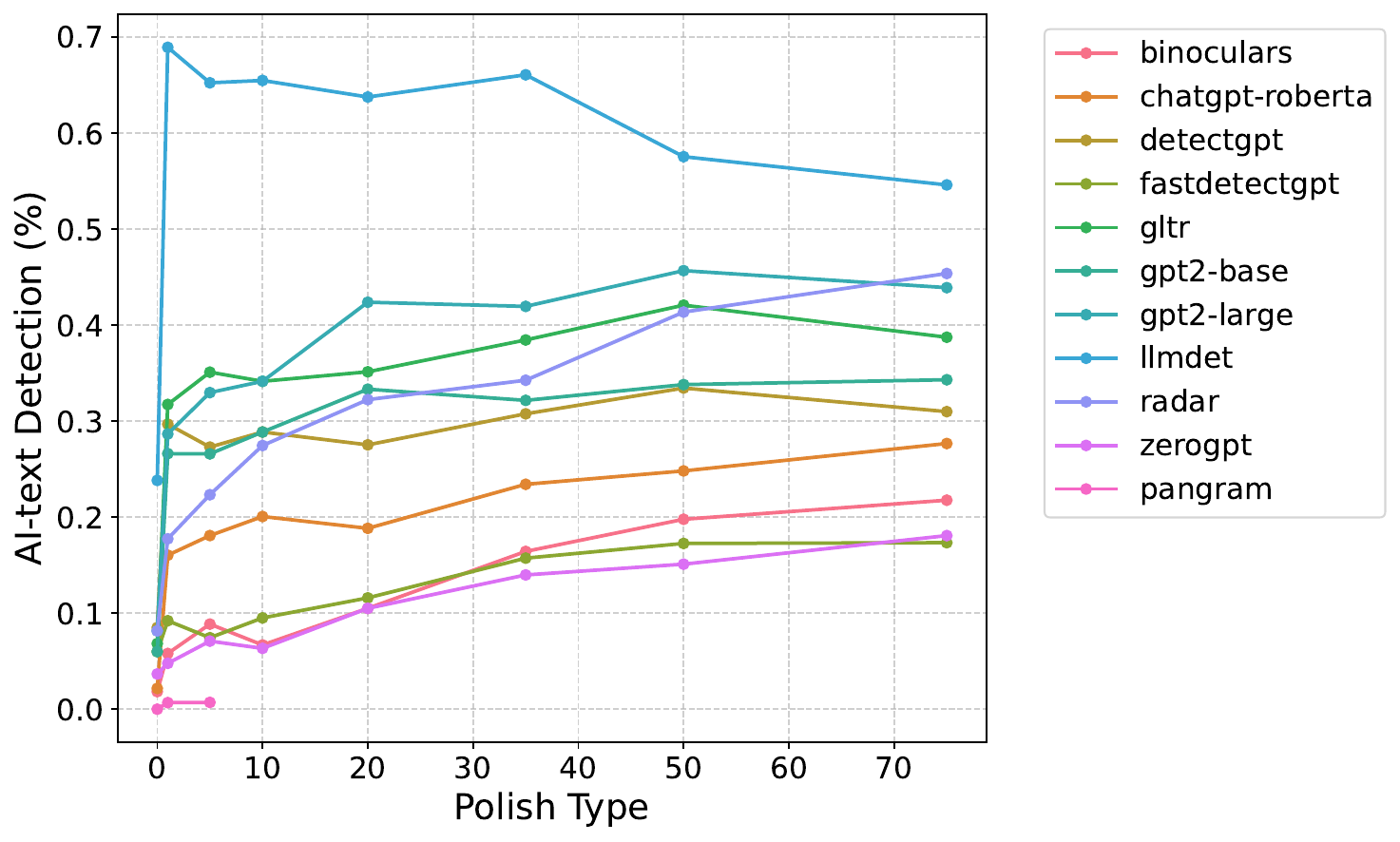}
            \caption{Llama3-8B}
        \end{subfigure}
    \end{minipage}

    \vspace{0.5cm} 
    \begin{minipage}{\textwidth}
    \centering
        \begin{subfigure}{0.48\textwidth}
            \centering
            \includegraphics[width=\textwidth]{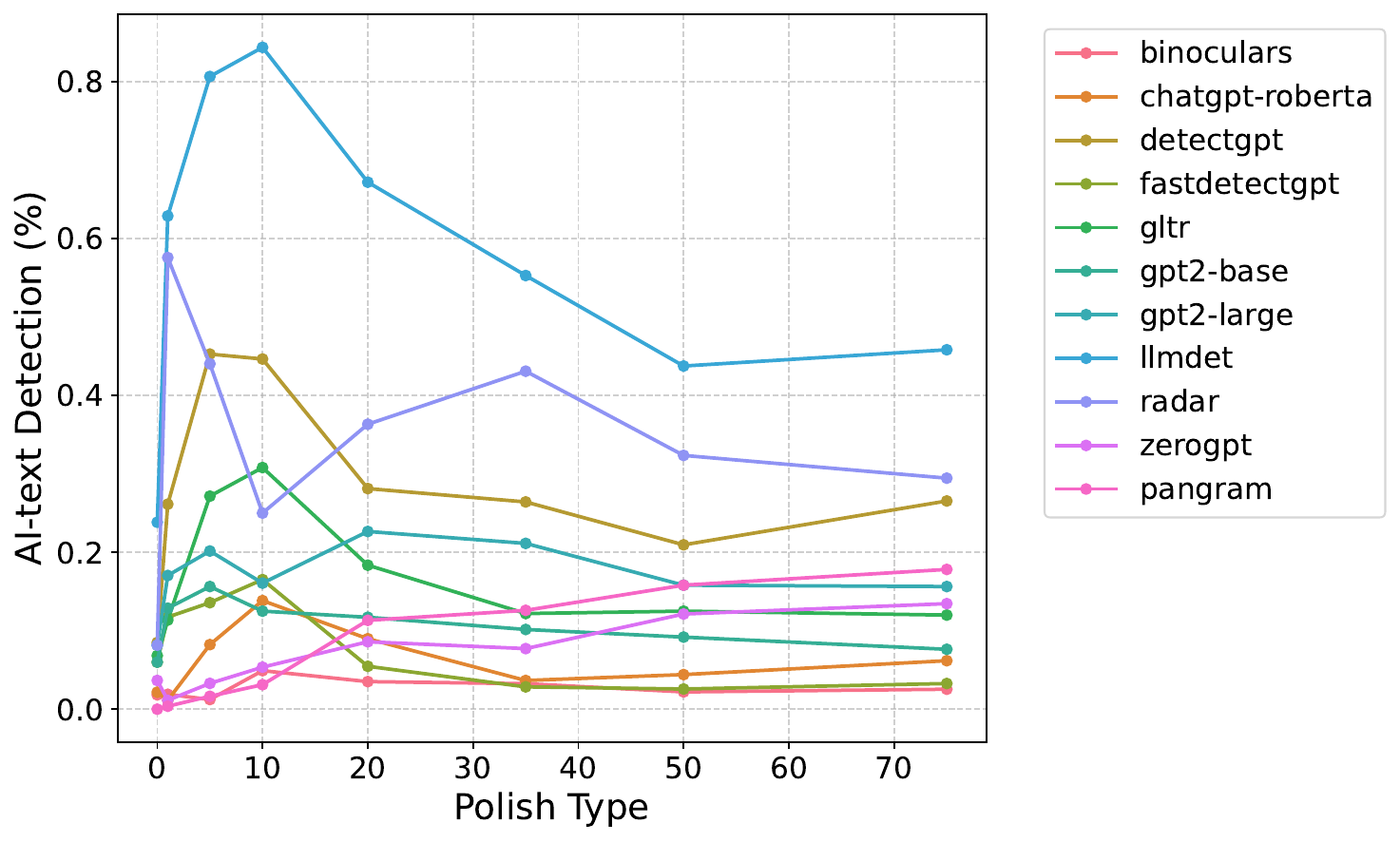}
            \caption{DeepSeek-V3}
        \end{subfigure}
    \end{minipage}
    
    \caption{AI-text detection rate for \textbf{percentage-based} AI-polished-texts (APT) by all detectors.}
    \label{fig:result_combined_prct}
\end{figure*}

\clearpage
\newpage

\subsection{Polisher LLM Specific Results} \label{app:polisher_results}
\begin{figure}[h!]
    \centering
    \includegraphics[width=0.7\linewidth]{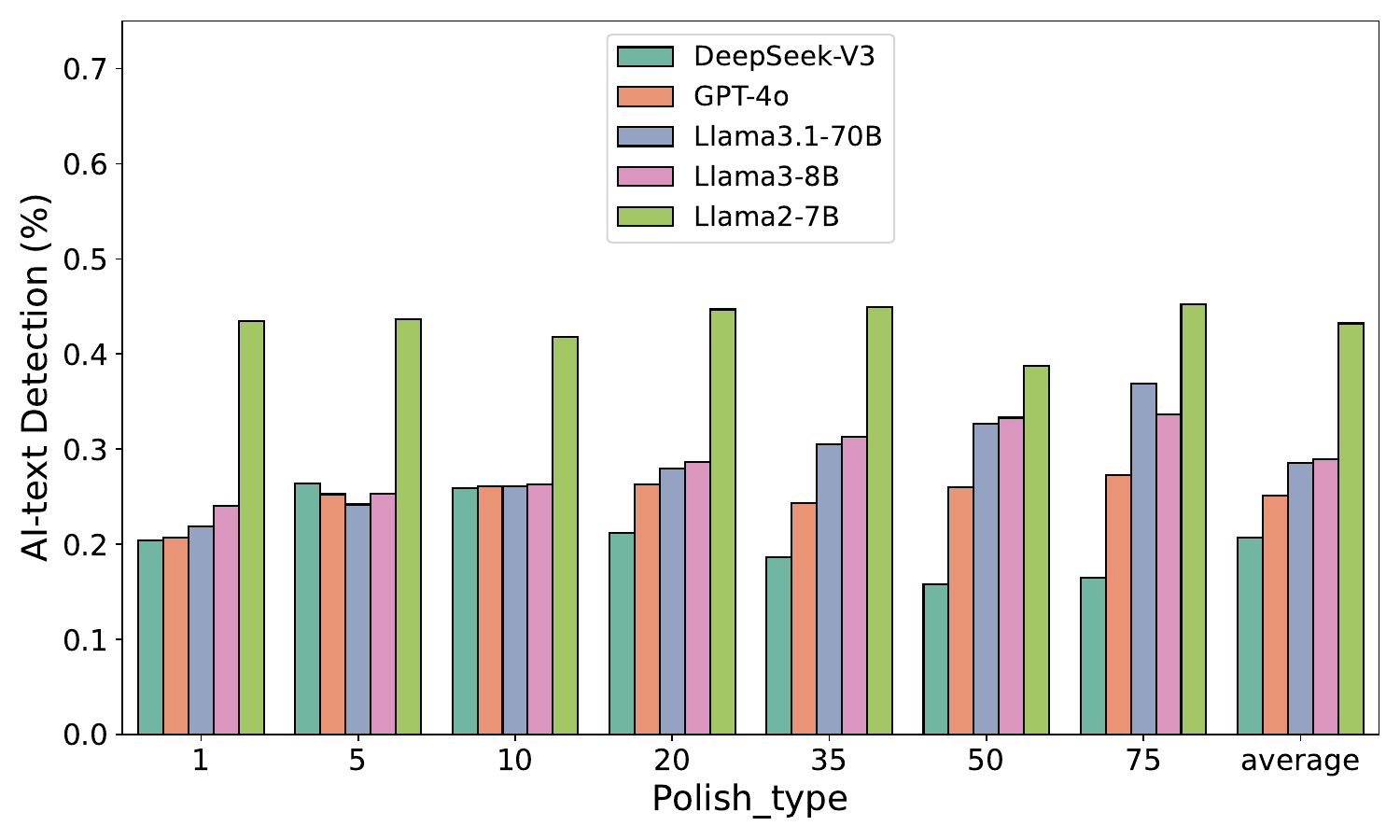}
    \caption{AI-text detection rate for \textbf{percentage-based} AI-polished-texts from different polisher LLMs.}
    \label{fig:polisher_vary_ratio}
\end{figure}

\begin{figure}[h!]
    \centering
    \includegraphics[width=0.7\linewidth]{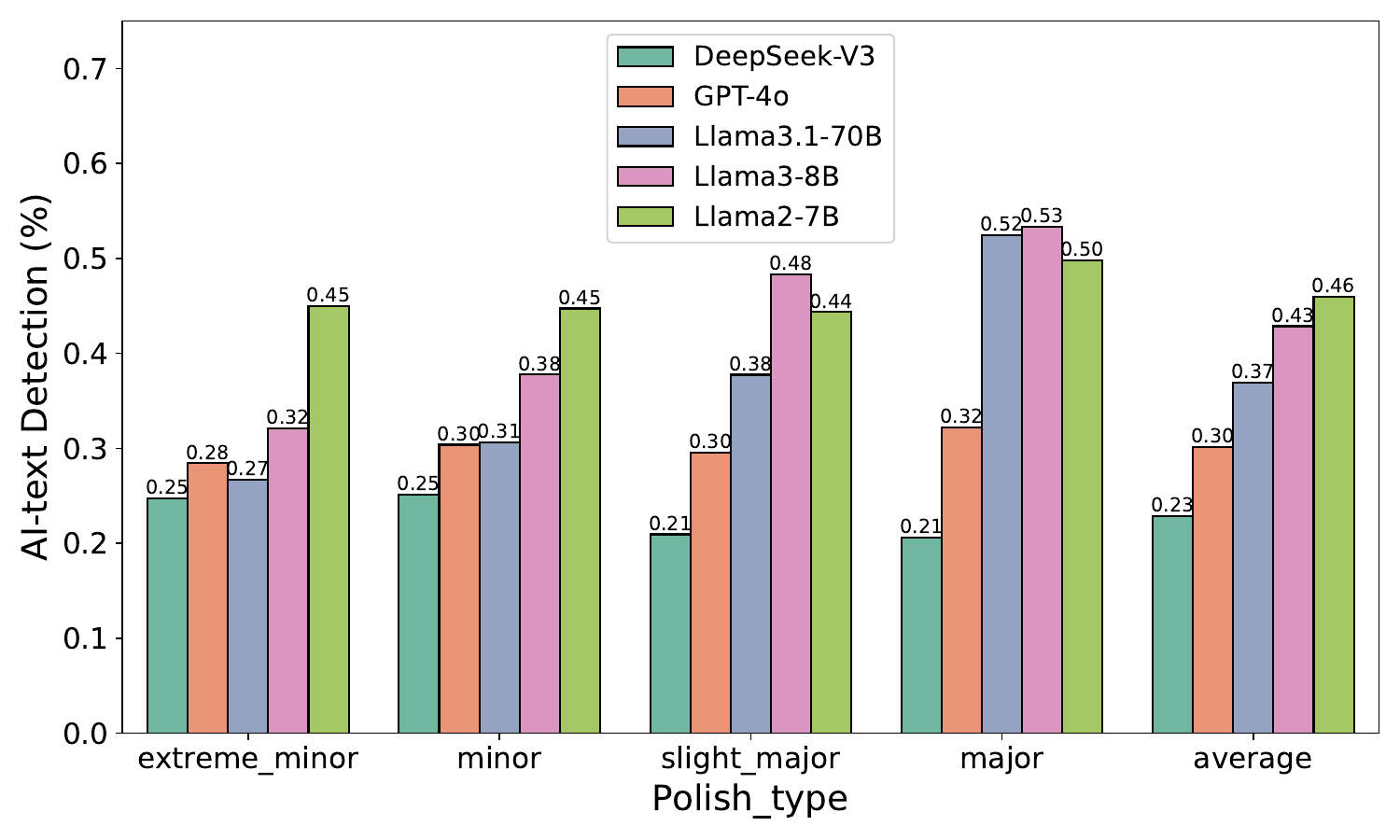}
    \caption{AI-text detection rate for \textbf{degree-based} AI-polished-texts from different polisher LLMs.}
    \label{fig:polisher_vary_degree}
\end{figure}

\clearpage
\newpage

\subsection{Domain Specific Results} \label{app:domain_results}
\begin{figure*}[h!]
    \centering
    \begin{minipage}{\textwidth} 
        \centering
        \begin{subfigure}{0.24\textwidth}
            \centering
            \includegraphics[width=\textwidth]{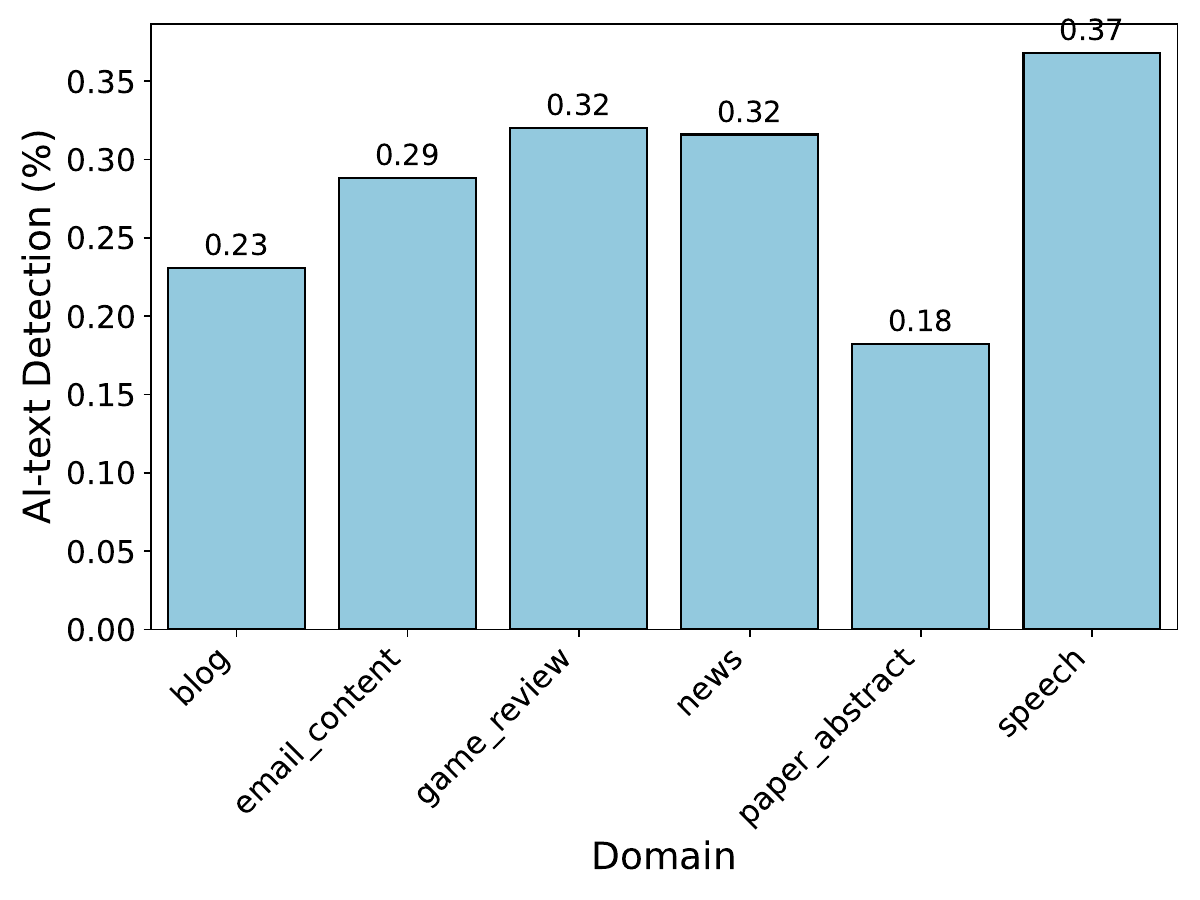}
            \caption{GPT-4o}
        \end{subfigure}
        \hfill
        \begin{subfigure}{0.24\textwidth}
            \centering
            \includegraphics[width=\textwidth]{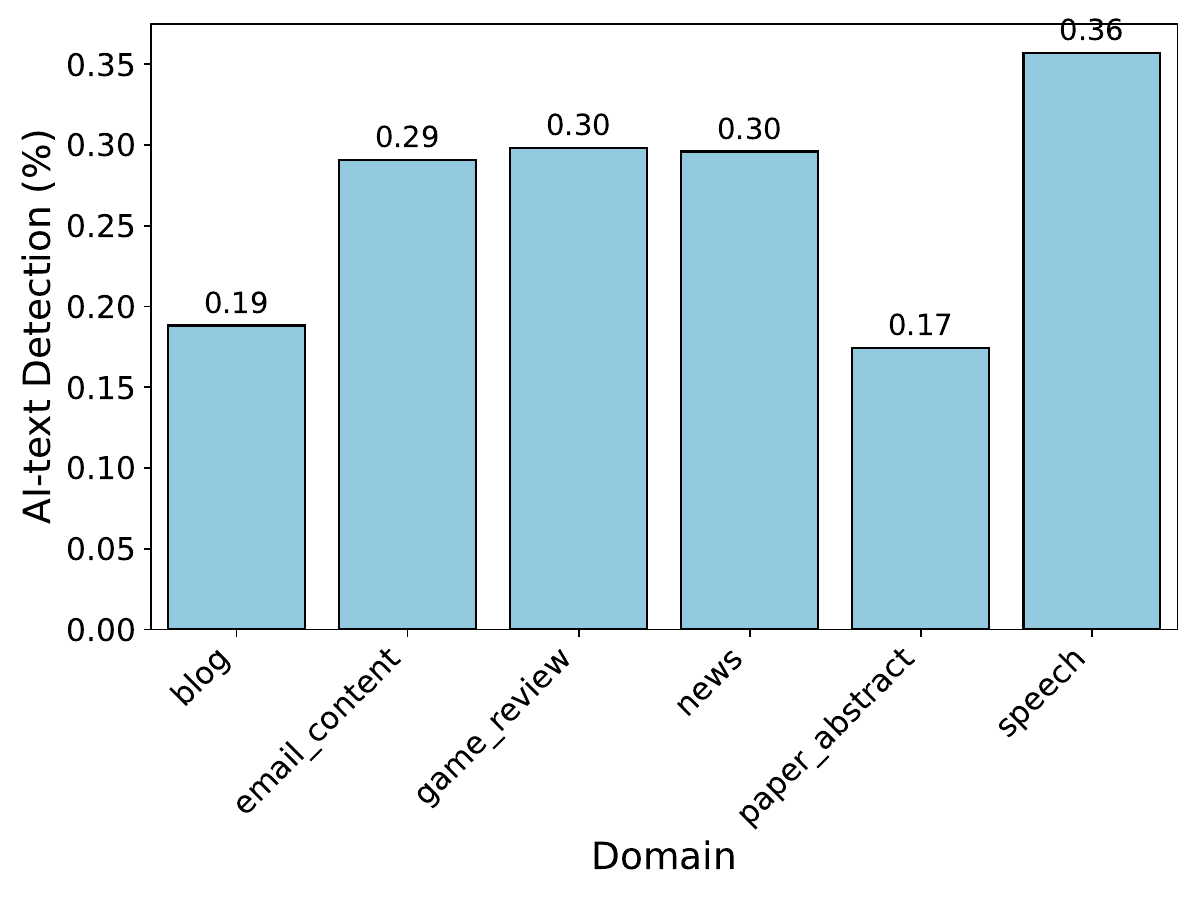}
            \caption{Llama3.1-70b}
        \end{subfigure}
        \hfill
        \begin{subfigure}{0.24\textwidth}
            \centering
            \includegraphics[width=\textwidth]{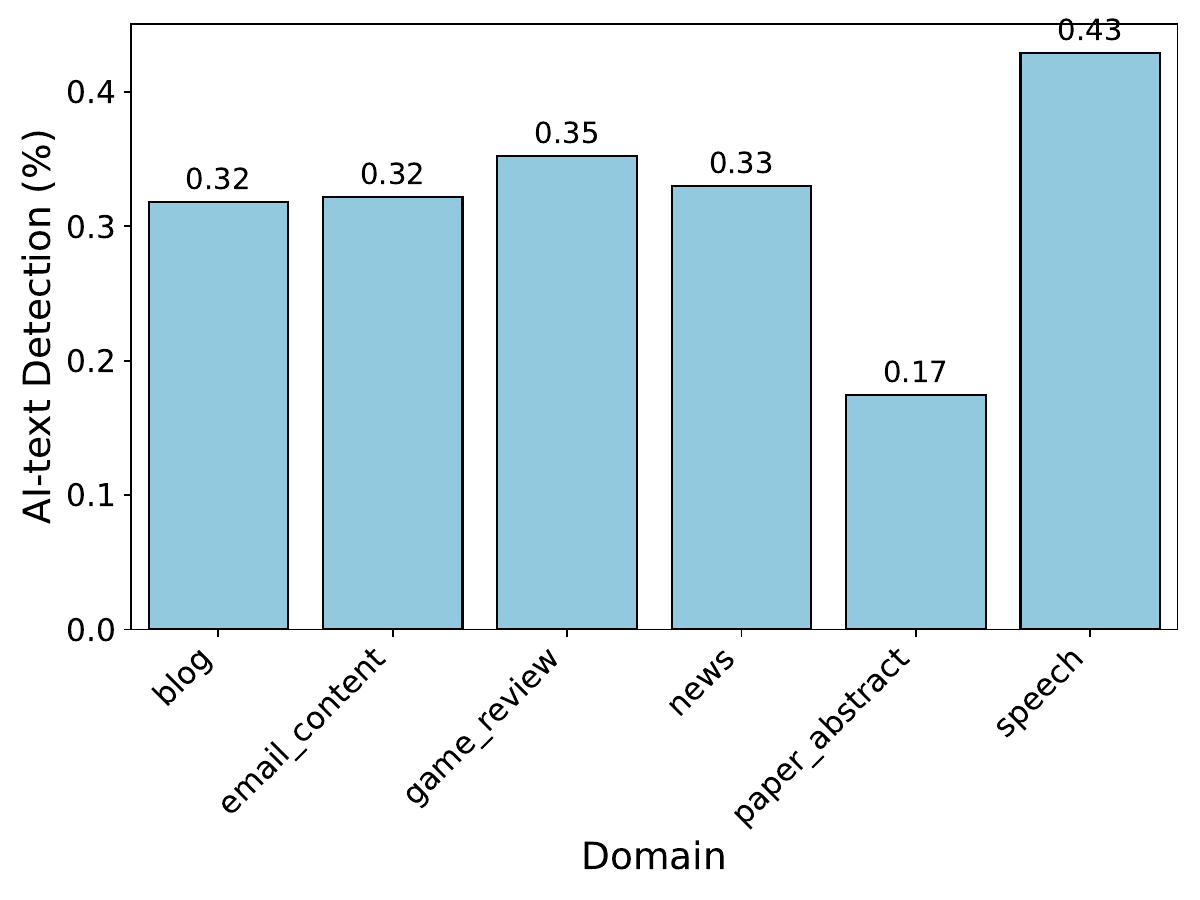}
            \caption{Llama3-8b}
        \end{subfigure}
        \hfill
        \begin{subfigure}{0.24\textwidth}
            \centering
            \includegraphics[width=\textwidth]{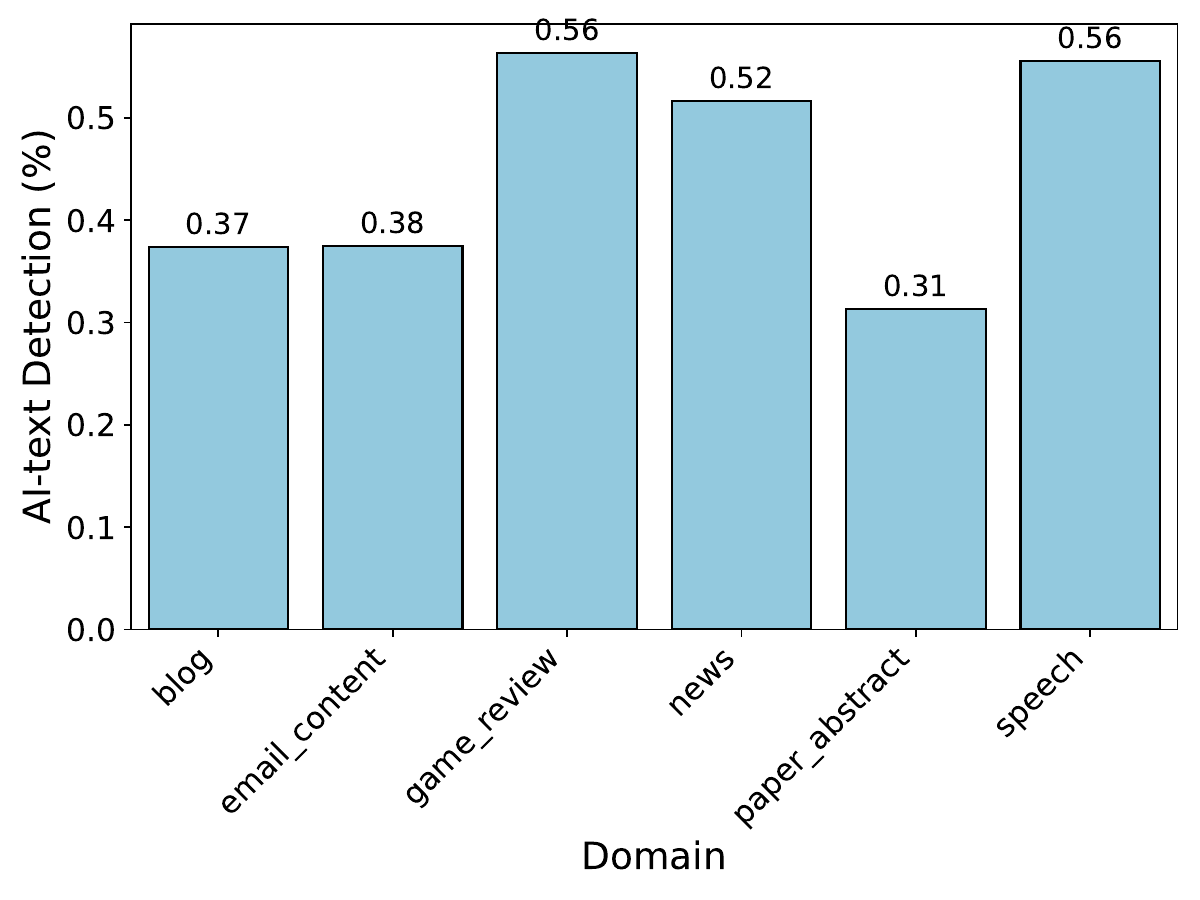}
            \caption{Llama2-7b}
        \end{subfigure}
        \\ 
        \vspace{0.2cm}
        \small \textbf{Extreme-minor Polishing}
    \end{minipage}
    
    \vspace{0.5cm} 
    \begin{minipage}{\textwidth}
    \centering
        \begin{subfigure}{0.24\textwidth}
            \centering
            \includegraphics[width=\textwidth]{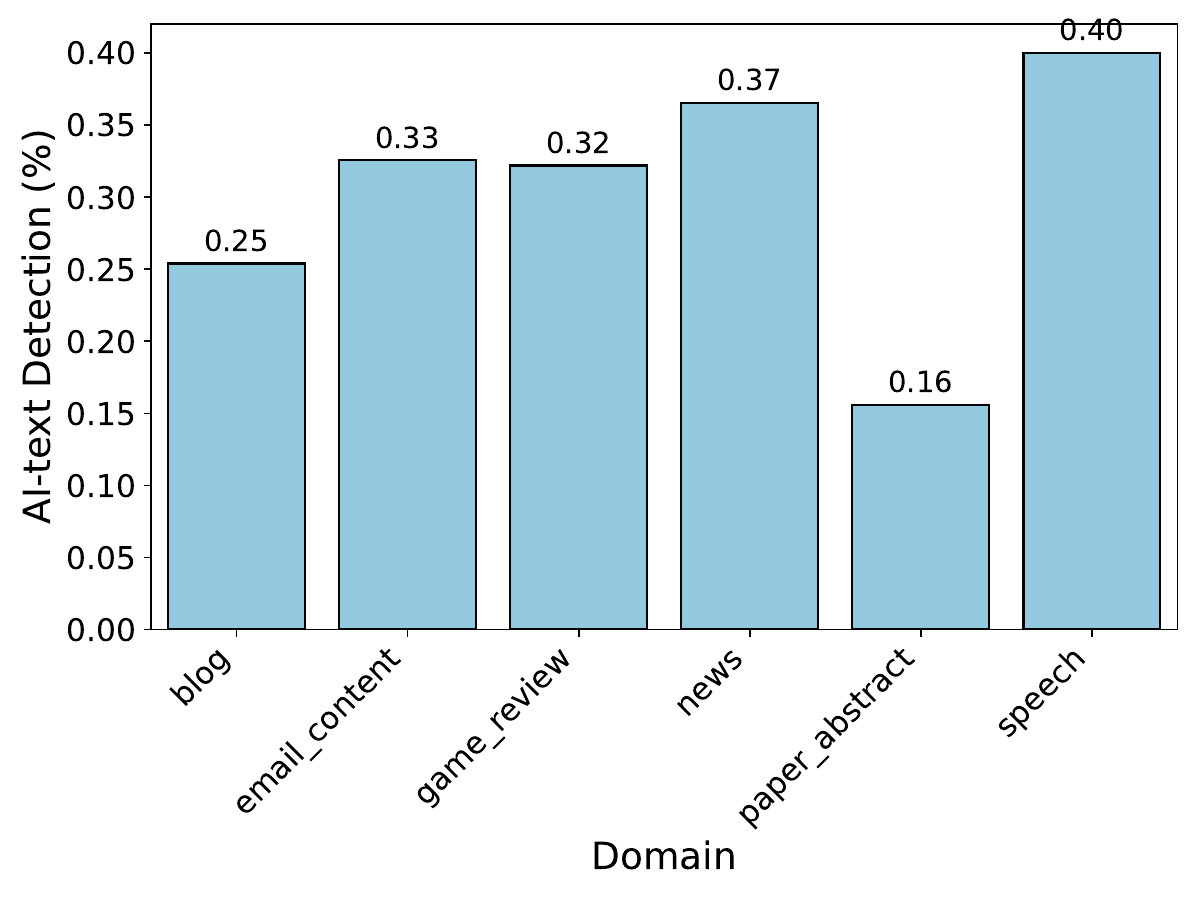}
            \caption{GPT-4o}
        \end{subfigure}
        \hfill
        \begin{subfigure}{0.24\textwidth}
            \centering
            \includegraphics[width=\textwidth]{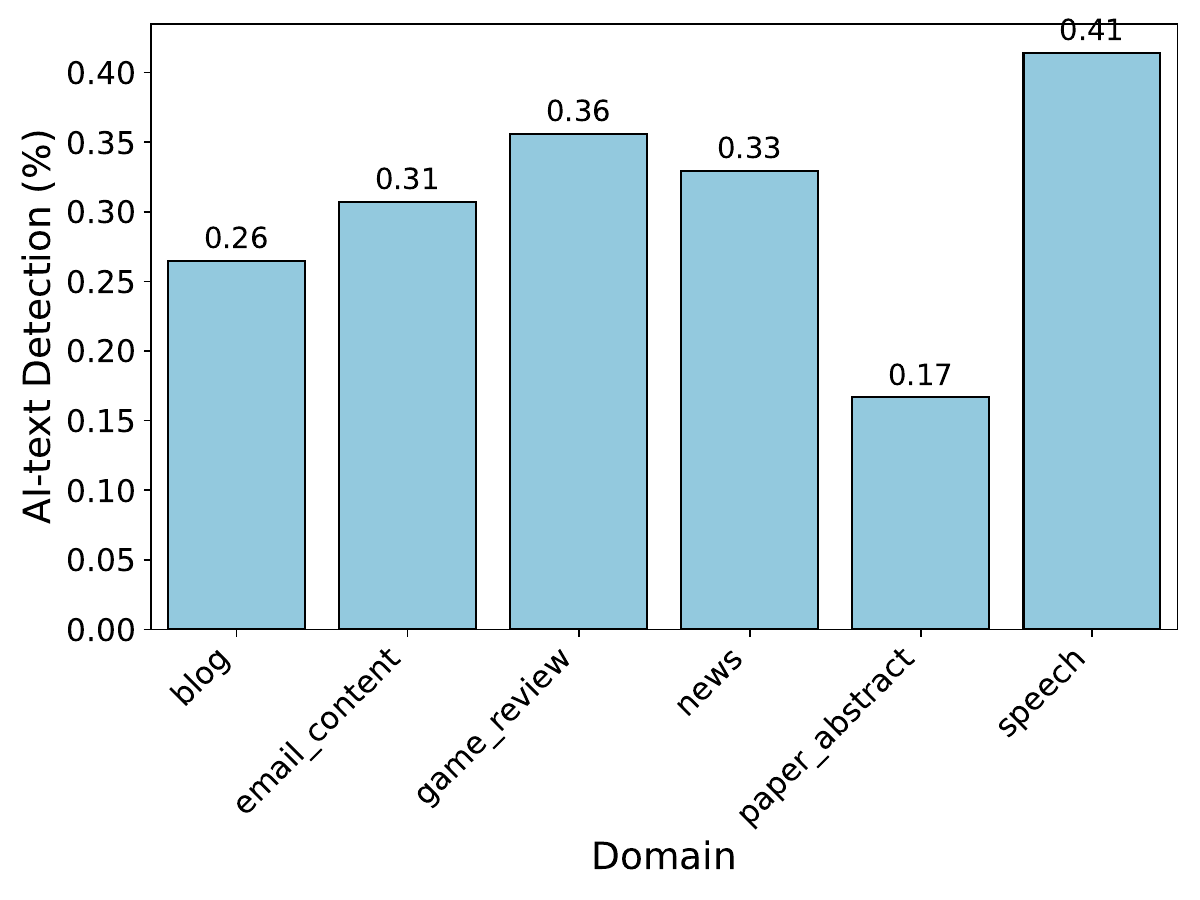}
            \caption{Llama3.1-70b}
        \end{subfigure}
        \hfill
        \begin{subfigure}{0.24\textwidth}
            \centering
            \includegraphics[width=\textwidth]{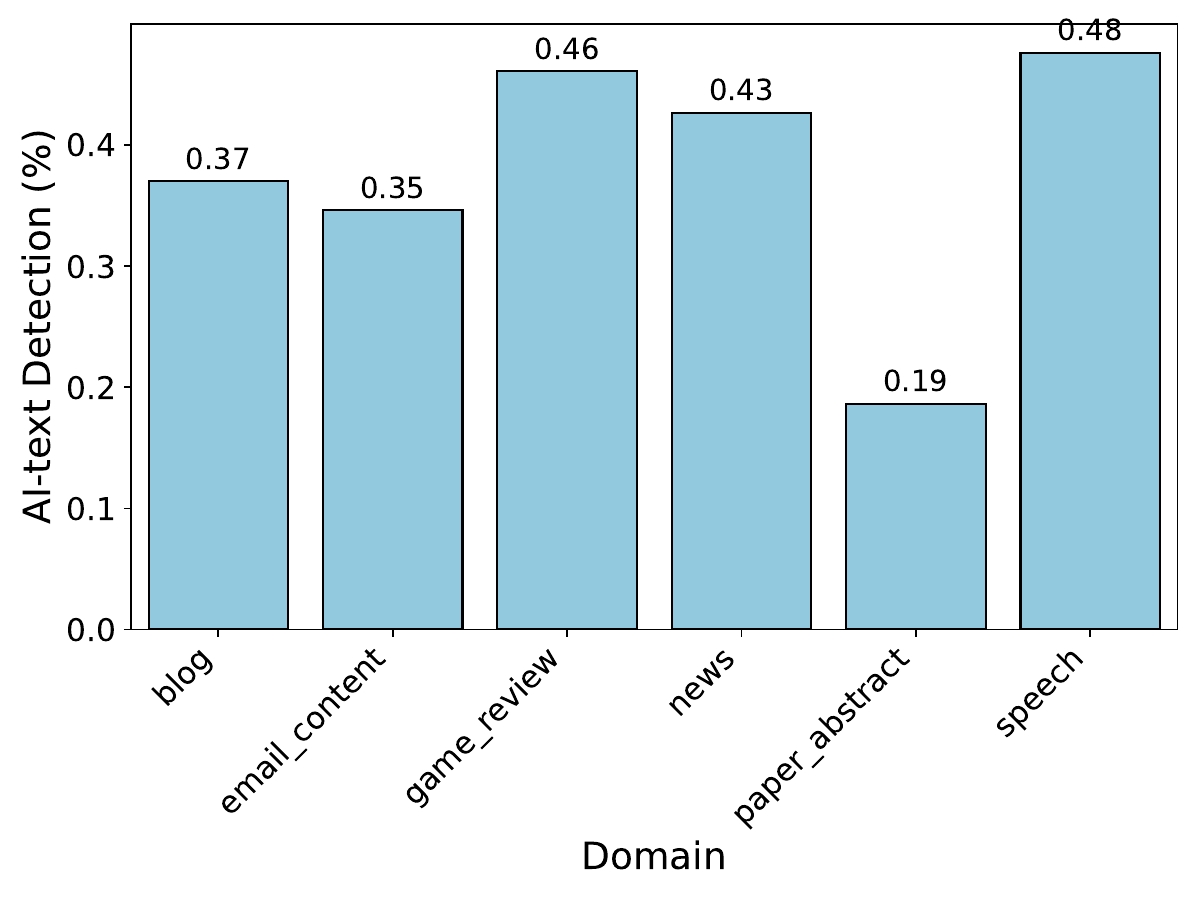}
            \caption{Llama3-8b}
        \end{subfigure}
        \hfill
        \begin{subfigure}{0.24\textwidth}
            \centering
            \includegraphics[width=\textwidth]{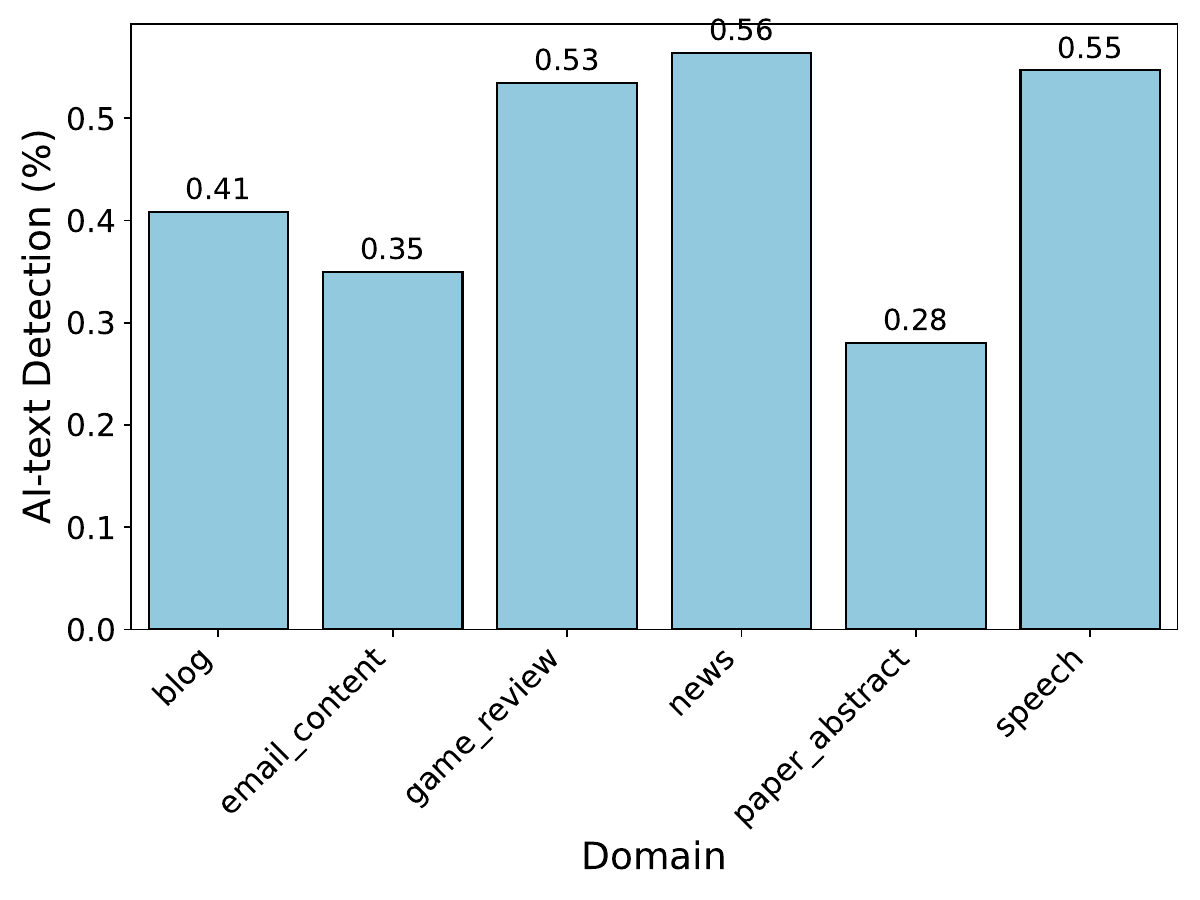}
            \caption{Llama2-7b}
        \end{subfigure}
        \\ 
        \vspace{0.2cm}
        \small \textbf{Minor Polishing}
    \end{minipage}

    \vspace{0.5cm} 
    \begin{minipage}{\textwidth}
    \centering
        \begin{subfigure}{0.24\textwidth}
            \centering
            \includegraphics[width=\textwidth]{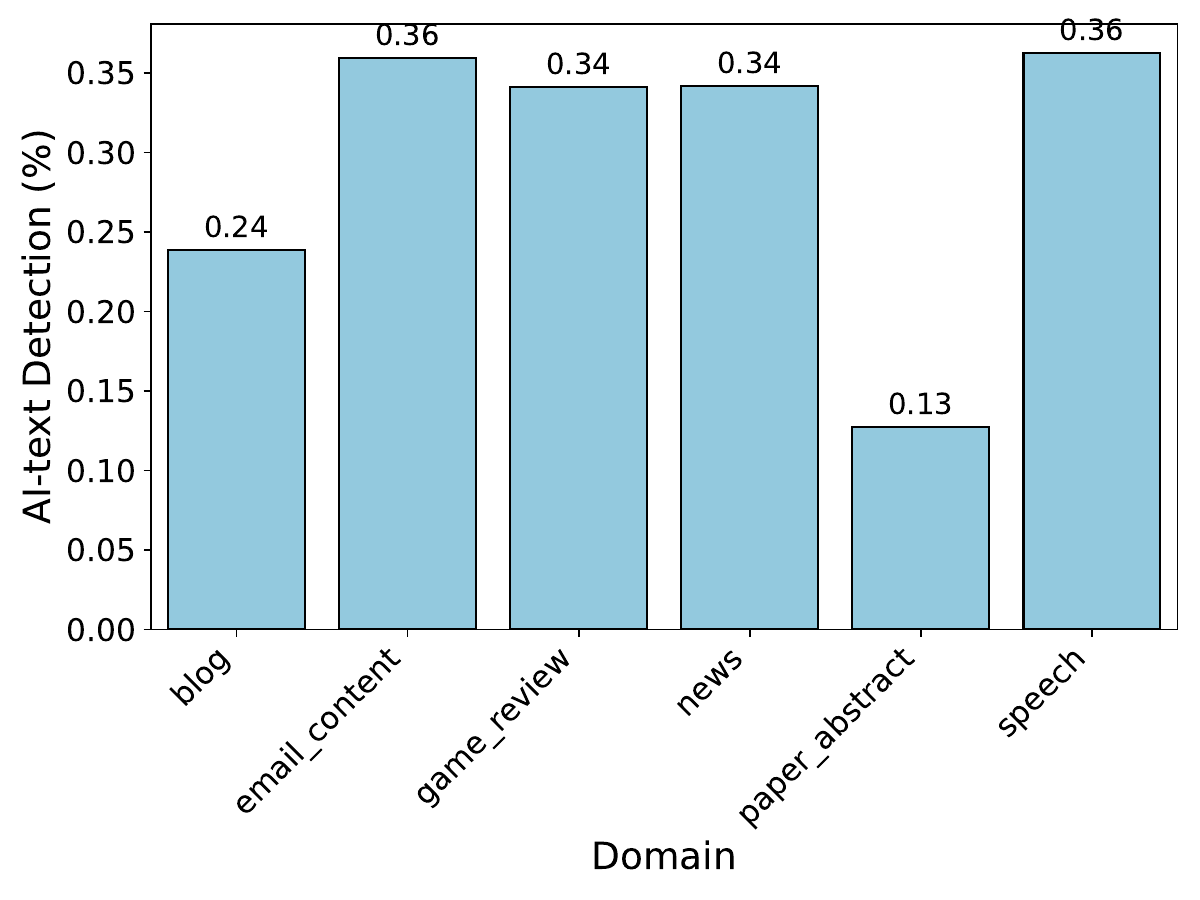}
            \caption{GPT-4o}
        \end{subfigure}
        \hfill
        \begin{subfigure}{0.24\textwidth}
            \centering
            \includegraphics[width=\textwidth]{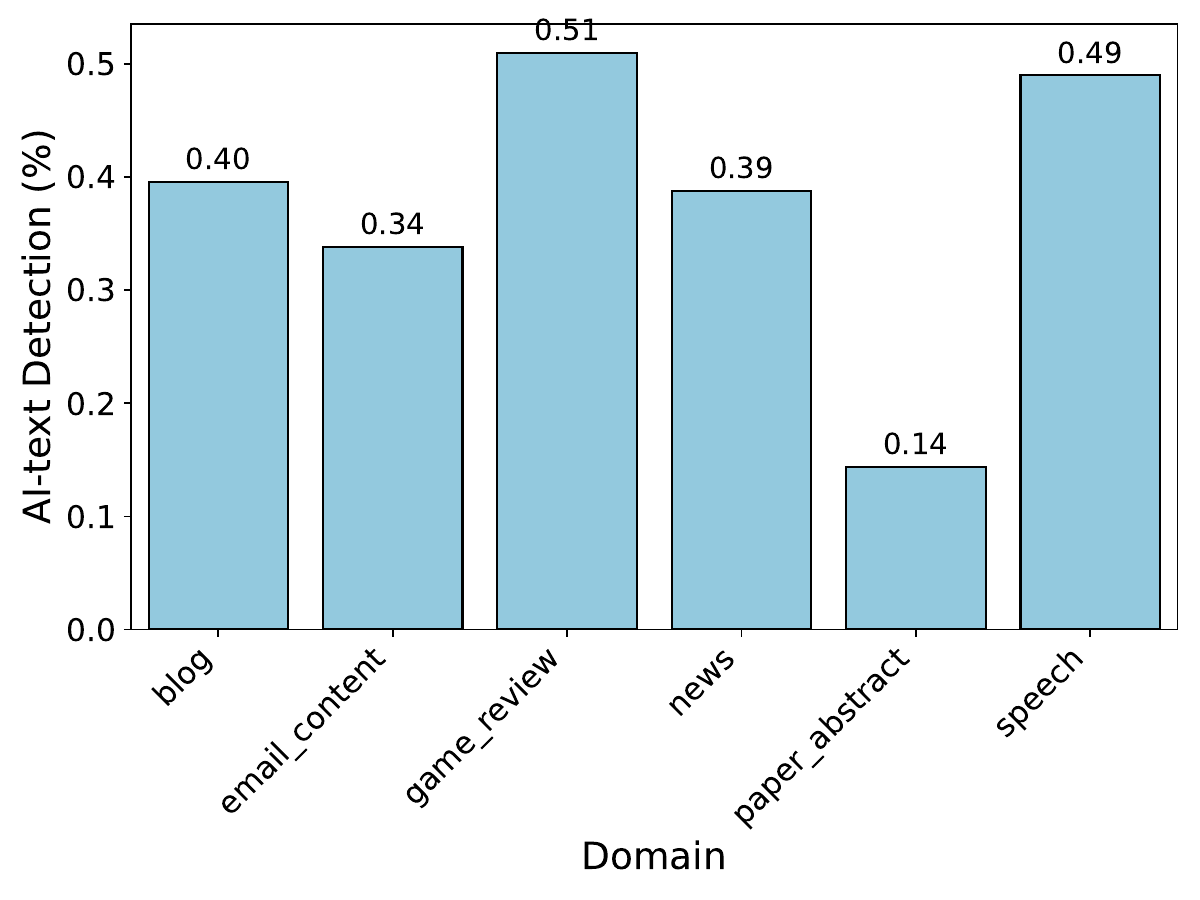}
            \caption{Llama3.1-70b}
        \end{subfigure}
        \hfill
        \begin{subfigure}{0.24\textwidth}
            \centering
            \includegraphics[width=\textwidth]{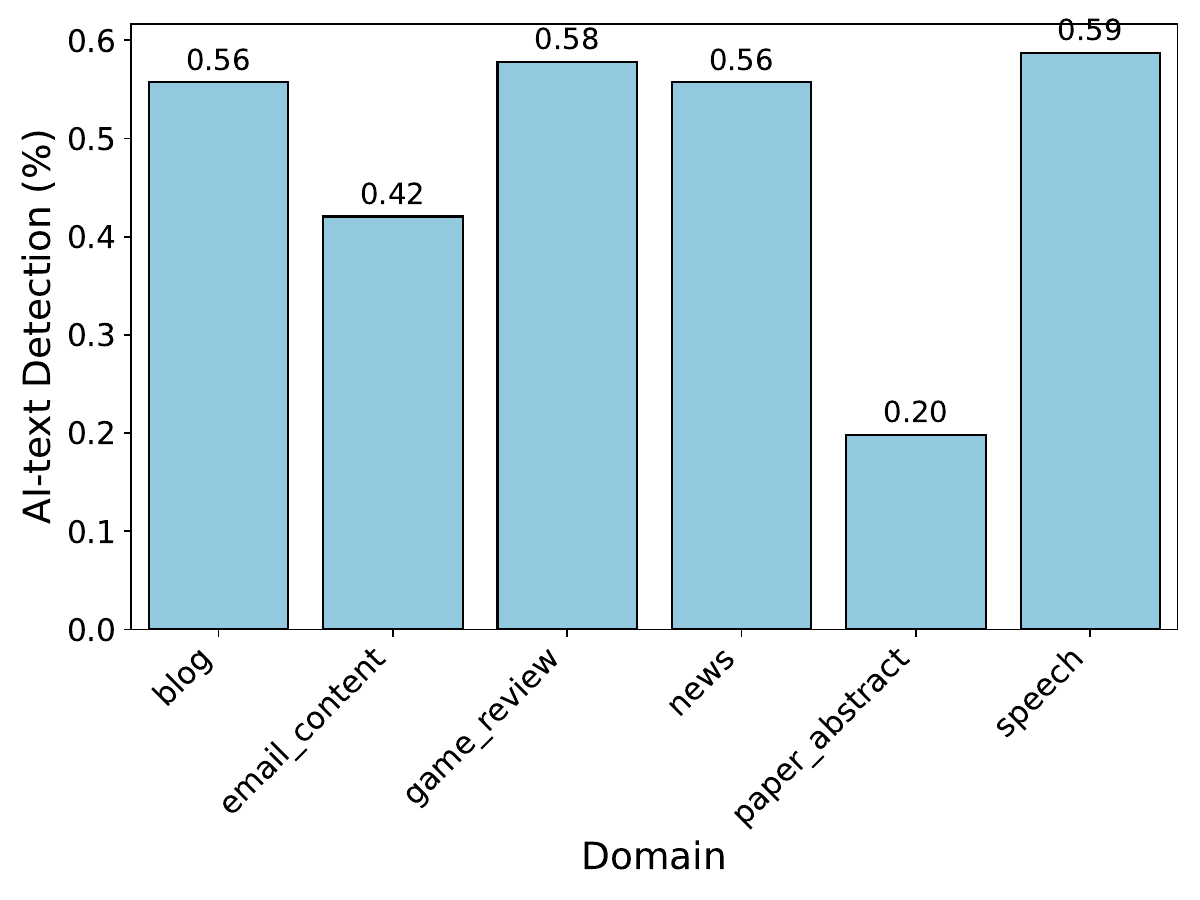}
            \caption{Llama3-8b}
        \end{subfigure}
        \hfill
        \begin{subfigure}{0.24\textwidth}
            \centering
            \includegraphics[width=\textwidth]{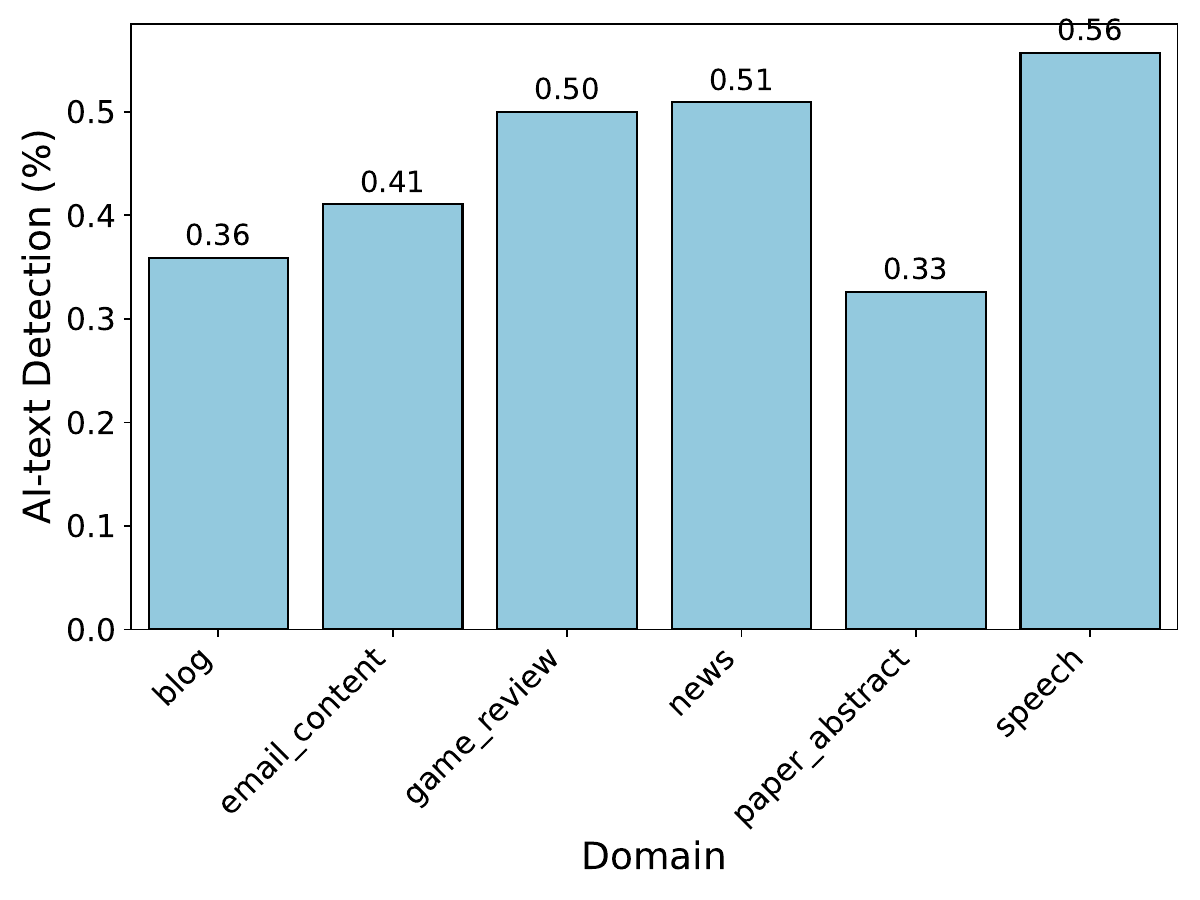}
            \caption{Llama2-7b}
        \end{subfigure}
        \\ 
        \vspace{0.2cm}
        \small \textbf{Slight-major Polishing}
    \end{minipage}

    \vspace{0.5cm} 
    \begin{minipage}{\textwidth}
    \centering
        \begin{subfigure}{0.24\textwidth}
            \centering
            \includegraphics[width=\textwidth]{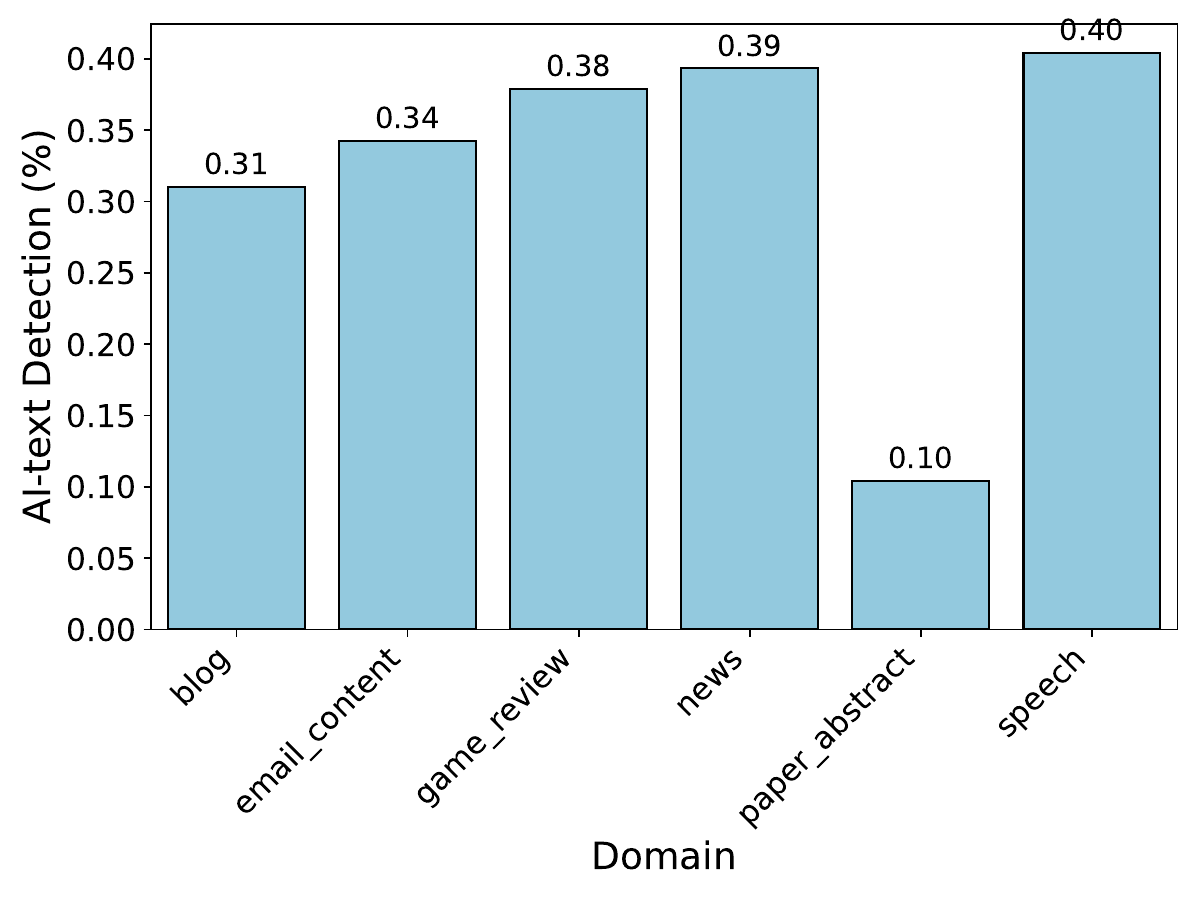}
            \caption{GPT-4o}
        \end{subfigure}
        \hfill
        \begin{subfigure}{0.24\textwidth}
            \centering
            \includegraphics[width=\textwidth]{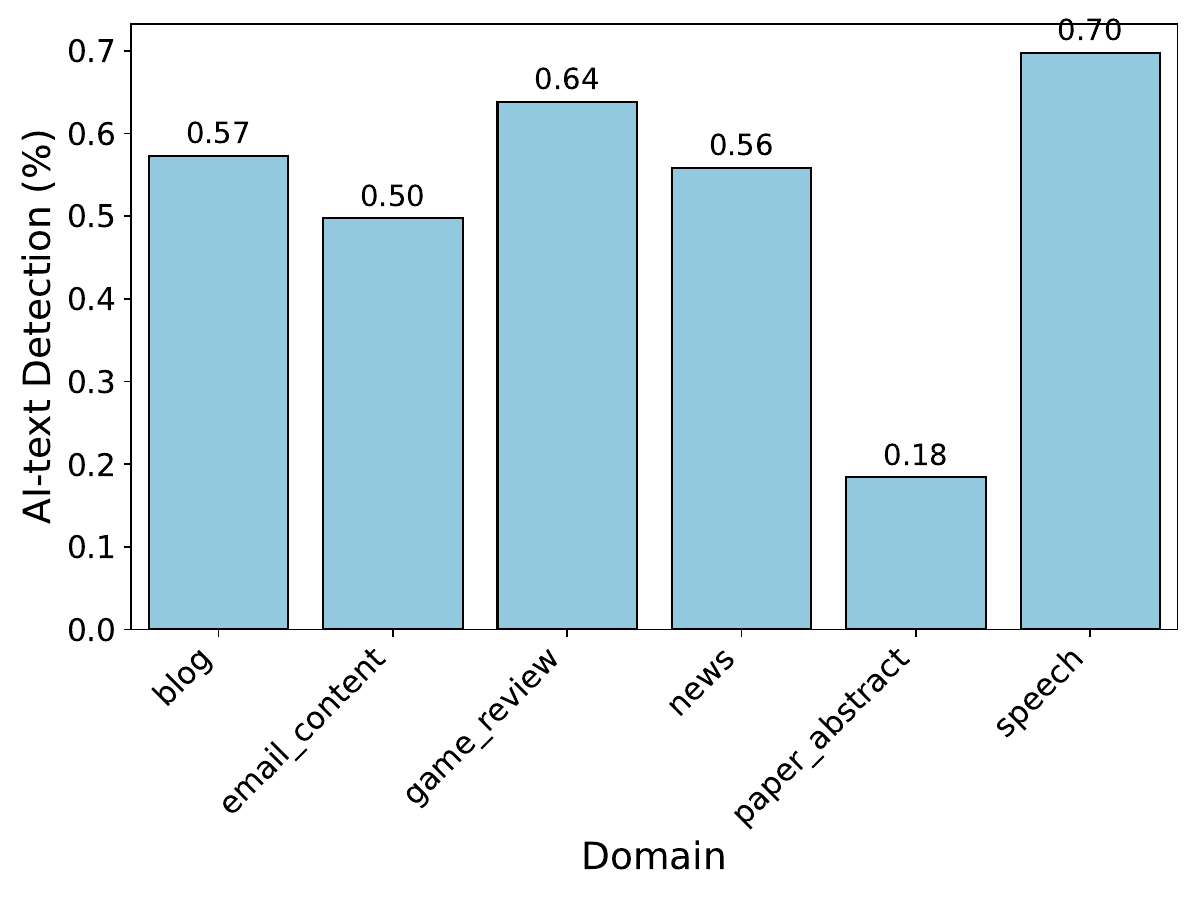}
            \caption{Llama3.1-70b}
        \end{subfigure}
        \hfill
        \begin{subfigure}{0.24\textwidth}
            \centering
            \includegraphics[width=\textwidth]{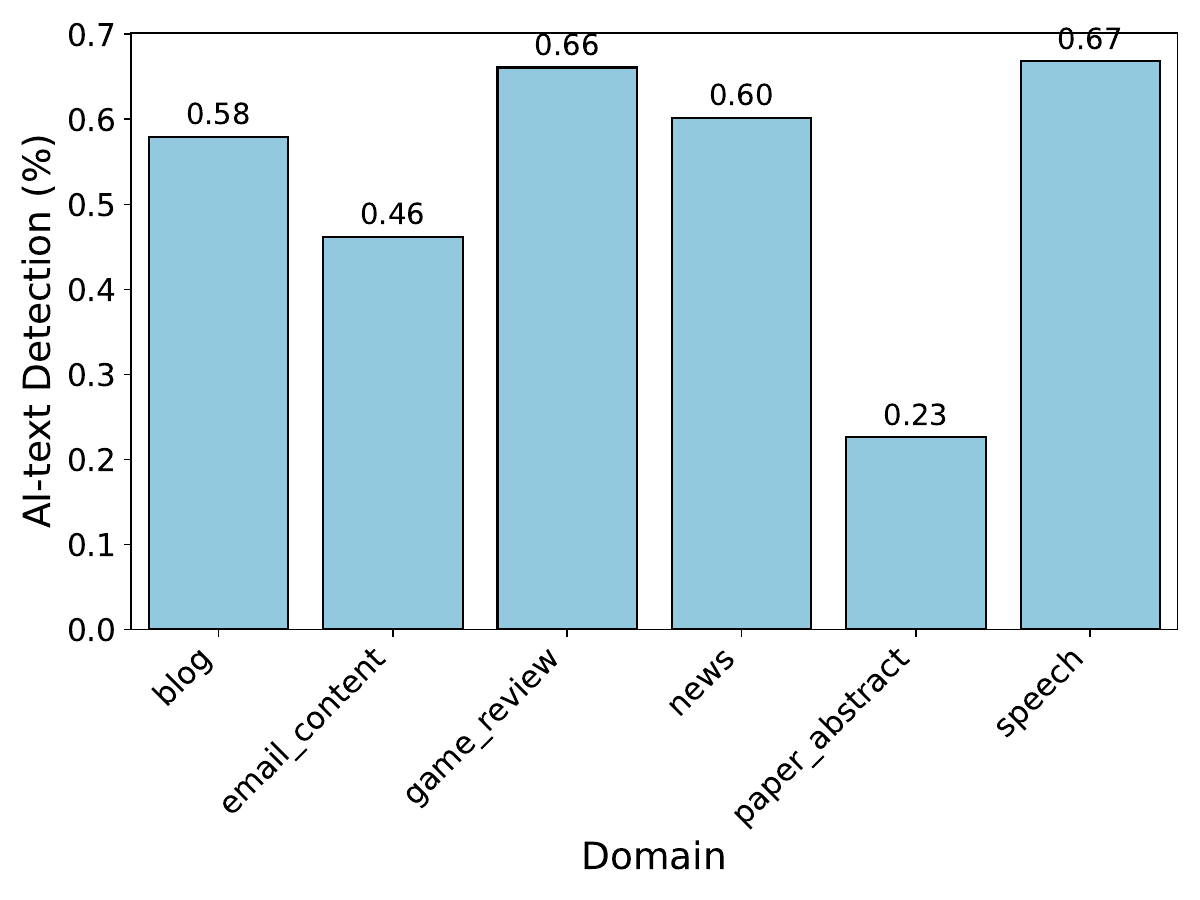}
            \caption{Llama3-8b}
        \end{subfigure}
        \hfill
        \begin{subfigure}{0.24\textwidth}
            \centering
            \includegraphics[width=\textwidth]{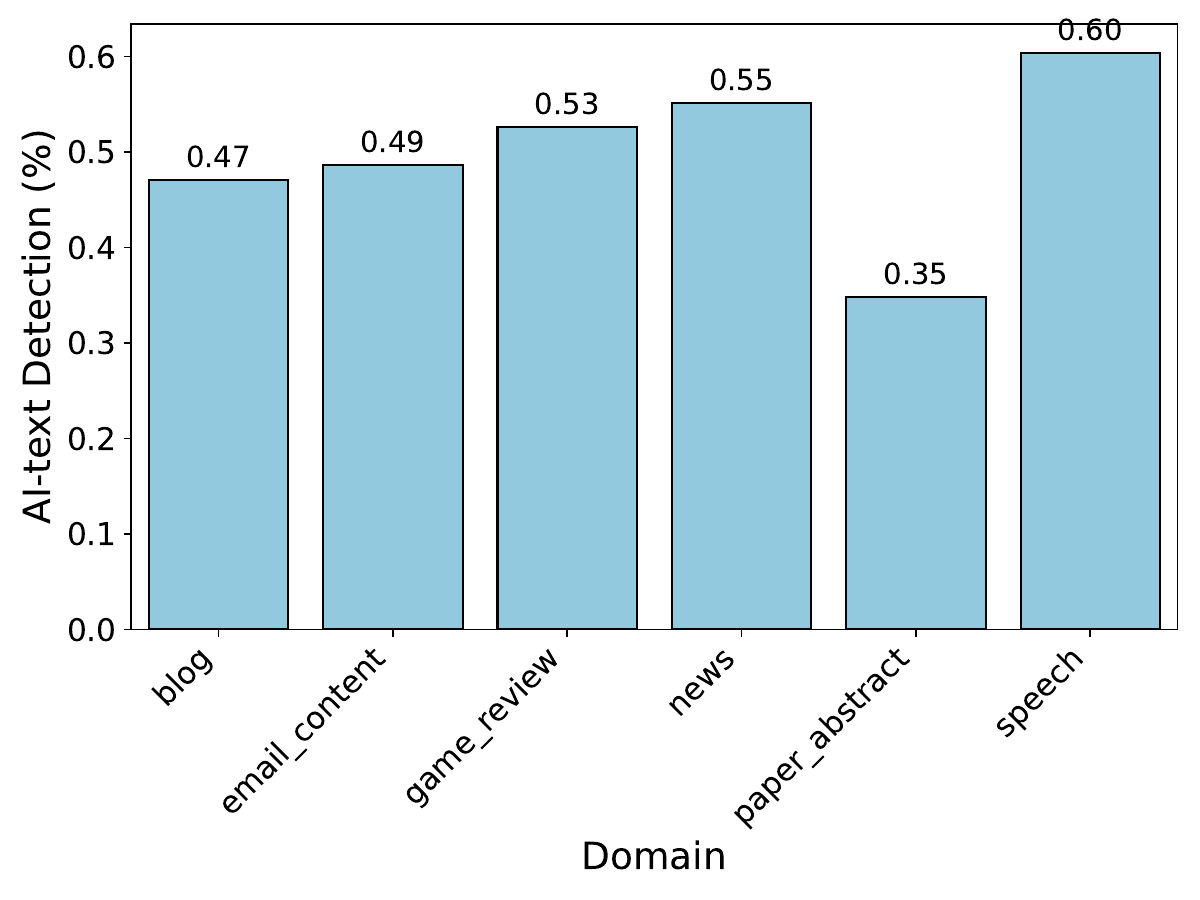}
            \caption{Llama2-7b}
        \end{subfigure}
        \\ 
        \vspace{0.2cm}
        \small \textbf{Major Polishing}
    \end{minipage}

    \vspace{0.5cm} 
    \begin{minipage}{\textwidth}
    \centering
        \begin{subfigure}{0.24\textwidth}
            \centering
            \includegraphics[width=\textwidth]{images_new/domain_plots/average_domain_accuracy_gpt.pdf}
            \caption{GPT-4o}
        \end{subfigure}
        \hfill
        \begin{subfigure}{0.24\textwidth}
            \centering
            \includegraphics[width=\textwidth]{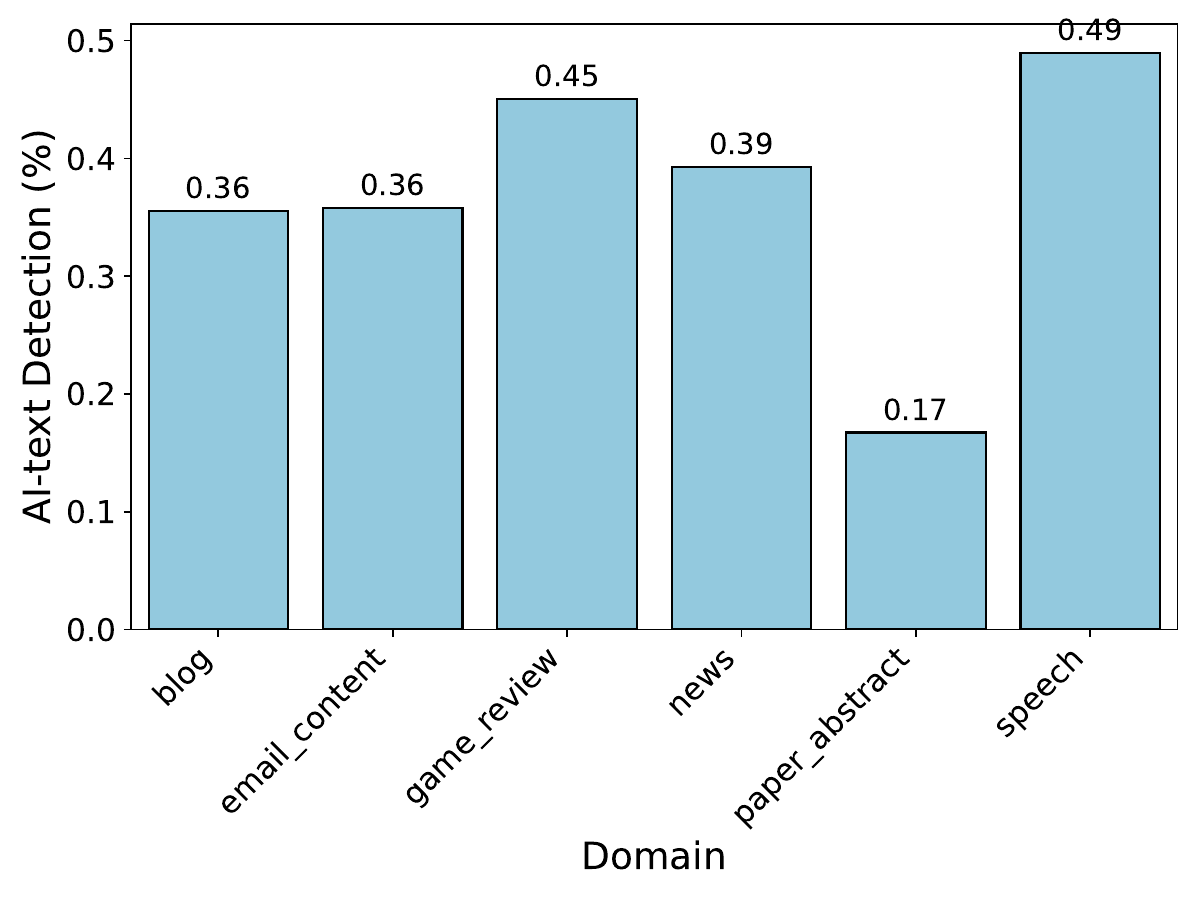}
            \caption{Llama3.1-70b}
        \end{subfigure}
        \hfill
        \begin{subfigure}{0.24\textwidth}
            \centering
            \includegraphics[width=\textwidth]{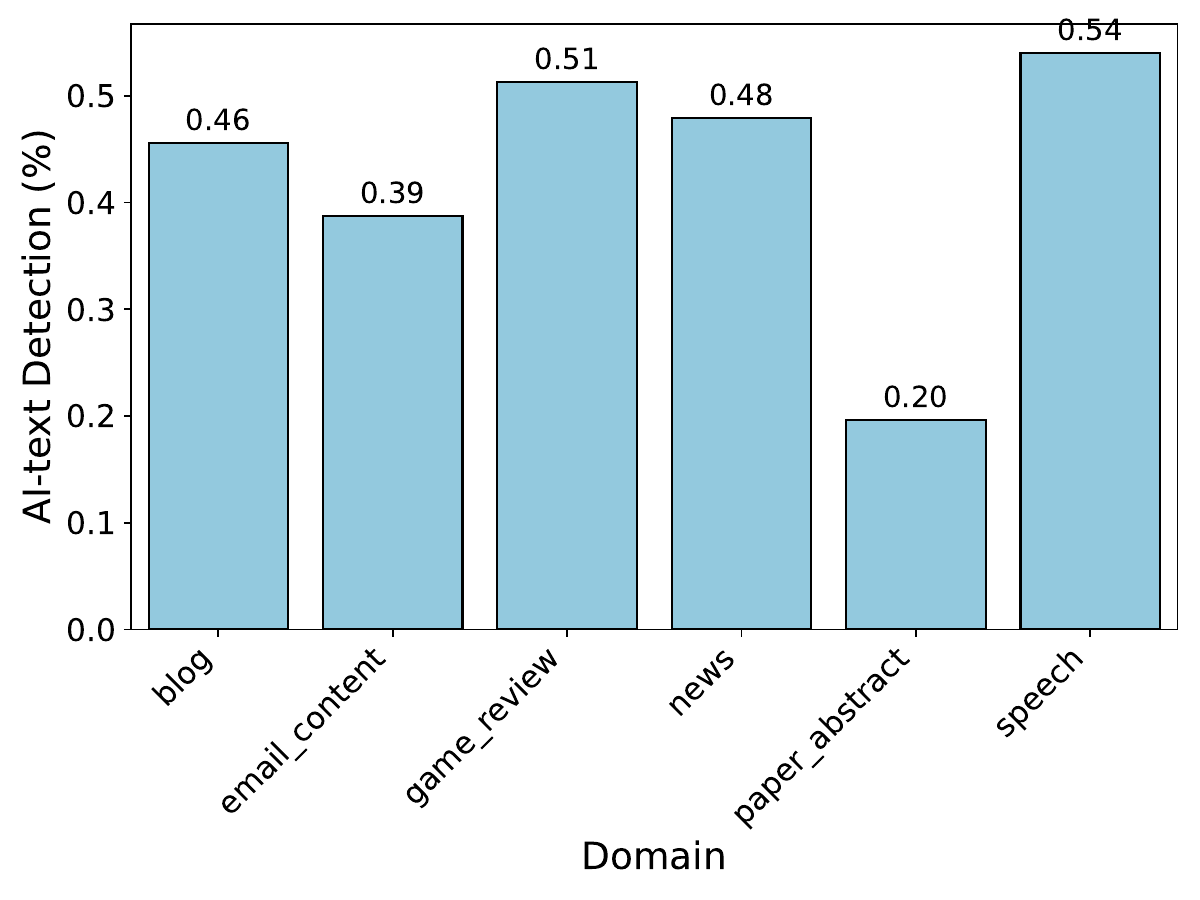}
            \caption{Llama3-8b}
        \end{subfigure}
        \hfill
        \begin{subfigure}{0.24\textwidth}
            \centering
            \includegraphics[width=\textwidth]{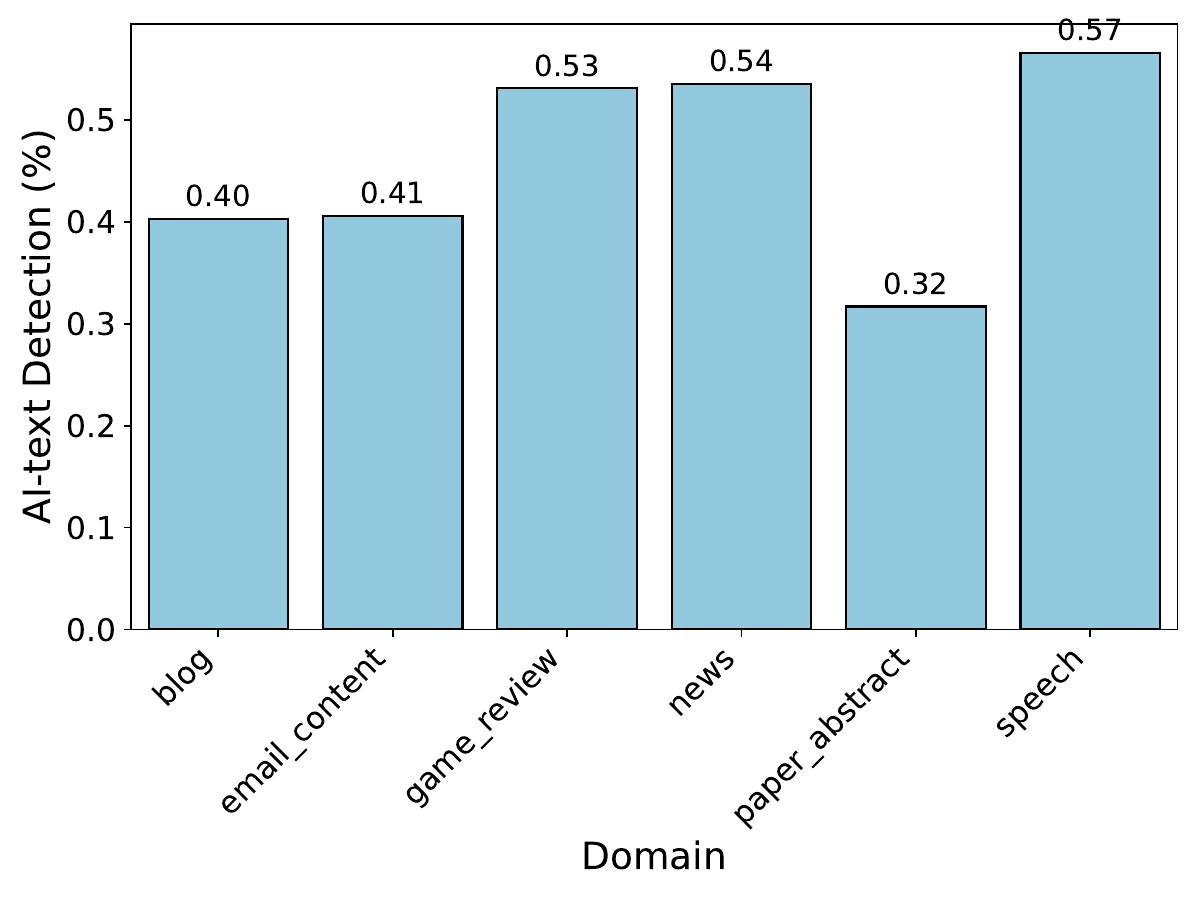}
            \caption{Llama2-7b}
        \end{subfigure}
        \\ 
        \vspace{0.2cm}
        \small \textbf{Average}
    \end{minipage}
    
    \caption{AI-text detection rate across all domains for different degree-based polishing.}
    \label{fig:domain_all_results}
\end{figure*}


\begin{figure*}[htbp]
    \centering
    \begin{minipage}{\textwidth} 
        \centering
        \begin{subfigure}{0.48\textwidth}
            \centering
            \includegraphics[width=\textwidth]{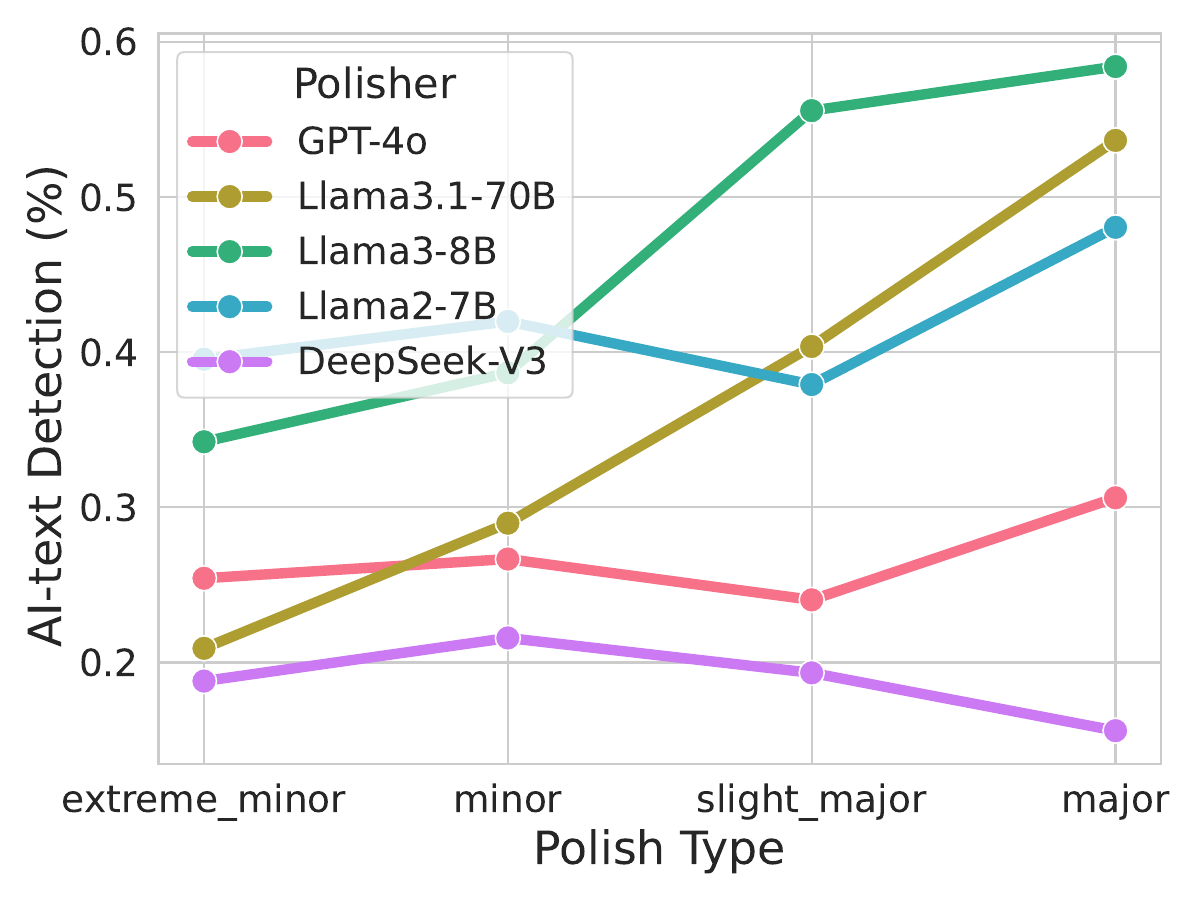}
            \caption{Blog}
        \end{subfigure}
        \hfill
        \begin{subfigure}{0.48\textwidth}
            \centering
            \includegraphics[width=\textwidth]{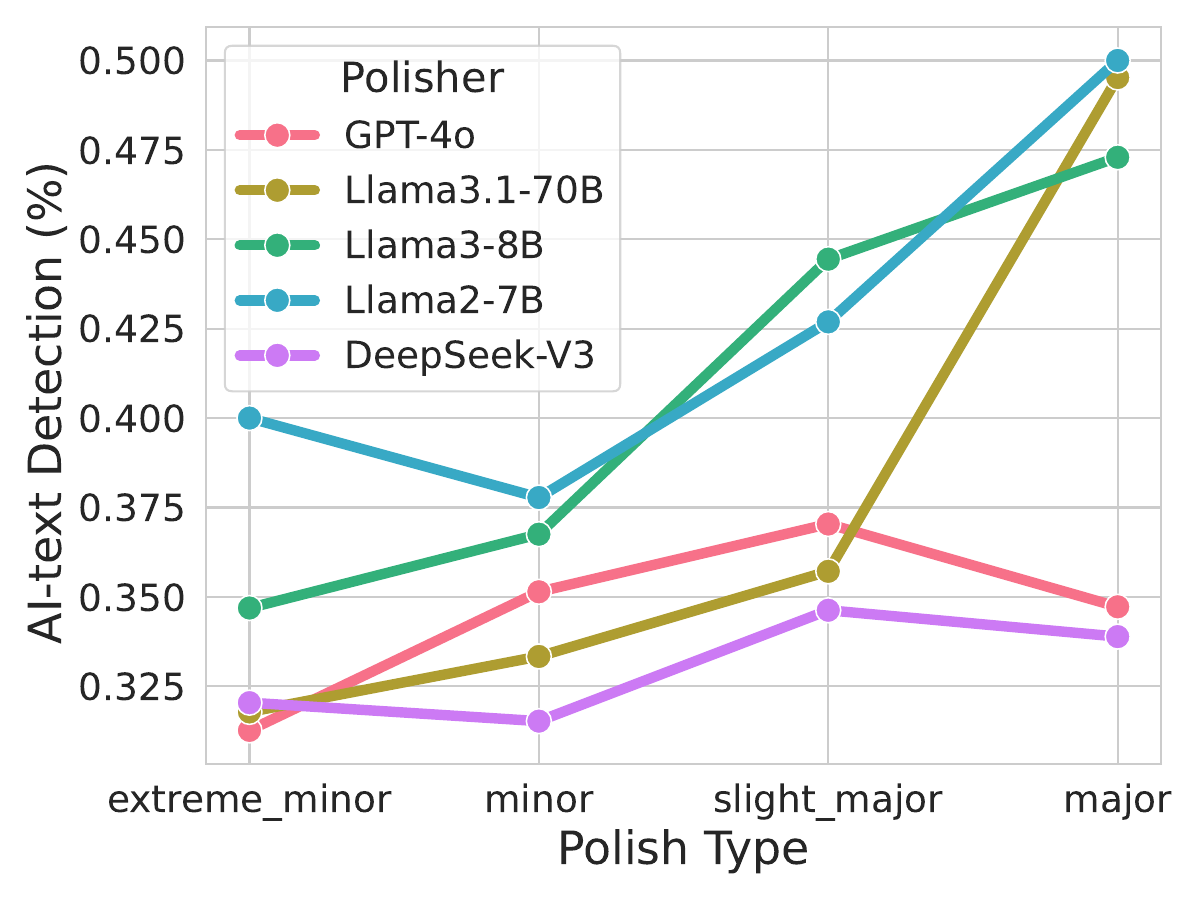}
            \caption{Email content}
        \end{subfigure}
    \end{minipage}
    
    \vspace{0.5cm} 
    \begin{minipage}{\textwidth}
    \centering
        \begin{subfigure}{0.48\textwidth}
            \centering
            \includegraphics[width=\textwidth]{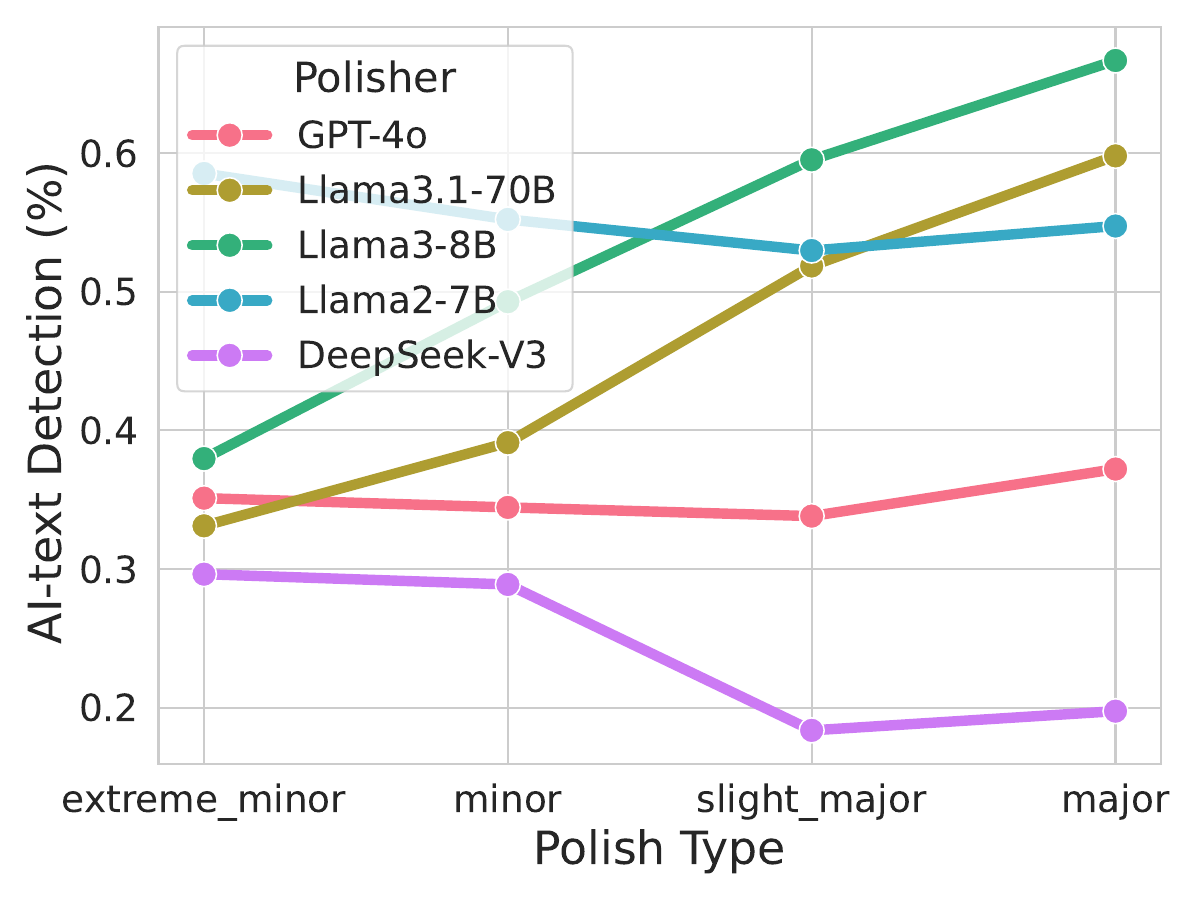}
            \caption{Game review}
        \end{subfigure}
        \hfill
        \begin{subfigure}{0.48\textwidth}
            \centering
            \includegraphics[width=\textwidth]{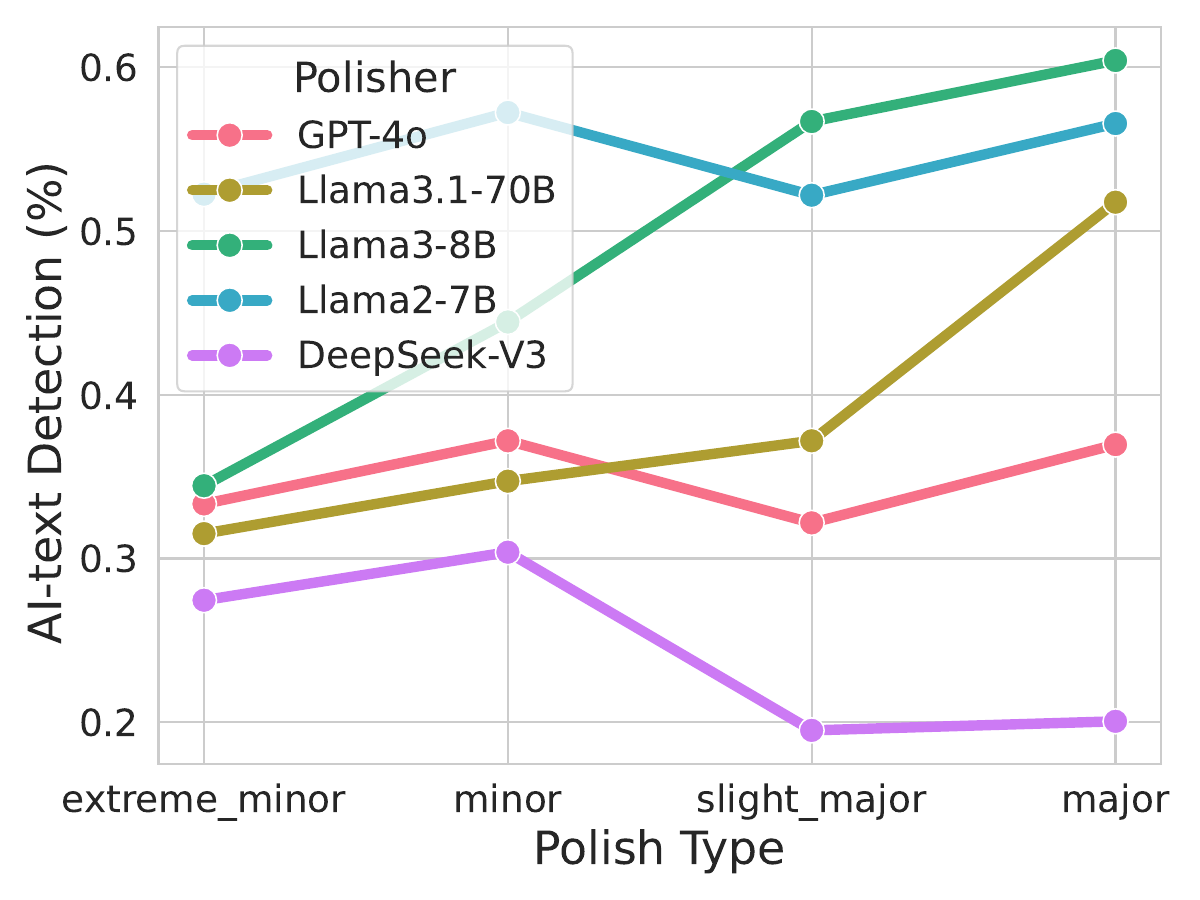}
            \caption{News}
        \end{subfigure}
    \end{minipage}

    \vspace{0.5cm} 
    \begin{minipage}{\textwidth}
    \centering
        \begin{subfigure}{0.48\textwidth}
            \centering
            \includegraphics[width=\textwidth]{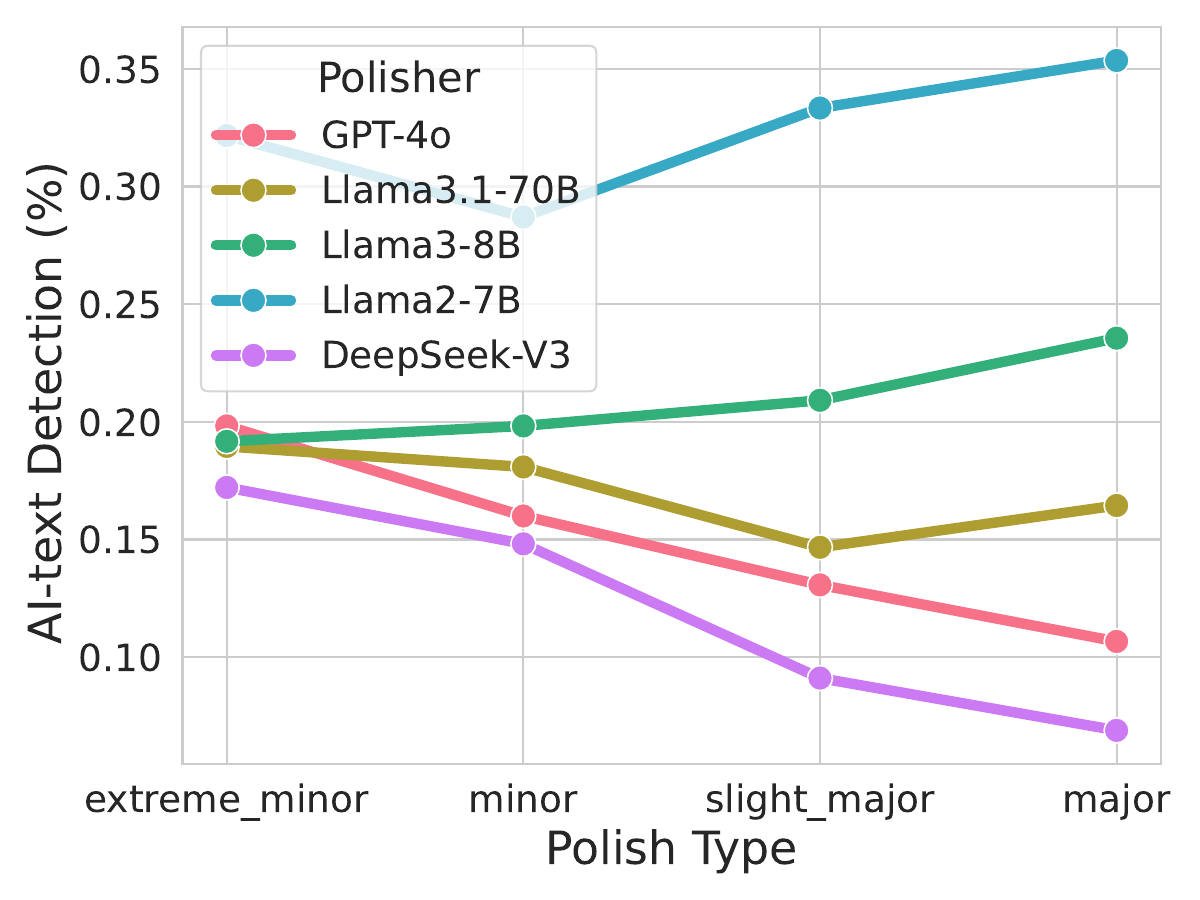}
            \caption{Paper Abstract}
            \label{fig:paper_abstract_change}
        \end{subfigure}
        \hfill
        \begin{subfigure}{0.48\textwidth}
            \centering
            \includegraphics[width=\textwidth]{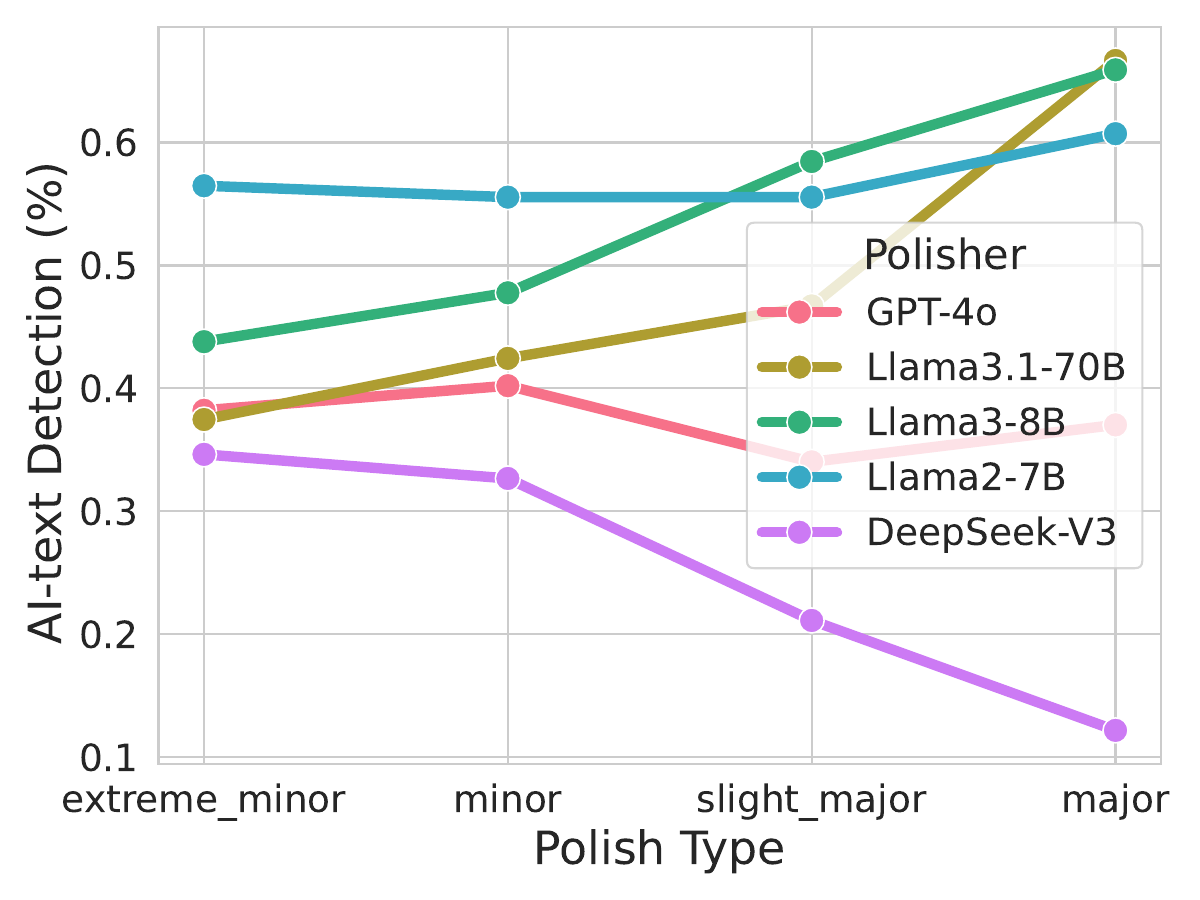}
            \caption{Speech}
        \end{subfigure}
    \end{minipage}
    
    \caption{Average AI-text detection for different domains across all polisher LLMs.}
    \label{fig:domain_change_plot}
\end{figure*}


\end{document}